\documentclass{article}

\usepackage{mathtools}
\usepackage{bbm}
\DeclarePairedDelimiter\ceil{\lceil}{\rceil}
\DeclarePairedDelimiter\floor{\lfloor}{\rfloor}
\usepackage{amsthm}

\usepackage[pdfauthor={Shahaf Bassan}, pdftitle={What makes an Ensemble (Un) Interpretable?}]{hyperref}

\usepackage{microtype}
\usepackage{graphicx}
\usepackage{subfigure}
\usepackage{multirow}
\usepackage{booktabs}
\usepackage[table]{xcolor}

\newcommand{\mysubsection}[1]{\medskip\noindent\textbf{#1}}

\newcommand{\z}{\textbf{z}}
\newcommand{\x}{\textbf{x}}
\newcommand{\y}{\textbf{y}}
\newcommand{\ub}{\textbf{u}}
\newcommand{\vb}{\textbf{v}}
\newcommand{\wb}{\textbf{w}}

\newcommand{\yes}{\textit{Yes}}
\newcommand{\GSSP}{\textit{GSSP}}
\newcommand{\kGSSP}{\textit{k-GSSP}}
\newcommand{\ckGSSP}{\textit{{k-GSSP}\textsuperscript{*}}}
\newcommand{\no}{\textit{No}}
\newcommand{\true}{\textit{True}}
\newcommand{\false}{\textit{False}}

\newcommand{\countPComplexity}{\#P}
\usepackage[accepted]{icml2025}

\usepackage{amsmath}
\usepackage{amssymb}
\usepackage{paralist}
\usepackage{enumitem}

\newcommand{\NPComplexity}{NP}

\newcommand{\StoPComplexity}{$\Sigma_{2}^{P}$} 
\newcommand{\StoPCompleteComplexity}{$\Sigma_{2}^{P}$-complete} 

\usepackage[capitalize,noabbrev]{cleveref}

\theoremstyle{plain}
\newtheorem{theorem}{Theorem}[section]
\newtheorem{proposition}[theorem]{Proposition}
\newtheorem{lemma}[theorem]{Lemma}

\newtheorem{observation}{Observation}
\newtheorem{claim}{Claim}

\theoremstyle{definition}
\newtheorem{definition}[theorem]{Definition}

\theoremstyle{remark}

\usepackage[textsize=tiny]{todonotes}

\icmltitlerunning{What makes an Ensemble (Un) Interpretable?}

\begin{document}

\twocolumn[
\icmltitle{What makes an Ensemble (Un) Interpretable?}




\begin{icmlauthorlist}
\icmlauthor{Shahaf Bassan}{yyy}
\icmlauthor{Guy Amir}{comp}
\icmlauthor{Meirav Zehavi}{zzz}
\icmlauthor{Guy Katz}{yyy}

\end{icmlauthorlist}

\icmlaffiliation{yyy}{The Hebrew University of Jerueslaem}
\icmlaffiliation{comp}{Cornell University}
\icmlaffiliation{zzz}{Ben-Gurion University of the Negev}

\icmlcorrespondingauthor{Shahaf Bassan}{shahaf.bassan@mail.huji.ac.il}

\icmlkeywords{ensembles, interpretability, explainability, computational complexity}

\vskip 0.3in
]



\printAffiliationsAndNotice{}  

\begin{abstract}

Ensemble models are widely recognized in the ML community for their limited interpretability. For instance, while a single decision tree is considered interpretable, ensembles of trees (e.g., boosted trees) are often treated as black-boxes. Despite this folklore recognition, there remains a lack of rigorous mathematical understanding of what particularly makes an ensemble (un)-interpretable, including how fundamental factors like the \begin{inparaenum}[(i)] \item \emph{number}, \item \emph{size}, and \item \emph{type} \end{inparaenum} of base models influence its interpretability. In this work, we seek to bridge this gap by applying concepts from computational complexity theory to study the challenges of generating explanations for various ensemble configurations. Our analysis uncovers nuanced complexity patterns influenced by various factors. For example, we demonstrate that under standard complexity assumptions like P$\neq$
NP, interpreting ensembles remains intractable even when base models are of constant size. Surprisingly, the complexity changes drastically with the number of base models: small ensembles of decision trees are efficiently interpretable, whereas interpreting ensembles with even a constant number of linear models remains intractable. We believe that our findings provide a more robust foundation for understanding the interpretability of ensembles, emphasizing the benefits of examining it through a computational complexity lens.

\end{abstract}

\section{Introduction}


Ensemble learning is a widely acclaimed technique in ML that leverages
the strengths of multiple models instead of relying on a single one. This approach has been proven to enhance predictive accuracy, mitigate variance, and handle imbalanced or noisy datasets effectively~\cite{dong2020survey, sagi2018ensemble}.

However, a significant challenge with ensemble models is their
perceived lack of interpretability~\cite{guidotti2018survey,
  sagi2021approximating, hara2018making, benard2021interpretable,
  kook2022deep}. The reason behind this is straightforward ---
utilizing several models simultaneously makes the
decision-making process inherently more complex, and thus more challenging to understand. For instance, while it is feasible to trace the decision-making path in a single decision tree, this level of straightforward traceability is not achievable in tree ensembles~\cite{sagi2021approximating, hara2018making, benard2021interpretable}.

Despite a general acknowledgment of this issue in the ML community, some fundamental questions remain regarding the precise nature of what makes an ensemble (un)-interpretable. For instance: \begin{inparaenum}[(i)]
    \item How much ``harder'' is it to interpret ensembles compared to their base models? Is the difference in complexity \emph{polynomial} or \emph{exponential}? 
    \item How does the interpretability of ensembles vary depending on the \emph{type of explanation} used? 
    \item How does the interpretability depend on the \emph{type of base models}? 
    \item How do structural parameters, such as the \emph{number} of base models or the \emph{size} of individual models, influence interpretability? For example, is an ensemble with many small models more interpretable than one with a few large models? Does this answer depend on the explanation type or the nature of the base models? \item Does \emph{simplifying} an ensemble by reducing the number or the size of base models enhance its interpretability?
\end{inparaenum}

In this work, we aim to address these questions by examining the interpretability of ensembles through the lens of \emph{computational complexity}. The computational investigation of interpretability has been explored in several recent studies~\cite{BaMoPeSu20, bassanlocal, ArBaBaPeSu21, adolfi2024computational}. Notably, \cite{BaMoPeSu20} introduced the term ``computational interpretability'' as a formal and mathematical approach to assessing interpretability, distinguishing it from the more common emphasis on human-centered understanding~\cite{rudin2022interpretable}. Within computational interpretability, a model is considered interpretable if explanations for its decisions can be generated efficiently (i.e., it is ``easy'' to explain). Conversely, if generating explanations is computationally challenging, the model is classified as uninterpretable.



\textbf{Our Contributions.} We perform a formal analysis to investigate the interpretability of ensemble models by studying the computational complexity of deriving various types of explanations in different settings and comparing these complexities to those of their individual base models. Our work encompasses a wide range of complexity scenarios, focusing on five distinct explanation types examined in previous research~\citep{BaMoPeSu20, bassanlocal, van2022tractability}. These include \begin{inparaenum}[(i)] \item feature selection explanations (covering three relevant forms), \item contrastive explanations, and \item Shapley value feature attributions. \end{inparaenum}

We additionally explore a broad range of ensemble configurations encompassing various interpretability scenarios and examine three key factors: \begin{inparaenum}[(i)] \item the \emph{type} of base model, \item the \emph{number} of base models in the ensemble, and \item the \emph{size} of the base models. \end{inparaenum} To examine diverse base-model types, we focus on three widely used base models that span the extremes of the interpretability spectrum: \begin{inparaenum}[(i)] \item linear models, \item decision trees, and \item neural networks. \end{inparaenum} To examine the number and size of base models, we explore their impact through concepts from \emph{parameterized complexity}~\cite{downey2012parameterized}, a key area of computational complexity that examines how various structural parameters affect complexity behavior.

\subsection{Negative Complexity Corollaries}

\begin{itemize}
    \item \textbf{Ensembles are computationally hard to interpret (and the explanation type matters).} We provide
      a range of intractability results (NP, $\Sigma^P_2$,
      $\#$P-Hardness, etc.) for generating multiple types of explanation, for diverse ensembles, emphasizing foundational limitations of computing explanations for these models. Interestingly, within these intractability results, we uncover substantial complexity differences among the analyzed explanation forms, demonstrating that certain forms are significantly more challenging than others.
\item \textbf{Ensembles are substantially \emph{less} interpretable than their base models (but the base model type matters).} To demonstrate that the intractability arises from the ensemble aggregation itself, we establish \emph{strict complexity separations} between explaining an ensemble's base models (e.g., linear models, decision trees) --- which can often be done in polynomial or linear time --- and explaining the ensembles themselves, which is computationally intractable (e.g., NP-hard). This highlights, from a computational standpoint, that ensembles are (\emph{exponentially}) ``less interpretable'' than their base models. Interestingly, we also show that no such complexity gaps exist for ensembles consisting of more expressive base models like neural networks.
\item \textbf{Even ensembles with significantly small base models remain hard to interpret.} 
Interestingly, we demonstrate that in a general setting --- including all the instances we examined --- ensembles consisting of even \emph{constant}-size base models remain computationally intractable to interpret (e.g., NP, $\Sigma^P_2$, $\#$P-Hard). From a practical standpoint, this means that even if a practitioner attempts to simplify an ensemble by drastically reducing the sizes of its base models, the overall ensemble remains computationally infeasible to interpret, highlighting a strikingly negative complexity result.
\item \textbf{Even ensembles with only two linear models are already computationally hard to interpret.} We present another strong negative complexity result that applies broadly: across all analyzed settings, ensembles of linear models (often regarded as highly interpretable) become intractable to interpret, when consisting of only two linear models. This underscores how rapidly the integration of linear models leads to a significant loss of interpretability, resulting in a substantially negative complexity outcome in a highly simplified scenario.

\end{itemize}

\subsection{Positive Complexity Corollaries}
\begin{itemize}

\item \textbf{Ensembles with a small number of decision trees can be interpreted efficiently.} 
Surprisingly, we observe strikingly different complexity behaviors between ensembles of linear models and those of decision trees. We present more optimistic complexity results, demonstrating that reducing the number of decision trees in an ensemble (e.g., Random Forests, XGBoost) allows for tractable (poly-time) computation of various explanation types --- unlike in the case of linear models. These findings open the door to practical and efficient algorithmic implementations in this context.
\end{itemize}

We believe these corollaries provide a more rigorous, and mathematically grounded perspective on ensemble interpretability. While some of our results confirm common ML beliefs, others reveal unexpected complexity variations based on model type, number, and size, underscoring the importance of a complexity analysis in this context.

Due to space constraints, we include an outline of our various theorems and corollaries within the paper, and relegate the complete proofs
to the appendix.

\section{Preliminaries}




\textbf{Complexity Classes.} This paper assumes that readers have a
basic understanding of standard complexity classes, including
polynomial time (PTIME) and nondeterministic polynomial time (NP and
coNP). We also discuss the common class within the second order of the
polynomial hierarchy, \StoPComplexity. This class contains problems
that can be solved within \NPComplexity{} when provided access to an
oracle capable of solving coNP problems in constant
time. It clearly holds that \NPComplexity{}, coNP
$\subseteq$ \StoPComplexity; and it is also widely believed that \NPComplexity{}, coNP $\subsetneq$ \StoPComplexity~\cite{arora2009computational}. The paper additionally discusses the complexity class \countPComplexity{} which represents the total count of accepting paths in a polynomial-time nondeterministic Turing machine. It is widely believed that
\StoPComplexity $\subsetneq$ \countPComplexity~\cite{arora2009computational}.

\textbf{Setting.} We consider a set of $n$ input features $\{1, \ldots, n\}$, represented by assignments $\x := (\x_1, \ldots, \x_n)$, within a feature space $\mathbb{F}$. The classifier $f: \mathbb{F} \to [c]$, where $c \in \mathbb{N}$ is the number of classes, is analyzed through \emph{local} explanations that interpret predictions for specific instances $\x$. We note that for clarity, we follow common conventions~\cite{ArBaBaPeSu21, waldchen2021computational, BaMoPeSu20, bassanlocal, adolfi2024computational} and assume Boolean inputs and outputs (i.e., $\mathbb{F} := \{0,1\}^n$, $c := 1$). This simplification aids readability, but our findings extend to discrete or real-valued inputs and multi-class classifiers. See Appendix~\ref{framework_extensions_section} for additional information.




\section{Base-Model and Explanation Types}

We aim to explore the interpretability of ensembles through a computational complexity perspective, encompassing as broad and diverse a range as possible. To achieve this, we analyze the complexity of various ensembles composed of base models spanning different points on the interpretability spectrum. Additionally, we examine the complexity of generating a wide array of explanation forms commonly studied in the literature. 

\subsection{Ensemble and Base-Model Types}

We define an ensemble as a classification function consisting of $k$ base models, each itself a classification function. Since our focus is on post-hoc explanations, we analyze the ensemble's \emph{inference} process, specifically for majority voting and weighted voting inference. These approaches cover a broad range of ensembles, including boosting, voting, and bagging ensembles (see Appendix~\ref{appendix:model_types} for details). In majority voting, the ensemble prediction is determined by the majority vote among the $k$ base models, while in weighted voting, predictions are aggregated based on weights assigned to each model. We examine three types of base models: \begin{inparaenum}[(i)]\item axis-aligned decision trees (following conventions~\citep{BaMoPeSu20, ArBaBaPeSu21, bassanlocal}), \item linear classifiers, and \item neural networks with ReLU activations. Formal definitions can be found in Appendix~\ref{appendix:model_types}.\end{inparaenum}


This formalization encompasses a wide range of ensemble techniques, including \emph{random forests}, \emph{boosted trees} (e.g., XGBoost), and other diverse ensembles composed of decision trees, neural networks, or linear models (e.g., logistic regression, SVMs). While our focus is on classification models, many findings also apply to regression (see Appendix~\ref{framework_extensions_section}), making them relevant for methods like linear regression. Although we primarily address \emph{homogeneous} ensembles (identical model types), several results extend to \emph{heterogeneous} ensembles (mixed model types), as detailed in Appendix~\ref{framework_extensions_section}.


\subsection{Explanation Types}

We focus on several widely recognized forms of explanations from the literature. In line with previous research~\cite{BaMoPeSu20, ArBaBaPeSu21, bassanlocal}, we conceptualize each type of explanation as an \emph{explainability query}. An explainability query takes both $f$ and $\x$ as inputs and aims to address specific inquiries while providing some type of interpretation for the prediction $f(\x)$. 


\textbf{Sufficient reason Feature Selection.}  We consider the common \emph{sufficiency} criterion for feature selection, which is based on common explainability methods~\cite{ribeiro2018anchors, carter2019made, ignatiev2019abduction}. A  \emph{sufficient reason} is
a subset of input features, $S\subseteq [n]$, such that when we fix the features of $S$ to their corresponding values in $\x\in\mathbb{F}$, then
the
prediction always remains $f(\x)$, regardless of any different assignment to the features in the subset $\overline{S}$. We use the notation of $(\mathbf{x}_S;\mathbf{z}_{\Bar{S}})$ to denote an assignment where the values $\x$ are assigned to $S$ and the values of $\z$ are assigned to $\overline{S}$. We can hence formally define $S$ to be a sufficient reason with respect
to $\langle f,\x\rangle$ iff it holds that for all $\z\in \mathbb{F}$: $f(\x_{S};\z_{\Bar{S}})= f(\x)$.


		
		

A typical assumption that is made in the literature suggests that
smaller sufficient reasons (that is, those with a lesser cardinality
of $|S|$) are more useful than larger ones~\cite{ribeiro2018anchors,
carter2019made, ignatiev2019abduction}. This leads to a particular
interest in obtaining \emph{cardinally minimal sufficient reasons},
also referred to as \emph{minimum sufficient reasons}, and consequently
to our first explainability query:

\noindent\fbox{%
	\parbox{\columnwidth}{%
		\mysubsection{MSR (Minimum Sufficient Reason)}:
		
		\textbf{Input}: Model $f$, input $\x$, and $d\in\mathbb{N}$
		
		\textbf{Output}: 
		\yes{} if there exists some $S\subseteq [n]$ such that $S$ is a sufficient reason with respect to $\langle f,\x\rangle$ and $|S|\leq d$, and \no{} otherwise.
	}%
}

To provide a comprehensive understanding of the complexity results of sufficient reasons, we study two additional common feature selection explainability queries~\cite{BaMoPeSu20, bassanlocal} which represent refinements of the MSR query: \begin{inparaenum}[(i)] \item 
    The \emph{Check-Sufficient-Reason} (\emph{CSR}) query, which, given a subset $S$, checks whether it is a sufficient reason; \item The \emph{Count-Completions} (\emph{CC}) query, which represents a generalized version of the CSR query, where given a subset of features, we return the relative portion of assignments that maintain a prediction. This form of explanation relates to the probability of maintaining a classification.
\end{inparaenum} Due to space constraints, we relegate the full
formalization of the CSR and CC queries to
Appendix~\ref{Additional_query_formalizations}.

\textbf{Contrastive Explanations.} An alternative approach to
interpreting models involves examining subsets of features which, when
modified, could lead to a change in the model's
classification~\cite{dhurandhar2018explanations,
  guidotti2022counterfactual}. We consider a subset $S\subseteq [n]$
as \emph{contrastive} if changing its values has the potential to
alter the original classification $f(\x)$; or, more formally, if there exists some $\z\in \mathbb{F}$ for which $f(\x_{\Bar{S}};\z_{S})\neq f(\x)$. Similarly to sufficient reasons, smaller contrastive reasons are generally assumed to be more meaningful. These represent the minimum change required to alter the original prediction. Hence, it is also natural to focus on \emph{cardinally-minimal contrastive reasons}. 

\vspace{0.5em} 

\noindent\fbox{%
	\parbox{\columnwidth}{%
		\mysubsection{MCR (Minimum Change Required)}:
		
		\textbf{Input}: Model $f$, input $\x$, and $d\in\mathbb{N}$.
		
		\textbf{Output}: \yes{}, if there exists some contrastive reason $S$ such that $|S| \leq d$ for $f(\x)$, and \no{} otherwise.
	}%
}

\vspace{0.5em} 

\textbf{Shapley Values.} In the additive feature attribution setting, each
feature $i\in[n]$ is assigned an importance weight $\phi_i$. A common
method for  allocating  weights is by using the \emph{Shapley value} attribution index~\cite{lundberg2017unified}, defined as follows:

\begin{equation}
    \label{eq:explanation}
    \phi_i(f,\x):=\sum_{S\subseteq [n]\setminus \{i\}}\frac{|S|!(n-|S|-1)!}{n!}(v(S\cup \{i\})-v(S))
\end{equation}

where $v(S)$ is the \emph{value function}, and we use the common
\emph{conditional expectation} value function
$v(S):=\mathbb{E}_{\z\sim \mathcal{D}_p}[f(\z)|
\z_S=\x_S]$~\cite{sundararajan2020many, lundberg2017unified}.

In our complexity analysis, we assume feature independence, which follows common practice in computational complexity frameworks~\cite{arenas2023complexity, van2022tractability}, as well as in practical methods for computing Shapley values, such as the KernelSHAP approach in the SHAP library~\cite{lundberg2017unified}. For a complete formalization, refer to Appendix~\ref{Additional_query_formalizations}.

\noindent\fbox{%
	\parbox{\columnwidth}{%
		\mysubsection{SHAP (Shapley Additive Explanation)}:
		
		\textbf{Input}: Model $f$, input \x, and $i\in[n]$
		
		\textbf{Output}: The shapley value $\phi_i(f,\x)$.
	}%
}



\section{Ensemble vs. Base-Model Interpretability}
\label{Non_paramterized_section}


We seek to compare the complexity of solving an explainability query
$Q$ for an ensemble, to that of solving it for the ensemble's
constituent base models. We are particularly interested in cases where
there exists \emph{strict computational complexity gaps} between the
two settings (e.g., solving $Q$ for a single base model can be
performed in polynomial time, whereas solving it for the ensemble is
NP-Complete). Identifying such gaps is particularly important as they pinpoint the specific situations where the ensemble aggregation process directly influences the (lack of) interpretability. To identify these gaps, we use the notion of \emph{c-interpretability} (computational interpretability) as defined in \cite{BaMoPeSu20}.:

\begin{definition}

Let $\mathcal{C}_1$ and $\mathcal{C}_2$ be two classes of models and let $Q$ be an explainability query for which $Q(\mathcal{C}_1)$ is in complexity class $\mathcal{K}_1$ and $Q(\mathcal{C}_2)$ is in complexity class $\mathcal{K}_2$. We say that \emph{$\mathcal{C}_1$ is strictly more c-interpretable than $\mathcal{C}_2$ with respect to $Q$} iff $Q(\mathcal{C}_2)$ is hard for the complexity class $\mathcal{K}_2$ and $\mathcal{K}_1\subsetneq \mathcal{K}_2$. 
\end{definition}

\subsection{Complexity Gap in Simple Base Models}

We begin with examining two types of base-models known for their simplicity and interpretability --- decision trees and linear models. Our results affirm the existence of a complexity gap when an ensemble consists of these base-models, as depicted in Table~\ref{table:Complexityclassgap}. Our initial complexity results are presented in the following proposition, with the proof provided in Appendix~\ref{non_paramaterized_appendix_section}:

\begin{table*}[h]
        \small 
	\centering
	\def\arraystretch{1.31}%
	\setlength{\tabcolsep}{0.95em} 
	\caption{The ``complexity gap'' when interpreting a single model and an ensemble of models in the case of decision trees and linear models. Cells highlighted in blue represent novel
                results, presented here; whereas the rest were
                already known previously.}
	\begin{tabular}{lccccccc}
		\\
		\toprule
		\multirow{3}{*}{} &
		\multicolumn{2}{c}{\textbf{Decision Trees}} &
		\multicolumn{2}{c}{\textbf{Linear Models}} \\
		& {\ \ Base-Model} & {\ Ensemble} & {\  Base-Model} & { Ensemble} \\
		\midrule
		\textbf{Check Sufficient Reason (CSR)} & PTIME & coNP-C  & PTIME  &   
		\cellcolor{blue!10}coNP-C  \\
    	\textbf{Minimum Contrastive Reason (MCR)} & PTIME & NP-C  & PTIME  &   
		\cellcolor{blue!10}NP-C \\
		\textbf{Minimum Sufficient Reason (MSR)} & NP-C & \cellcolor{blue!10}$\Sigma^P_2$-C  & PTIME  &   
		\cellcolor{blue!10}$\Sigma^P_2$-C  \\
  		\textbf{Count Completions (CC)} & PTIME & \cellcolor{blue!10}$\#$P-C  & $\#$P-C  &   
		\cellcolor{blue!10}$\#$P-C  \\
    	\textbf{Shapley Additive Explanations (SHAP)} & PTIME & $\#$P-H  & $\#$P-C  &   
		\cellcolor{blue!10}$\#$P-H  \\
		\bottomrule
	\end{tabular}
	\label{table:Complexityclassgap}
\end{table*}

\begin{proposition}
\label{main_result_non_paramterized_main_text}
    Ensembles of decision trees and ensembles of linear models are
    \begin{inparaenum}[(i)] \item coNP-Complete with respect to CSR, \item NP-Complete with respect to MCR, \item $\Sigma^P_2$-Complete with respect to MSR, \item $\#P$-Complete with respect to CC, and \item $\#P$-Hard with respect to SHAP.\end{inparaenum}
\end{proposition}

Where the results for majority-voting decision tree ensembles under the CSR and MCR queries were shown in~\citep{ordyniak2024explaining}, and we provide all of the remaining results. Based on these findings, we leverage our previously established notation to deduce an interpretability separation between ensembles of decision trees/linear models and their corresponding base models:

\emph{Proof Sketch.} We build upon a
proof from~\cite{izzaexplaining} that reduces DNF formulas to random
forest models to demonstrate that computing prime implicants for
random forests is $D^P$-Complete. We expand this by showing that DNF
formulas can be transformed into ensembles of decision trees or linear models in
polynomial time, fitting our broader category of \emph{poly-subset
  constructable functions} which efficiently represent any disjunction
of literals. We confirm that decision trees and linear models belong to this
category and proceed to obtain the complexity of various queries
through reductions: particularly, the MSR query reduction is obtained from the
Shortest-Implicant-Core problem~\cite{umans2001minimum}. We note
that while~\cite{audemard2022trading} aims to address the MSR query through a reduction from minimal
unsatisfiable sets to DNFs, there is a noted technical gap in this
proof (details in Appendix~\ref{proof_of_msr_appendix}), also observed
in other similar proofs~\cite{huang2021efficiently}. To the best of
our knowledge, we are the first to address this issue effectively with
a non-trivial approach, enabling us to confirm $\Sigma^P_2$-Hardness
for \emph{DNFs} and related ensembles of poly-subset-constructable functions.

\begin{theorem}
Decision trees are strictly more c-interpretable than ensembles of decision trees with respect to CSR, MSR, MCR, CC, and SHAP. The same result holds for linear models (and ensembles of linear models) with respect to CSR, MSR, and MCR.
\end{theorem}



In the previous theorem, there is no complexity gap for the CC
and SHAP queries in the context of linear models. Nevertheless,
it is still possible to demonstrate a complexity separation if we
assume that the weights and biases are given in unary form, a concept often termed as \emph{pseudo-polynomial time}. This is demonstrated in the following proposition, with its proof provided in Appendix~\ref{pseudo_plynomial_theorem_appendix_section}.

\begin{proposition}
\label{CC_and_SHAP_proposition}
    While CC and SHAP can be solved in pseudo-polynomial time for linear models, ensembles of linear models remain $\#P$-Hard even if the weights and biases are given in unary. Therefore, assuming that the weights and biases are given in unary, linear models are strictly more c-interpretable than ensembles of linear models with respect to CC and SHAP.
\end{proposition}

\emph{Proof sketch.} The findings for ensembles are derived from those in Proposition~\ref{main_result_non_paramterized_main_text}. We base the pseudo-polynomial algorithm for CC on the work of~\cite{BaMoPeSu20}. Additionally, we achieve similar outcomes for SHAP using a non-trivial dynamic programming algorithm that solves the SHAP query for Perceptrons in pseudo-polynomial time.

\subsection{No Complexity Gap in Complex Base Models}

We revealed a complexity gap between interpreting simple models (e.g., decision trees, linear models) and their ensemble counterparts. However, our findings show no such gap for neural networks. In fact, we establish a much stronger result: this holds for \emph{any} explainability query whose complexity class remains consistent under polynomial reductions, including all $\mathcal{K}$-Complete queries within the polynomial hierarchy (e.g., PTIME, NP, $\Sigma^P_2$) and their corresponding counting classes (e.g., $\#$P).


\begin{proposition}
\label{no_gap_original_paper_prop}
There is no explainability query $Q$ for which the class of neural networks is strictly more c-interpretable than the class of ensemble-neural networks.
\end{proposition}

The proof is provided in Appendix~\ref{no_gap_in_complex_models_appendix_sec}. The result is intuitive: since an ensemble of neural networks can be reduced to a single network in polynomial time, no distinct complexity classes (closed under polynomial reductions) differ in complexity between interpreting a single model and an ensemble. We generalize this property to other models, calling it \emph{closed under ensemble construction}.


\begin{definition}
We say that a class of models $\mathcal{C}$ is \emph{closed under ensemble construction} if given an ensemble $f$ containing models from $\mathcal{C}$, we can construct in polynomial time a model $g\in \mathcal{C}$ for which $\forall\x\in\mathbb{F}, f(\x)=g(\x)$.
\end{definition}

Clearly, the aforementioned property correlates to the expressiveness of the base model and does not apply to linear models or decision trees (assuming 
P$\neq$NP).

\section{Impact of Base-Model Count and Size on Ensemble Interpretability}
\label{paramterized_complexity_part_main_paper_section}
\begin{figure*}[h]
	\centering
	\hspace*{-1cm}\includegraphics[width=0.69\textwidth]{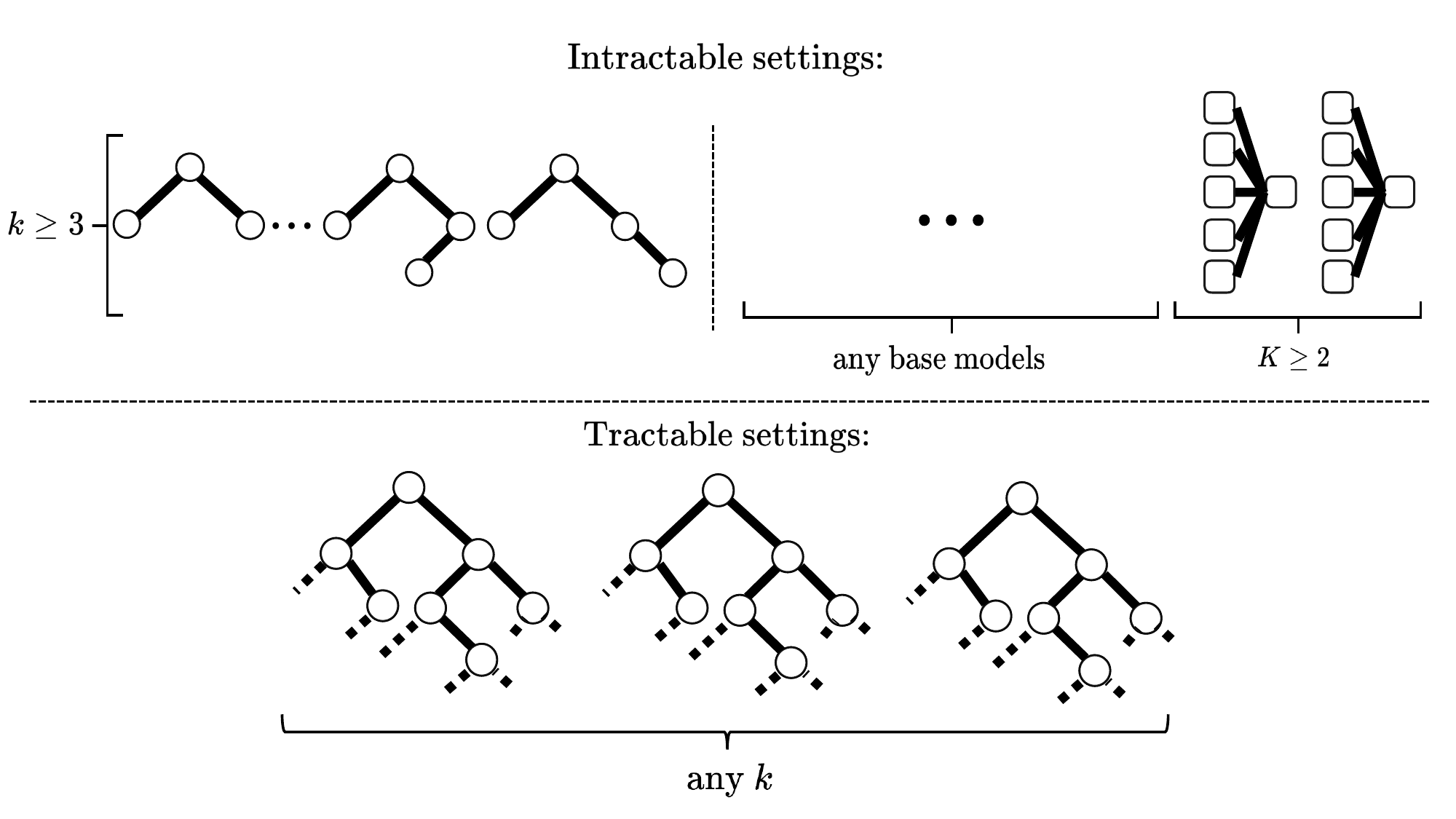}
	\caption{Illustration of insights from our parameterized complexity results: even highly simplified ensembles with constant-size base models (e.g., three base models) remain intractable to interpret. Moreover, ensembles with just two linear models already pose intractability, highlighting the substantial difficulty of interpreting linear model ensembles, even in simplified cases. However, reducing the number of trees in tree ensembles (e.g., Random Forest, XGBoost) can make explanation computation tractable if the number of trees is fixed.
}\label{fig:pdfplot}
\end{figure*}

Up until now, our analysis did not consider any structural
parameters. For example, the MCR query is polynomial-time solvable for
a single decision tree, but becomes NP-Complete for an ensemble of $k$
models. This raises questions about how different parameters, such as
the size of the participating base-models and the number of base
models, contribute to this effect. We explore these aspects by
using \emph{parameterized
  complexity}~\cite{downey2012parameterized}, an important branch of
computational complexity theory, to study how different parameters
affect the entire complexity of interpreting ensembles. We begin by outlining
some fundamental concepts in parameterized complexity, which are
crucial for this study. In this field, we deal with two-dimensional
instances denoted as $\langle \mathcal{X},k\rangle$ where
$\mathcal{X}$ represents the original encoding, and $k$ is a specified
parameter. We briefly describe the three main scenarios in
parameterized complexity.

 \textbf{1. Problems solvable in $|\mathcal{X}|^{O(1)}\cdot g(k)$ time.}   This is the best-case scenario concerning the parameter $k$ and includes \emph{fixed parameter tractable} (\emph{FPT}) problems concerning $k$, meaning their complexity is \emph{primarily controlled by the parameter $k$}. For example, the Vertex-Cover problem is FPT when $k$ is the vertex cover size, allowing efficient solutions even for large graphs if $k$ is small.


\textbf{2. Problems solvable in $O(|\mathcal{X}|^{g(k)})$ time.} This class includes XP problems, solvable in $O(|\mathcal{X}|^{g(k)})$ time. For fixed $k$, they are polynomial-time solvable, but large $|\mathcal{X}|$ can make them hard, even for small $k$. The W-hierarchy~\cite{downey2012parameterized}, including $W[t]$ for $t \geq 1$ and W[P], provides a finer classification within XP. For example, \emph{Clique} is $W[1]$-Complete. Here, $t$ relates to the Boolean circuit depth used in reductions (Appendix~\ref{parametrieed_background}), with W[P] allowing arbitrary depth. It is assumed that FPT $\subsetneq$ W[1] $\subsetneq$ W[2] $\subsetneq \ldots \subsetneq$ XP~\cite{downey2012parameterized}. Thus, XP problems lack FPT algorithms and remain inefficient to solve for W[1]-Hard cases if $|\mathcal{X}|$ grows arbitrarily large.

\textbf{3. Problems that are NP-Hard when $k$ is constant.} At the extreme of the parameterized complexity spectrum is \emph{para-NP}, representing the highest sensitivity to the parameter $k$. A problem is para-NP-Hard if it remains NP-hard even when $k$ is fixed. Assuming P$\neq$NP, we have XP $\subsetneq$ para-NP. For example, the Graph-Coloring problem is NP-Hard for any $k \geq 3$, meaning that even for a significantly small $k$, the problem is intractable --- unlike FPT and XP.



\textbf{Extensions of parameterized complexity.} We briefly mention extensions of parameterized complexity to counting problems~\cite{flum2004parameterized} and higher orders of the polynomial hierarchy~\cite{de2019parameterized} (e.g., $\Sigma^P_2$). This includes the $\#W$-hierarchy, extending the W-hierarchy to counting problems, where it is believed that FPT $\subsetneq \#$W[1] $\subsetneq \#$W[2] $\subsetneq \ldots \#$W[P] $\subsetneq$ XP. Similarly, XNP extends XP to the second order of the polynomial hierarchy, with XNP $\subsetneq$ para-$\Sigma^P_2$~\cite{de2017parameterized}. The concept of para-NP also generalizes to other classes, including para-coNP, para-$\Sigma^P_2$, and para-$\#$P~\cite{de2019parameterized}, where a problem is para-$\mathcal{K}$-Hard if it remains $\mathcal{K}$-Hard when $k$ is constant.

\subsection{Impact of Base-Model Sizes on Ensemble Interpretability}

We start by studying how the sizes of an ensemble's base models
influence its interpretability. In this setting, we take the size of the
largest base model in an ensemble as our parameter $k$. We prove that
with this parameterization, \emph{all} of the aforementioned
explainability queries become intractable already for a \emph{constant} value of $k$.

\begin{proposition}
    \label{paramaterized_model_size}
    An ensemble consisting of either linear models, decision trees or neural networks, parameterized by the maximal base-model size is \begin{inparaenum}[(i)]
\item para-coNP Complete with respect to CSR, \item para-NP-Complete with respect to MCR, \item para-$\Sigma^P_2$-Complete with respect to MSR, \item para-$\#P$-Complete with respect to CC, and \item para-$\#P$-Hard with respect to SHAP. 
\end{inparaenum}
\end{proposition}

\emph{Proof Sketch.} The proof appears in Appendix~\ref{linear_model_para_hardness_results_appendix_section}. To prove para-NP/para-coNP hardness for MCR and CSR, we reduce from the NP-Complete Subset-Sum Problem to the problem of solving CSR/MCR for ensembles consisting of only two Perceptrons. To establish para-$\Sigma^P_2$-Hardness for MSR, we employ a more intricate reduction from the lesser-known Generalized Subset-Sum Problem (GSSP)~\cite{schaefer2002completeness, berman2002complexity}, a $\Sigma^P_2$-Complete problem. This reduction demonstrates that solving MSR in an ensemble of only five perceptrons is already $\Sigma^P_2$-Hard.

The results for majority-voting decision tree ensembles under the CSR and MCR queries were shown in~\citep{ordyniak2024explaining}, and we present here the remaining results. The proof of Proposition~\ref{paramaterized_model_size} appears in Appendix~\ref{paramaterized_model_size_appendix_section} and provides
a \emph{negative} outcome regarding the interpretability of
ensembles. The proposition implies that the uninterpretability of
ensembles is not a result of the sizes of the participating base
models, but rather of the aggregation process itself, such as the majority vote in voting ensembles. This implies that even if we reduce the size of our corresponding models to a constant size, the ensemble still remains intractable to interpret, and hence uninterpretable from a complexity perspective. This result applies to all of our base-model types, ensembles, and explanation forms.

\subsection{Impact of Number of Base-Models on Ensemble Interpretability}

A more intricate dynamic emerges when we take the \emph{number of base
  models} that participate in the ensemble as our parameter
$k$. Table~\ref{table:paramterizedresults} shows our parameterized
complexity results under this setting. The results for neural networks are
straightforward since they are closed under ensemble
construction. Decision trees and linear models however reveal an interesting
trend --- while linear models lose their tractability with a constant number of models in the ensemble, tree ensembles tend to remain XP-tractable, meaning they can be solved in polynomial time when the number of base models is fixed. The following subsections explore these findings in more detail.

\begin{table*}[h]
	\centering
	\def\arraystretch{1.31}%
	\setlength{\tabcolsep}{0.95em} 
	\caption{Parameterized complexity classes for explainability
          queries of ensemble models, parametrized by the number of
          models participating in the ensemble, $k$.}
          \scalebox{0.87}{
	\begin{tabular}{lccccccc}
		\\
		\toprule
		\multirow{3}{*}{} &
		\multicolumn{1}{c}{\textbf{Decision Trees}} &
		\multicolumn{1}{c}{\textbf{Linear Models}} &
      	\multicolumn{1}{c}{\textbf{Neural Networks}} \\
		\midrule
		\textbf{Check Sufficient Reason (CSR)} & coW[1]-C & para-coNP-C (\textbf{k=2})   & para-coNP-C (k=1)  \\
		\textbf{Minimum Contrastive Reason (MCR)} & W[1]-H, in W[P] & para-NP-C (\textbf{k=2}) & para-NP-C (k=1) \\
  		\textbf{Minimum Sufficient Reason (MSR)} & para-NP-H (k=1), in XNP & para-$\Sigma^P_2$-C (\textbf{k=5})  &  para-$\Sigma^P_2$-C (k=1) \\
		\textbf{Count Completions (CC)} & \#W[1]-C & para-\#P-C (k=1) & para-\#P-C (k=1)  \\
  		\textbf{Shapley Additive Explanations (SHAP)} & \#W[1]-H, in XP & para-\#P-C (k=1) & para-\#P-C (k=1)  \\
		\bottomrule
	\end{tabular}}
	\label{table:paramterizedresults}
\end{table*}

\subsubsection{The Number of Linear Models in an Ensemble}
We find that ensembles of linear models become hard to interpret with only a \emph{constant} number of base-models. Interestingly, we demonstrate that this property holds quite generally, as it applies across all the explanation types we analyzed, as stated in the following proposition, with the proof provided in Appendix~\ref{linear_model_para_hardness_results_appendix_section}:

\begin{proposition}
\label{k_ensemble_perceptrons_hardness_paper}
    $k$-ensembles of linear models are
\begin{inparaenum}[(i)]
\item para-coNP Complete with respect to CSR; \item para-NP-Complete with respect to MCR; \item para-$\Sigma^P_2$-Complete with respect to MSR; \item para-$\#P$-Complete with respect to CC, and \item para-$\#P$-Hard with respect to SHAP. 
\end{inparaenum}
\end{proposition}

\subsubsection{The Number of Decision Trees in an Ensemble}

Unlike ensembles composed of linear models, ensembles of decision trees (including widely used models like random forests and XGBoost) yield more optimistic complexity results. Specifically, explanations become computationally tractable when the number of trees in the ensemble is reduced. The precise degree of tractability, however, varies depending on the type of explanation:

\textbf{XP tractable explanations.} We demonstrate that four of the five queries analyzed (CSR, MCR, CC, and SHAP) fall within XP, with some also belonging to lower complexity classes in the W-hierarchy (the full proofs are relegated in Appendix~\ref{xp_proofs_appendix}).

\begin{proposition}
\label{main_paper_fbdd_ensemble_paramterized_results}
For ensembles of $k$-decision trees, \begin{inparaenum}[(i)] \item the CSR query is coW[1]-Complete; \item the MCR query is W[1]-Hard and in W[P]; \item the CC query is $\#$W[1]-Complete; and \item the SHAP query is $\#W[1]$-Hard and in XP.
\end{inparaenum}
\end{proposition}

\emph{Proof Sketch.} Membership for the CSR query is established through a many-to-one FPT reduction to the complementary version of the $k$-Clique problem, while hardness was shown by~\citep{ordyniak2024explaining}. The CC query extends these results, as the counting versions of these tasks are $\#$W[1]-Complete. $\#W[1]$-Hardness for SHAP can be inferred from the complexity of CC, given that the tractability of SHAP is~linked to model counting~\cite{van2022tractability}. For membership, a more detailed proof places SHAP within XP, which is made possible by the assumption of feature independence. Specifically, when the distribution is uniform, SHAP is proven to be $\#W[1]$-Complete (see Lemma~\ref{lemma_of_uniform_shap}). Lastly, while hardness of the MCR query was demonstrated by~\citep{ordyniak2024explaining}, its membership is established by reducing it to the weighted circuit satisfiability (WCS) problem for circuits with arbitrary depth.


Intuitively, this result indicates that these types of explanations can be computed in polynomial time when the number of trees in the ensemble is fixed. The levels of tractability, corresponding to different explanation types, align with various classes within the W-hierarchy.

\textbf{XNP tractable explanations.} The only explainability query that is not in XP for tree ensembles is the minimum sufficient reason (MSR) query. This makes sense since this problem is known to be NP-Hard only for a \emph{single} decision tree~\cite{BaMoPeSu20}, and hence it is para-NP-Hard for an ensemble of $k$ trees. However, we can show a similar behavior, by proving its membership in XNP, the variant of XP for the second order of the polynomial hierarchy~\cite{flum2004parameterized}. This is illustrated in the following proposition, with its proof provided in Appendix~\ref{msr_k_ensemble_fbdds_appendix_sec}.

\begin{proposition}
\label{msr_prop_main_paper}
    The MSR query for a $k$-ensemble of decision trees is para-NP-Hard and in XNP.
\end{proposition}

\emph{Proof Sketch.} To prove membership, we devise an algorithm that initially computes all minimal contrastive reasons during a preprocessing phase in $O(|\mathcal{X}|^k)$ time. In the second phase, the algorithm leverages the Minimal-Hitting-Set (MHS) duality between sufficient and contrastive reasons~\cite{ignatiev2020contrastive}, allowing the algorithm to non-deterministically identify the MHS among all minimal contrastive reasons.

Intuitively, while this query is NP-Hard when $k$ is constant, its membership in XNP suggests it remains significantly more tractable compared to para-$\Sigma^P_2$-Hard problems (as encountered with the MSR query for ensembles of linear models). When $k$ is fixed, these problems can be addressed using NP oracles (e.g., SAT solvers~\citep{moskewicz2001chaff}), whereas para-$\Sigma^P_2$-Hard problems demand a worst-case exponential number of queries to such solvers.

Although the MSR query is not in XP, a \emph{relaxed} version is. This version seeks a \emph{subset-minimal} (or ``locally minimal'') sufficient reason, i.e., a subset $S\subseteq [n]$ that is a sufficient reason, where removing any $i$ from $S$ makes it no longer sufficient. This result was previously shown in~\citep{ordyniak2024explaining}; in Appendix~\ref{xp_subset_minimal_proof_sec}, we discuss its connection to our other findings.


\begin{proposition}
\label{subset_minimal_proof_main}
    Obtaining a subset-minimal sufficient reason for a $k$-ensemble of decision trees is in XP.
\end{proposition}



\textbf{FPT tractable explanations.} Although the previous XP-tractable explanations are \emph{relatively} more tractable for tree ensembles (compared to ensembles of linear models) and solvable in polynomial time with a fixed number of trees, our results show they are either W[1] or coW[1]-hard. Thus, assuming FPT $\subsetneq$ W[1], no FPT algorithms exist to solve these queries in $|\mathcal{X}|^{O(1)} \cdot g(k)$ time~\cite{downey2012parameterized}. Consequently, they are not primarily driven by the parameter $k$ and grow more challenging as tree size increases.


However, if we assume that the number of leaf nodes $m$ in each tree is bounded by a constant (even if the total size $|f_i|$ of each tree is arbitrarily large), it is possible to develop FPT algorithms to solve these queries. This follows from the fact that all the aforementioned algorithms have a runtime of $O(m^k)$. This leads to the following conclusion, which can also be inferred from~\citep{ordyniak2024explaining} for majority-voting tree ensembles under the CSR and MCR queries and we extend this result to all remaining settings:

\begin{proposition}
\label{bounded_leaves_proposition}
If the maximal number of leaves in each tree $m$ is constant, there exist FPT algorithms that solve the CSR, MCR, CC, and SHAP queries for $k$-ensembles of decision trees (even if the size $|f_i|$ of each base-model, and hence the size of the ensemble $|f|$, is arbitrarily large).
\end{proposition}

\subsubsection{Tree Ensembles vs. Linear Model Ensembles} 

The previous
subsections showed that decision tree and linear model ensembles
exhibit very distinct behaviors when parameterized by the number of
base models $k$, giving rise to the following corollary: 

\begin{theorem}
$k$-ensemble decision trees are strictly more c-interpretable than $k$-ensemble linear models with respect to CSR, MSR, MCR, CC, and SHAP. 
\end{theorem}

From a practical perspective, if a practitioner seeks to \emph{simplify} an ensemble by reducing the number of base models, this simplification enhances computational interpretability for decision tree ensembles. However, this is not the case for ensembles composed of linear models, where interpretability becomes intractable even with just two linear models. Notably, these complexity challenges also extend to heterogeneous ensembles. In other words, any ensemble that includes a mix of model types and only two linear models is already intractable to interpret.




Finally, Proposition~\ref{bounded_leaves_proposition} shows that algorithms for CSR, MCR, CC, and SHAP on decision tree ensembles run not only in XP but in FPT time with a fixed number of leaves per tree, even for arbitrarily large input spaces and tree sizes. In contrast, ensembles of linear models remain computationally intractable unless the model size --- and thus the input space size --- is constrained, even with just two models.






\section{Practical Implications}

While our work is theoretical, it offers key insights for practitioners focused on ensemble interpretability. Specifically, we show that generating explanations for ensembles is substantially intractable, with exponential gaps compared to single base-models. This remains true even for ensembles with constant-sized base models, indicating that using ensembles with small base-models alone does not improve interpretability from a computational standpoint. A key practical takeaway, however, is that using ensembles with fewer but deeper trees is computationally more favorable than many shallow ones. This observation can guide tree architecture choices when interpretability is a priority. Conversely, incorporating linear models --- even with only a few models --- can drastically increase complexity. This underscores a substantial comparative advantage for decision-tree-based ensembles over those using linear models.


More concretely, our tractability results yield \emph{polynomial-time algorithms} for computing explanations. While some cases are W[1]-, coW[1]-, or \#W[1]-hard, bounding the number of leaves per tree makes them FPT-tractable, enabling efficient computation even for arbitrarily large ensembles. Many of these tasks also support parallelization, boosting performance. Finally, since our problems fall within well-studied classes like W[1], they can be reduced to canonical problems like $k$-Clique, for which effective heuristics already exist~\citep{vassilevska2009efficient, chang2019efficient, chen2019robustness}, opening new directions for practical applications.


\section{Related Work}

Our work relates to the field of \emph{formal XAI}, which focuses on
explanations with mathematical guarantees
\cite{marques2020explaining}. Previous studies have explored the computational complexity of generating such explanations for various ML models \cite{BaMoPeSu20, bassanlocal, adolfi2024computational}. Here, we focus on the complexity of \emph{ensemble} models and their interpretability. Prior work has examined the complexity of interpretations for specific ensembles (e.g., random forests) \cite{izza2021efficient, audemard2022trading, audemard2023computing, huang2024updates} and used parameterized complexity to distinguish between shallow and deep neural networks \cite{BaMoPeSu20} or parameterize explanations by size \cite{ordyniak2023parameterized}. In a more recent and independent study, the notable work by~\cite{ordyniak2024explaining} examines the complexity of various explainability queries --- primarily local and global abductive and contrastive explanations, which correspond to our CSR, MSR, and MCR queries --- under different parameterizations, including those related to majority-voting ensembles of models such as decision trees, lists, and sets. We provide a more technical and comprehensive discussion of all relevant prior computational complexity results, including this work and others, in Appendix~\ref{appendix:extended_related_work}.



Another line of work focuses on obtaining explanations with formal
guarantees on tree ensembles by encoding them as propositional logic
formulas and then solving these queries with Boolean satisfiability (SAT)
\cite{izzaexplaining}, Maximum satisfiability (MaxSAT), \cite{ignatiev2022using}, Mixed
integer linear programming (MILP) \cite{parmentier2021optimal,
  chen2019robustness} or satisfiability modulo (SMT)
\cite{ignatiev2019validating} solvers. 


Finally, in our work, we used terms that have sometimes appeared in the literature under different names. For example, ``sufficient reasons'' are also called abductive explanations \cite{ignatiev2019relating}. Subset minimal sufficient reasons are related, though not identical, to prime implicants in Boolean formulas \cite{darwiche2002knowledge}. Similarly, the CC query aligns with the concept of a $\delta$-relevant set \cite{waldchen2021computational, izza2021efficient}, which identifies whether the completion count exceeds a threshold $\delta$.

\section{Limitations and Future Work}

Similarly to other studies on the computational complexity of obtaining explanations, our work is limited to specific explanation forms and base-model types. However, we believe it still offers a broad overview of various formats and settings, covering diverse facets of ensemble interpretability and laying the groundwork for exploring the complexity of additional forms in future work. Moreover, while most of our findings extend from classification to regression, some require further investigation, as discussed in Appendix~\ref{framework_extensions_section}, along with other potential extensions and open problems.

\section{Conclusion}
We introduce a complexity-theoretic framework for evaluating the interpretability of ensemble models. Our study encompasses a wide range of popular explanation forms, base-model types, and varying structural parameters, such as the number and size of base models. We believe our findings offer a significantly more rigorous understanding of some of the fundamental factors shaping the interpretability limitations of ensembles. While some of our results support widely held assumptions in the ML community, others are unexpected and surprising --- for instance, the significantly differing impacts of base-model sizes, numbers, and types across various contexts. These insights provide a novel perspective on ensemble interpretability, informing both the practical development of explanation algorithms (the tractable cases) and a deeper understanding of scenarios where explanations may be infeasible to obtain (the intractable cases). Overall, we believe that our work underscores the importance of computational complexity in advancing our understanding of interpretability in ML.


\section*{Impact Statement}
We acknowledge the potential social implications of interpretability. Our work provides a deeper understanding of the circumstances under which explanations can or cannot be obtained efficiently. However, as our work is predominantly theoretical, we believe it does not have any immediate or direct social ramifications.

\section*{Acknowledgments}

This work was partially funded by the European Union
(ERC, VeriDeL, 101112713). Views and opinions expressed
are however those of the author(s) only and do not neces-
sarily reflect those of the European Union or the European
Research Council Executive Agency. Neither the European
Union nor the granting authority can be held responsible
for them. This research was additionally supported by a grant from the Israeli Science Foundation
(grant number 558/24). The work of Amir was further supported by a
scholarship from the Clore Israel Foundation.


\bibliography{example_paper}
\bibliographystyle{icml2025}


\newpage
\appendix
\onecolumn

\newcounter{definition}
\newcounter{proposition}
\newcounter{lemma}

\setcounter{definition}{0}
\setcounter{proposition}{0}
\setcounter{theorem}{0}
\setcounter{lemma}{0}
\begin{center}\begin{huge} Appendix\end{huge}\end{center}    
\noindent{The appendix contains formalizations and proofs that were mentioned throughout the paper:}

\newlist{MyIndentedList}{itemize}{4}
\setlist[MyIndentedList,1]{%
	label={},
	noitemsep,
	leftmargin=0pt,
}

\begin{MyIndentedList}
	\item \textbf{Appendix~\ref{appendix:extended_related_work}} contains an extended discussion of related work.
	\item \textbf{Appendix~\ref{appendix:model_types}} contains formalizations of base-model types and ensembles.
  	\item \textbf{Appendix~\ref{parametrieed_background}} contains background on parameterized complexity.
  	\item \textbf{Appendix~\ref{Additional_query_formalizations}} contains additional formalizations concerning explainability queries.
   	\item \textbf{Appendix~\ref{framework_extensions_section}} contains information regarding possible expansions of our framework to \begin{inparaenum}[(i)]
   	    \item discrete and continuous input/output domains, \item regression tasks, and \item heterogeneous ensembles.
   	\end{inparaenum}
    \item \textbf{Appendix~\ref{non_paramaterized_appendix_section}} contains the proof of Proposition~\ref{main_result_non_paramterized_main_text}.
    \item \textbf{Appendix~\ref{pseudo_plynomial_theorem_appendix_section}} contains the proof of Proposition~\ref{CC_and_SHAP_proposition}.
    \item \textbf{Appendix~\ref{no_gap_in_complex_models_appendix_sec}} contains the proof of Proposition~\ref{no_gap_original_paper_prop}.
    \item \textbf{Appendix~\ref{paramaterized_model_size_appendix_section}} contains the proof of Proposition~\ref{paramaterized_model_size}. 
\item \textbf{Appendix~\ref{linear_model_para_hardness_results_appendix_section}} contains the proof of Proposition~\ref{k_ensemble_perceptrons_hardness_paper}.
\item \textbf{Appendix~\ref{xp_proofs_appendix_sec}} contains the proof of Proposition~\ref{main_paper_fbdd_ensemble_paramterized_results}.
\item \textbf{Appendix~\ref{msr_k_ensemble_fbdds_appendix_sec}} contains the proof of Proposition~\ref{msr_prop_main_paper}.
\item \textbf{Appendix~\ref{xp_subset_minimal_proof_sec}} contains the proof of Proposition~\ref{subset_minimal_proof_main}. 
	
\end{MyIndentedList}

\section{Extended Related Work}
\label{appendix:extended_related_work}

This section provides a more detailed discussion of related work and key complexity results explored in previous research.

\textbf{The Complexity of Computing Explanations.} Our work builds on a growing body of research analyzing the computational complexity of generating various types of explanations with different guarantees across a wide range of ML models~\citep{BaMoPeSu20, amir2024hard, bhattacharjee2024auditing, adolfi2024computational, blanc2022query, huang2023feature, ArBaBaPeSu21, cooper2023tractability}. Prior studies have examined the complexity of computing explanations based on sufficiency~\citep{BaMoPeSu20, bassanlocal}, contrastiveness~\citep{blanc2022query, cooper2023tractability, BaMoPeSu20}, Shapley values~\citep{marzouk2025on, marzouk2024tractability, van2022tractability, arenas2023complexity}, and probabilistic reasoning~\citep{izza2023computing, blanc2021provably, waldchen2021computational}. These analyses span a variety of model classes --- from inherently interpretable models like linear classifiers~\citep{subercaseaux2024probabilistic}, monotonic models~\citep{marques2021explanations}, KNNs~\citep{barcelo2025explaining}, and decision trees~\citep{bounia2023approximating, arenas2022computing, huang2021efficiently} --- to more complex black-box models such as neural networks~\citep{adolfi2024computational, bassanlocal, BaMoPeSu20}, where the computational challenges are typically greater.


\textbf{The Complexity of Computing Explanations for Ensembles.} Most closely related to our work are studies that analyze the complexity of generating explanations for ensemble models. Notably,~\citep{izzaexplaining} show that computing a subset-minimal sufficient reason for a random forest is $D^P$-complete, a result extended by~\citep{audemard2023computing} to weighted voting ensembles. The significant work of~\citet{audemard2022trading} establishes complexity results for sufficient and contrastive reasons in majority-voting tree ensembles --- results we expand upon in the relevant sections. More recently, the highly notable work of~\citep{ordyniak2024explaining} provide a thorough parameterized complexity analysis for majority-voting ensembles of decision trees, decision lists, and sets, considering both explanation-size parameters and structural parameters of the models. We elaborate on the connection to their results throughout the paper and in the appendix. Lastly, both~\citep{huang2024updates} and~\citep{marzouk2025on} analyze the complexity of computing SHAP values on different models, which include tree ensembles; our work extends these findings by exploring additional ensemble settings and introducing new parameterized results.

\textbf{Formal XAI.} More broadly, our work lies within the subfield of \emph{formal XAI}~\citep{marques2022delivering}, which seeks to provide explanations for ML models with formal guarantees~\citep{ignatiev2020towards, bassan2023towards, darwiche2020reasons, darwiche2022computation, ignatiev2019abduction, audemard2022preferred, bassanfast}. Such explanations are typically obtained using formal reasoning tools like SMT solvers~\citep{barrett2018satisfiability} (e.g., for tree ensembles~\citep{audemard2022trading}) or neural network verifiers~\citep{katz2017reluplex, wu2024marabou, wang2021beta} (e.g., for neural networks~\citep{izza2024distance, bassan2023formally}), or by performing other manipulations such as relaxing the sufficiency definitions~\citep{jin2025probabilistic, izza2023computing, wang2021probabilistic, chockler2024causal, jin2025probabilistic}, applying smoothing~\citep{xue2023stability}, or using self-explaining methods~\citep{bassan2025explain, alvarez2018towards, bassan2025self}. A central concern in the area of Formal XAI is the computational complexity of generating these explanations~\citep{BaMoPeSu20,waldchen2021computational, cooper2023tractability, bassanlocal, blanc2021provably, amir2024hard, adolfi2024computational, barcelo2025explaining, calautti2025complexity, ordyniak2023parameterized}.


\section{Ensembles and Base-Model Types}

\label{appendix:model_types}
In this appendix, we describe the various ensembles that are incorporated in our work and the corresponding base-model types that consist in these models.

\subsection{Ensemble Formalization}

Numerous well-known ensemble techniques exist; however, our research is geared towards \emph{post-hoc} interpretation, thus we emphasize the \emph{inference} phase rather than the training of these ensembles. Our analysis is focused on ensemble families that utilize either \emph{majority voting} or \emph{weighted-voting} methods during inference. This includes \emph{bagging} ensembles such as random forests, which implement majority-voting-based inference, and \emph{boosting} ensembles like XGBoost, Gradient Boosting, and Adaboost, which employ weighted voting for inference. Moreover, we will examine how all of our results hold to either \emph{hard-voting} or \emph{soft-voting} inference as well as the potential to apply our findings to alternative inference techniques such as \emph{weighted averaging}, or \emph{meta-model decision} inference. Such inference techniques are commonly found in other ensemble types like \emph{stacking} ensembles, or those utilized in regression tasks.
    
\mysubsection{Majority Voting Inference.}

In \emph{majority voting} inference, the condition $f(\x)=1$ is satisfied if and only if there are at least $\ceil{\frac{k}{2}}$ base-models within an ensemble $f$ where $f_i(\x)=1$. Put simply, the decision for $f(\x)$ is determined by the majority consensus of the models involved in $f$. Formally, for any $\x\in\mathbb{F}$ we define $f$ as follows:

 \begin{equation}
f(\x):=\begin{cases}
    1 \quad if \ \ | \{ \ i \ | \ f_i(\x)=1 \ \}| \ \geq \ \ceil{\frac{k}{2}}\\
    0 \quad otherwise
\end{cases}
\end{equation}


\mysubsection{Weighted Voting Inference.}

For \emph{weighted voting} inference, we consider a weight $\phi_i\in\mathbb{Q}$ that describes the importance of each model participating in the ensemble. Considering this is a binary classification, we can define the prediction as being determined by the sign of the total aggregation of all weights. Formally, for any $\x\in\mathbb{F}$ we define $f$ as follows:
\begin{equation}
\label{weighted_voting_formalization}
f(\x):=step(\sum_{1\leq i\leq n} \phi_i\cdot f_i(\x))
\end{equation}

where $step(\z) = 1 \iff \z \geq 0$.

\textbf{Weighted voting is harder than majority voting}: When examining the explainability query $Q$, our analysis aims to evaluate the complexity classes of two families of ensembles: majority voting ensembles, denoted as $\mathcal{C}_{\mathcal{M}}$, and weighted voting ensembles, denoted as $\mathcal{C}_{\mathcal{W}}$. It is straightforward to demonstrate that the following relationship is true:

\begin{lemma}
    Let $\mathcal{C}$ denote a class of models, $\mathcal{C}_{\mathcal{M}}$ the class of majority voting ensembles of models from $\mathcal{C}$, and $\mathcal{C}_{\mathcal{W}}$, the class of weighted voting ensembles of models from $\mathcal{C}$. Then for any explainability query $Q$ it holds that $Q(\mathcal{C}_\mathcal{M})\leq_P Q(\mathcal{C}_\mathcal{W})$. 
\end{lemma}

\emph{Proof.} The lemma holds by a simple reduction that starts with a majority voting ensemble $f$ and constructs a weighted voting ensemble $f'$ where each weight is assigned an equal attribution. In other words: $\phi'_i = \frac{1}{n}$ for all $i\in[n]$.

$\qedsymbol{}$

The previous statement holds technical significance as it simplifies the process of establishing complexity class completeness results for both $\mathcal{C}_{\mathcal{M}}$ and $\mathcal{C}_{\mathcal{W}}$:

\begin{observation}
    For proving that the complexity of solving both $Q(\mathcal{C}_{\mathcal{M}})$ and  $Q(\mathcal{C}_{\mathcal{W}})$ are complete for some complexity class $\mathcal{K}$ (closed under polynomial reductions), it suffices to prove membership for $Q(\mathcal{C}_{\mathcal{W}})$ and hardness for $Q(\mathcal{C}_{\mathcal{M}})$.
\end{observation}

From this point forward, whenever we mention the computational complexity of an ensemble of models in our text, it applies to both majority voting ensembles and weighted voting ensembles, as we prove the completeness of complexity classes for both families. We highlight this differentiation in our proofs, which are applicable to both types of ensembles.

\mysubsection{Extension to \emph{Soft} Voting.} In contrast to \emph{hard} voting that is common in the binary classification setting, within probabilistic classification soft voting can also be implemented. In this case, each model $f_i$ in the ensemble outputs some given probability value i.e, $f_i:\mathbb{F}\to [0,1]$. Then, in the case of \emph{majority soft voting}, the inference is computed by:

\begin{equation}
\label{majority_soft_voting_equation}
f(\x):=step(\sum_{1\leq i\leq n} \frac{f_i(\x)}{n})
\end{equation}

whereas in \emph{weighted soft voting} the inference is computed by incorporating equation~\ref{weighted_voting_formalization} for each $f_i:\mathbb{F}\to [0,1]$.

We start by defining a property that will be used in the next lemma. We say that a class $\mathcal{C}$ of models $f:\mathbb{F}\to [0,1]$ is \emph{scalar multiplicative} if given some constant $\lambda\in\mathbb{R}$ and for all $f\in\mathcal{C}$ we can construct, in polynomial time a model $f'\in\mathcal{C}$ for which $f'=\lambda f$.

\begin{lemma}
\label{soft_voting_lemma}
    Let $\mathcal{C}$ denote a class of models, $\mathcal{C}_{\mathcal{W}}$ the class of weighted (hard) voting ensembles of models from $\mathcal{C}$, $\mathcal{C}_{\mathcal{SW}}$, the class of  (soft) weighted voting ensembles of models from $\mathcal{C}$, and $\mathcal{C}_{\mathcal{SM}}$, the class of  (soft) majority voting ensembles of models from $\mathcal{C}$. Then for any explainability query $Q$ it holds that $Q(\mathcal{C}_\mathcal{W})=_P Q(\mathcal{C}_\mathcal{SW})=_PQ(\mathcal{C}_\mathcal{SM})$. The only restriction is for the condition $Q(\mathcal{C}_\mathcal{W})\geq_P Q(\mathcal{C}_\mathcal{SM})$ and it is that $\mathcal{C}_\mathcal{W}$ is scalar multiplicative.
\end{lemma}

\emph{Proof.} Given a soft majority voting ensemble $f$, we can construct a weighted hard (voting) ensemble $f'$ for which each weight $\phi'_i(\x):=\frac{f_i(\x)}{n}$. For the other direction, given a weighted hard (voting) ensemble $f$, and assuming that $\mathcal{C}_\mathcal{W}$ is scalar multiplicative, we can construct a soft majority voting ensemble $f'$. We do this by, for every $i\in[n]$, constructing $f'_i(\x)=\phi_i \cdot f_i(\x)$ (since $\mathcal{C}_\mathcal{W}$ being scalar multiplicative). Overall, from these two reductions, we get that $Q(\mathcal{C}_\mathcal{W})=_P Q(\mathcal{C}_\mathcal{SM})$.

For the second part of the claim --- we start with a weighted hard voting ensemble $f$ and construct a weighted soft voting ensemble $f'$ by assigning each weight $\phi'_i(\x):=\frac{\phi_i(\x)}{n}$. For the other direction, given a weighted soft voting ensemble $f$ we construct a weighted hard voting ensemble $f'$ by setting $\phi'_i(\x):=\phi_i(\x)\cdot f_i(\x)$. Overall, this implies that $Q(\mathcal{C}_\mathcal{W})=_P Q(\mathcal{C}_\mathcal{SW})$.

$\qedsymbol{}$

Lemma~\ref{soft_voting_lemma} establishes that our proofs (including those for membership and hardness) apply to \emph{soft-voting} ensembles, both for majority and weighted voting scenarios. This is because our proofs are conducted within the framework of \emph{weighted} (hard voting ensembles), and all the complexity classes we consider are closed under polynomial reductions.

\textbf{(Weighted) Averaging.} In the \emph{regression} setting, another common inference technique involves a \emph{weighted averaging} of all base-model predictions. Formally, given a set of $k$ regression base-models $f_i:\mathbb{F}\to\mathbb{R}$, then we define $f$ by incorporating equation~\ref{majority_soft_voting_equation} over each $f_i$. In essence, this is the same formalization as that of majority soft voting, and hence Lemma~\ref{soft_voting_lemma} holds for this family of inference models as well. However, when shifting our focus to regression, we must also consider different formalizations of some of the query forms discussed in our paper, such as the definition of a sufficient reason. We discuss these specific adjustments for the regression setting under Appendix~\ref{framework_extensions_section}. 

\mysubsection{Meta learner decision.} Another common ensemble inference method often used in \emph{stacking} ensembles involves employing a \emph{meta-model} to aggregate the outputs of the $k$ base models. In our particular scenario, this involves a model $g:\{0,1\}^k\to\{0,1\}$, which is trained to classify the outputs from each base model $f_i$ within a specified domain. It is important to note that if $g$ can function as a majority voting system among the $k$ models --- a capability all analyzed model types possess, including neural networks, linear classifiers, and decision trees -—-  then all the \emph{hardness} findings discussed in this paper automatically apply to this setup as well. For instance, a \emph{stacking} ensemble comprising a constant number of linear base-models remains intractable to interpret, as demonstrated in Proposition~\ref{k_ensemble_perceptrons_hardness_paper}. However, the examination of \emph{membership} results that were presented in this paper will vary depending on the type of model used for the meta-model $g$.

\subsection{Base-Model Types}

In this subsection, we formalize the three base-model types that were analyzed throughout the paper: \begin{inparaenum}[(i)] \item (axis-aligned) decision trees, \item linear classifiers, and \item neural networks with ReLU activations.
\end{inparaenum} 

\mysubsection{Decision Trees.} We regard a decision tree is an acyclic-directed graph and serves as a graphical model for the function $f$. This graph embodies the given Boolean function in the following manner: 
\begin{inparaenum}[(i)]
	\item Each internal node $v$ is associated with a unique binary input feature from the set $\{1,\ldots,n\}$;
	\item Every internal node $v$ has precisely two outgoing edges, corresponding to the values $\{0,1\}$ which are assigned to $v$;
	\item In the decision tree, each variable is encountered no more than once on any given path $\alpha$;
 	\item Each leaf node is labeled either \true{} or \false{}.
\end{inparaenum}

Thus, assigning a value to the inputs $\x\in\mathbb{F}$ uniquely determines a specific path $\alpha$ from the root to a leaf in the decision tree. The function $f(\x)$ is assigned a $1$ if the terminal node leaf is labeled \true{}, and $0$ if it is labeled \false{}. The size of the decision tree, denoted as $\vert f \vert$, is measured by the total number of edges in its graph. We additionally assume that the decision tree permits different varying orderings of the input variables $\{1,\ldots,n \}$ across any two distinct paths, $\alpha$ and $\alpha'$. This ensures that no two nodes along any single path $\alpha$ share the same label.

\mysubsection{Neural Networks.} We focus on neural networks with ReLU activations. It is straightforward to show that any neural network using ReLU activations can be straightforwardly transformed into a fully connected network by assigning ``missing'' connections a weight and bias of $0$ for any non-connected neurons. Following standard conventions~\citep{BaMoPeSu20, bassanlocal, adolfi2024computational}, we assume the network is fully connected. In other words, our analysis pertains to multi-layer perceptrons (MLPs). Formally, an MLP, denoted by $f$, consists of $t-1$ \emph{hidden layers} ($g^{j}$ where $j$ ranges from $1$ to $t-1$) and a single output layer ($g^{t}$). The layers are defined recursively --- each layer $g^{(j)}$ is computed by applying the activation function $\sigma^{(j)}$ to the linear combination of the outputs from the previous layer $g^{(j-1)}$, the corresponding weight matrix $W^{(j)}$, and the bias vector $b^{(j)}$. This is represented as $g^{(j)} := \sigma^{(j)}(g^{(j-1)}W^{(j)} + b^{(j)})$ for each $j$ in $\{1,\ldots,t\}$. The model includes $t$ weight matrices ($W^{(1)},\ldots,W^{(t)}$), $t$ bias vectors ($b^{(1)},\ldots,b^{(t)}$), and $t$ activation functions ($\sigma^{(1)},\ldots,\sigma^{(t)}$).

In the described MLP, the function $f$ is defined to output $f \coloneqq g^{(t)}$. The initial input layer $g^{(0)}$ is denoted by $\x \in {\{0,1\}}^n$, which serves as the model's input. The dimensions of the biases and weight matrices are specified by the sequence of positive integers $\{d_{0}, \ldots, d_{t}\}$. We specifically focus on weights and biases that are rational numbers, represented as $W^{(j)}\in \mathbb{Q}^{d_{j-1}\times d_{j}}$ and $b^{(j)}\in \mathbb{Q}^{d_{j}}$, which are parameters that are optimized during training. Given that the model is a binary classifier for indices $\{1, \ldots, n\}$, it follows that $d_{0} = n$ and $d_{t} = 1$. The primary activation function $\sigma^{(i)}$ that we focus on is the \emph{ReLU} activation function, defined as $reLU(x) = \max(0, x)$, except for the output layer, where a \emph{sigmoid} function is typically used for the classification. Since our focus is only on the \emph{post-hoc} interpretation of the corresponding model, we will equivalently assume the existence of a step function for the final layer activation, where we denote $step(\z) = 1 \iff \z \geq 0$.

\mysubsection{Linear Classifiers.} A linear classifier is essentially equivalent to a single-layer MLP classifier, which corresponds to a Perceptron model. To emphasize this equivalence --- also important for establishing the technical significance of hardness proofs for linear classifiers that extend to neural network classifiers --- we will refer to them as \emph{perceptrons} throughout this work. This terminology is consistent with prior research on the subject~\citep{BaMoPeSu20, bassanlocal}. Formally, a perceptron represents a single-layer MLP or in other words $t=1$. It is defined by the function $f(\x) = step((\mathbf{w}\cdot\x)+b)$ with $b$ belonging to the set of rational numbers, and $\mathbf{w}$ being a matrix in $\mathbb{Q}^{n\times d_{1}}$. Consequently, for the perceptron function $f$ it holds without loss of generality that $f(\x)=1$ if and only if $(\mathbf{w}\cdot\x)+b \geq 0$.



\section{Parameterized Complexity Background}
\label{parametrieed_background}

In parameterized complexity, we deal with \emph{parameterized problems} $L\subseteq \Sigma^*\times\mathbb{N}$ where $\Sigma$ is some finite alphabet. The elements of the paramaterized problems are hence two-dimensional instances denoted as $\langle \mathcal{X},k\rangle$ where $\mathcal{X}$ represents the original encoding and $k$ is the \emph{parameter}.

\subsection{Parameterized Reductions}

\textbf{FPT Reductions.} The parameterized complexity classes that we will discuss here are closed under a specific kind of reductions, known as fixed-parameter tractable (FPT) reductions. A given mapping $\phi:\Sigma^*\times\mathbb{N}\to\Sigma^*\times\mathbb{N}$ between instances from a parameterized problem $P_1$ to another parameterized problem $P_2$ is an \emph{FPT reduction} iff it holds that: \begin{inparaenum}[(i)]
    \item $(\mathcal{X},k)$ is in $P_1$ if and only if $\phi(\mathcal{X},k)$ is in $P_2$;
    \item there exists a computable function $g$ for which $k'\leq g(k)$ when $k'$ is the parameter of $\phi(\mathcal{X},k)$; and \item $\phi(\mathcal{X},k)$ can be computed in $|\mathcal{X}|^{O(1)}\cdot g'(k)$ time for some computable function $g'$.
\end{inparaenum}

\textbf{FPT Parsimonious Reductions.} For the counting version of FPT reductions~\cite{flum2004parameterized}, given two paramterized counting problems $F:\Sigma^*\times\mathbb{N}\to\mathbb{N}$, and  $G:\Sigma^*\times\mathbb{N}\to\mathbb{N}$ we define an FPT parsimonious reduction from $F$ to $G$ as an algorithm that computes for any instance $\langle \mathcal{X},k\rangle$ of $F$ an instance $\langle \mathcal{Y},\ell\rangle$ of $G$ in time $g_1(k)\cdot |\mathcal{X}|^c$ such that $\ell\leq g_2(k)$ and $F(\mathcal{X},k)=G(\mathcal{Y},\ell)$, for some computable functions $g_1, g_2:\mathbb{N}\to\mathbb{N}$, and a constant $c\in\mathbb{N}$.

\subsection{Parameterized Complexity Classes.}

We now will formalize the parameterized complexity classes that are relevant for this work.

\textbf{FPT.} A problem is \emph{fixed parameter tractable} (\emph{FPT}) concerning $k$ iff there exists a $|\mathcal{X}|^{O(1)}\cdot g(k)$ time algorithm solving the problem for some computable function $g$.

\textbf{XP and the W-Hierarchy.} The class XP describes all problems that can be solved in $O(|\mathcal{X}|^{g(k)})$ time for some computable function $g$. This class additionally encompasses the $W$-hirerchy~\cite{downey2012parameterized}, which can be described using boolean circuits. We recall that a boolean circuit $C$ is represented as a rooted directed acyclic graph. Nodes with no incoming edges are referred to as input gates, and the singular node without any outgoing edges is the output gate. The internal nodes of the circuit are designated as OR, AND, or NOT gates. NOT gates are characterized by having exactly one incoming edge. AND and OR gates can have up to two incoming edges, termed small gates, or more than two, termed large gates. The \emph{depth} of a circuit is measured by the longest path of edges from any input node to the output node. The \emph{weft} of a circuit refers to the largest number of large gates on any path from an input node to the output node. An assignment for $C$ maps the input gates to binary values \{0,1\}. The \emph{hamming weight} of an assignment reflects the count of input gates assigned the value $1$. This assignment determines the output at each gate based on its specific function. A circuit is satisfied by an assignment if it results in the output gate producing a value of $1$.

We can now characterize the $W$-Hierarchy through reductions to the general \emph{Weighted Circuit Satisfiability} problem (\emph{WCS}), which is defined as follows:

\vspace{0.5em} 

\noindent\fbox{%
    \parbox{\columnwidth}{%
\mysubsection{Weighted Circuit Satisfiability (WCS[$C_{t,d}$])}:

\textbf{Input}: A boolean Circuit $C$ with weft at most $t$ and depth at most $d$, and an integer $k$, 

\textbf{Parameter}: $k$

\textbf{Output}: \yes{}, if $C$ has a satisfying assignment of Hamming weight exactly $k$
    }%
}

\vspace{0.5em} 

We then say that a problem $Q$ belongs to $W[t]$ if there is an FPT reduction from $Q$ to WCS[$C_{t,d}$], for some fixed constant $d$. If there exists an FPT reduction from $Q$ to WCS[$C_{t,d}$], where the constructed circuit $C$ is allowed to have an arbitrary weft $t$, then we say that $Q$ belongs to $W[P]$. It is widely believed that the following relation holds~\cite{downey2012parameterized}:

\begin{equation}
    \text{FPT} \subsetneq W[1]\subsetneq W[2]\subsetneq \ldots \subsetneq W[t] \subsetneq W[P] \subsetneq \text{XP}
\end{equation}

\textbf{XNP.} Our paper will also briefly discuss the \emph{XNP} complexity class~\cite{de2017parameterized}, which is a generalization of the XP class to the second order of the polynomial hierarchy. XNP describes the set of problems that can be solved by a non-deterministic algorithm in $O(|\mathcal{X}|^{g(k)})$ time for some computable function $g$. It is widely believed that XNP $\subsetneq$ para-$\Sigma^P_2$.

\textbf{The $\#W$-Hirerchy.} The $W$-Hierarchy can be extended to its equivalent counting based hiearchy, which is termed the $\#W$-hierarchy~\cite{flum2004parameterized}. Similarly to the $W$-herarchy, where we use the problem WCS[$C_{t,d}$], we now denote $\#$WCS[$C_{t,d}$] as the problem of counting the number of assignments of Hamming weight $k$ for a Boolean circuit $C$ with depth $d$ and weft $t$ (that is not a fixed constant, but may depend on the size of $C$). Similarly to the $W$-hiearchy, we define $\#W[t]$ as any problem $Q$ for which there exists an FPT parsimonious reduction from $Q$ to $\#$WCS[$C_{t,d}$] for some constant $t$. If $t$ is arbitrary, then $Q$ is in  $\#W[P]$. Similarly to the $W$-hierarchy, it here too is believed that:

\begin{equation}
    \text{FPT} \subsetneq \#W[1]\subsetneq \#W[2]\subsetneq \ldots \subsetneq \#W[t] \subsetneq \#W[P] \subsetneq \text{XP}
\end{equation}

\textbf{Para-NP.} A problem is in \emph{para-NP} concerning some paramater $k$ if there exists a \emph{non-deterministic} algorithm which solves the problem in $O(|\mathcal{X}|^{O(1)}\cdot g(k))$ time for some computable function $g$. A problem is \emph{para-NP-Hard} if and only if the non-parameterized problem is NP-Hard when $k$ is set to some constant.

\textbf{Beyond Para-NP.} Other relevant classes are the extensions of the para-NP class to other classes, such as \emph{para-coNP}, \emph{para-$\Sigma^P_2$}, etc. ~\cite{flum2003describing}. Let $\mathcal{K}$ be a classical complexity class. Then para-$\mathcal{K}$ is the class of all parameterized complexity problems $P$, with $P\subseteq \Sigma^*\times\mathbb{N}$, for which there exists an alphabet $\Pi$, a computable function $g:\mathbb{N}\to \Pi^*$, and a problem $Q\subseteq \Sigma^*\times\Pi^*$ such that $\mathcal{X}\in \mathcal{K}$ and for all instances $(\mathcal{X},k)\in\Sigma^*\times\mathbb{N}$ of $P$ we have:

\begin{equation}
(\mathcal{X},k)\in P \iff (\mathcal{X},g(k))\in Q
\end{equation}

Intuitively, the class para-$\mathcal{K}$ contains all problems that are in $\mathcal{K}$ after a \emph{pre-computation} involving the parameter. Put differently, a problem is in para-$\mathcal{K}$ if it can be solved by two algorithms $\mathbb{P}$ and $\mathbb{A}$, where $\mathbb{P}$ is arbitrary and $\mathbb{A}$ has resources that are constrained by $\mathcal{K}$. The pre-computation that is performed by $\mathbb{P}$ involves only the paramater, which then transforms $k$ into a string $g(k)$. Then, the second algorithm $\mathbb{A}$ solves the problems given $g(k)$ from the pre-computation and the original input $x$, with resources that are constrained by the complexity class $\mathcal{K}$. We define a problem as \emph{para-$\mathcal{K}$-Hard} if there exists an FPT-reduction from any problem in para-$\mathcal{K}$ to it. Alternatively, a problem is para-$\mathcal{K}$-Hard if it is $\mathcal{K}$-Hard even when $k$ is constant.~\cite{flum2003describing, de2016parameterized}. It is widely believed that the following relations hold~\cite{downey2012parameterized, flum2004parameterized}:

\begin{equation}
    \text{XP}\subsetneq \text{para-NP}, \quad \text{XNP}\subsetneq \text{para-}\Sigma^P_2
\end{equation}

\section{Additional Query Formalizations}
\label{Additional_query_formalizations}

Here, we define the two remaining queries referenced throughout the paper: the \emph{Check-Sufficient-Reason} (\emph{CSR}) and \emph{Count Completions} (\emph{CC}) queries, which have also been analyzed in previous works~\cite{BaMoPeSu20, bassanlocal}. We will then provide further details on the computational complexity analysis of computing Shapley values, as discussed in the paper.

The \emph{CSR} query answers whether a specific subset $S$ is a sufficient reason. More formally:

\vspace{0.5em} 

\noindent\fbox{%
    \parbox{\columnwidth}{%
\mysubsection{Check Sufficient Reason (CSR)}:

\textbf{Input}: A model $f$, an instance $\x$, and a subset $S$.

\textbf{Output}: \yes{}, if $S$ is a sufficient reason of $\langle f,\x\rangle$, and \no{} otherwise.
    }%
}

\vspace{0.5em} 

For the \emph{CC} query we consider a \emph{relaxed} version of the \emph{CSR} query which instead of validating whether a specific subset is sufficient or not, asks for the relative \emph{portion} of assignments maintaining a given prediction, given that the other features are independently and uniformly distributed. We start by defining the \emph{completion count} of a given subset:

\begin{equation}
	\label{explanation_definition}
	c(S,f,\x):= \frac{{|\{\z\in \{0,1\}^{|\overline{S}|}; f(\x_{S};\z_{\Bar{S}}) = f(\x)\}|}}{{|\{\z\in \{0,1\}^{|\overline{S}|}|}}
\end{equation}

Now, the \emph{CC} query is defined as follows:

\noindent\fbox{%
	\parbox{\columnwidth}{%
		\mysubsection{CC (Count Completions)}:
		
		\textbf{Input}: Model $f$, input $\x$, and subset of features $S$.
		
		\textbf{Output}: The completion count $c(S,f,\x)$.
	}%
}

We provide here a more expanded formalization of the \emph{shapley-value} that is incorporated in the paper. The \emph{shapley value} attribution is:

\begin{equation}
    \label{eq:explanation}
    \phi_i(f,\x):=\sum_{S\subseteq [n]\setminus \{i\}}\frac{|S|!(n-|S|-1)!}{n!}(v(S\cup \{i\})-v(S))
\end{equation}

where $v(S)$ is the \emph{value function}, and we use the common \emph{conditional expectation} value function $v(S):=\mathbb{E}_{\z\sim \mathcal{D}_p}[f(\z)| \z_S=\x_S]$~\cite{sundararajan2020many, lundberg2017unified}. We follow common conventions in frameworks that assessed the computational complexity of computing \emph{exact} calculations of Shapley values~\cite{arenas2023complexity, van2022tractability}, as well as practical frameworks that compute Shapley values, such as the kernelSHAP method in the SHAP libary~\cite{lundberg2017unified}, and assume that each input feature is independent of all other features, or in other words, every feature $i\in[n]$ is assigned some probability value $[0,1]$, i.e., $p:[n]\to[0,1]$. These are called \emph{product distributions} in the work of \cite{arenas2023complexity} or \emph{fully factorized} in the work of \cite{van2022tractability}. We then formally define our distributions $\mathcal{D}_p(\x)$ as:

\begin{equation}
    \label{eq:explanation}
    \mathcal{D}_p(\x):=\Big({\displaystyle \prod_{i\in [n]; \x_i=1} p(i)}\Big)\cdot \Big({\displaystyle \prod_{i\in [n], \x_i=0} (1-p(i))}\Big)
\end{equation}

Clearly the uniform distribution is a special case of $\mathcal{D}_p$, obtained by setting $p(i):=\frac{1}{2}$ for every $i\in[n]$.

\section{Framework Extensions}
\label{framework_extensions_section}

\textbf{Input and Output Domains.} To make our proofs cleaner and easier to understand, we followed common conventions~\cite{BaMoPeSu20, arenas2022computing, waldchen2021computational, ArBaBaPeSu21, bassanlocal} and presented them using boolean input and output values. It should be emphasized that our analysis is not confined to binary input feature domains but is also applicable to features with $k$ possible \emph{discrete} values, where $k$ is any integer. Furthermore, we can modify our method to include inputs incorporating \emph{continuous} input domains. We will next provide a short overview of the various contexts in which this extension is applicable.

Regarding MLP explainability queries, 
earlier research indicates that the complexity of a \emph{satisfiability} 
query on an MLP extends to scenarios involving continuous inputs. Specifically, the work of~\cite{katz2017reluplex} and~\cite{salzer2021reachability} proves that verifying an arbitrary satisfiability query on an MLP with ReLU activations, over a  
continuous input domain, remains NP-complete. The \emph{CSR} query mentioned in this work, when $S:=\emptyset$ is akin to \emph{negating} a satisfiability query, and this implies that the \emph{CSR} query in MLPs remains coNP complete for the continuous case as well. We recall that the complexity of the \emph{MSR} query for MLPs is $\Sigma^P_2$-Complete~\cite{BaMoPeSu20}. This complexity arises from the use of a coNP oracle, which determines whether a subset of features is sufficient, essentially addressing the \emph{CSR} query. Given that \emph{CSR} can also be adapted to handle continuous outputs, the logic applied to \emph{CSR} can similarly be applied to demonstrate that the \emph{MSR} query can be extended to continuous domains.

%

For Perceptrons, the completeness proofs remain valid in a continuous domain for the explainability queries. The continuity of inputs does not 
alter the membership proofs, for the same reasons that hold for MLPs. For hardness proofs, notice that all 
reductions that were derived from the subset sum (\emph{SSP}) problem (or generalized subset sum (\emph{GSSP}) problem), can be adjusted to substitute any call to $\max\{{z_{i},z_{j}}\}$ in our original proof with $\max([z_{i},z_{j}])$.

Finally, the proofs that apply to tree classifiers (or ensemble tree classifiers) for queries are equally valid for continuous inputs. This extension to continuous inputs was demonstrated by previous works (see for instance,~\cite{huang2021efficiently}). This logic implies also to the complexity of ensembles consisting of the aggregation of a few decision trees.

\textbf{Regression and Probabilistic Classification.} 
Another avenue for extending our framework could involve redefining the explanation forms we have proposed to be more flexible, allowing them to be applied to different contexts such as probabilistic classification or regression.

Potential relaxations of our definitions might include integrating \emph{probabilistic} concepts of sufficiency~\cite{waldchen2021computational, izza2021efficient, arenas2022computing, wang2021probabilistic}, applying them within bounded $\epsilon$-ball domains~\cite{wu2024verix, izza2024distance, la2021guaranteed}, or focusing on meaningful distributions~\cite{yu2023eliminating, gorji2022sufficient}. Additionally, our definitions could be expanded beyond binary classification to address regression or probabilistic classification scenarios. For instance, in the context of a regression model $f:\mathbb{F}\to \mathbb{R}$, a \emph{sufficient reason} might be defined as a subset $S\subseteq\{1,\ldots,n\}$ of input features such that:

\begin{equation} \forall\z\in\mathbb{F} \quad ||f(\x_{S};\z_{\Bar{S}})-f(\x)||_p \leq \delta \end{equation}

for some $0\leq\delta\leq 1$ and an appropriate $\ell_p$-norm. Other notations explored in our work, such as contrastive explanations, can also be adapted within this framework. It is important to note, however, that some complexity results related to Shapley values may differ when transitioning from classification to regression. For example, while computing Shapley values for linear regression models is computationally feasible, the same task may become intractable when applied to classification~\cite{van2022tractability}.

\textbf{Homogeneous and Heterogeneous Ensembles.} The complexity analysis primarily focuses on homogeneous ensembles (ensembles containing base models of the same type). However, most of our results extend to heterogeneous ensembles as well. Clearly, all hardness results for homogeneous ensembles also apply to heterogeneous ensembles and will always align with the "hardest" complexity class among the associated models. For example, consider an ensemble composed of both linear models and decision trees. Suppose that for some explainability query $Q$, interpreting an ensemble of linear models is $\mathcal{K}_1$-Complete for a complexity class $\mathcal{K}_1$, and interpreting an ensemble of decision trees is $\mathcal{K}_2$-Complete for a complexity class $\mathcal{K}_2$. If we assume, without loss of generality, that $\mathcal{K}_1 \subsetneq \mathcal{K}_2$, then a heterogeneous ensemble consisting of both linear models and decision trees will be $\mathcal{K}_2$-Hard. This holds for both our non-parameterized results (Section~\ref{non_paramaterized_appendix_section}) and parameterized results (Section~\ref{paramterized_complexity_part_main_paper_section}). For instance, according to Proposition~\ref{main_result_non_paramterized_main_text}, for an ensemble comprising decision trees and Perceptrons, computing the MSR query would be $\Sigma^P_2$-Hard. Furthermore, based on Proposition~\ref{main_paper_fbdd_ensemble_paramterized_results} and Proposition~\ref{k_ensemble_perceptrons_hardness_paper}, obtaining the MCR query for a heterogeneous ensemble containing both Perceptrons and decision trees is para-NP-Hard (though when the ensemble consists only of decision trees, this query is in coW[1]).

\section{Proof of Proposition~\ref{main_result_non_paramterized_main_text}}
\label{non_paramaterized_appendix_section}

\begin{proposition}
\label{not_paramaterized_results_appendix}
    Ensembles of decision trees and ensembles of Perceptrons are
    \begin{inparaenum}[(i)] \item coNP-Complete with respect to CSR, \item NP-Complete with respect to MCR, \item $\Sigma^P_2$-Complete with respect to MSR, \item $\#P$-Complete with respect to CC, and \item $\#P$-Hard with respect to SHAP.\end{inparaenum}
\end{proposition}


\emph{Proof.} We will, in fact, prove this claim for a broader class of models, which we define as \emph{poly-subset-constructible} models (and for the \emph{MCR} query we will require an additional constraint). We will then demonstrate that both decision trees and Perceptrons fall into this category. Intuitively, poly-subset-constructable models are those for which, given a partial assignment over a subset $\x_S$, we can polynomially construct a function that returns $1$ if and only if the features in $S$ are assigned the values in $\x$. More formally:


\begin{definition}
We say that a class of models $\mathcal{C}$ is \textbf{poly-subset constructable} iff given an assignment $\x\in\mathbb{F}$, and a subset $S\subseteq \{1,\ldots,n\}$, it is possible to construct a model $f\in \mathcal{C}$, in polynomial time, for which for all $\y\in\mathbb{F}$ it holds that:
\begin{equation}
\begin{aligned}
	f(\mathbf{y})=\begin{cases}
		1 \quad if \ \
  \mathbf{y}_S=\mathbf{x}_S \\
		0 \quad otherwise
	\end{cases}
\end{aligned}
\end{equation}
\end{definition}

We will start by proving that each one of the models analyzed within our framework are poly-subset constructable:

\begin{lemma}
\label{perceptrons_fbdds_mlps_poly_subset}
    Perceptrons, decision trees, and MLPs are all poly-subset constructable.
\end{lemma}

\emph{Proof.} We will begin with decision trees. Given an assignment $\x$ and a subset $S$, we can simply construct a decision tree with a single accepting path $\alpha$ that corresponds to the partial assignment of the features in $S$ with their respective values from $\x$. It clearly follows that $\forall \z \in \mathbb{F}, \ f(\x_S; \z_{\Bar{S}}) = 1$. Additionally, for any $\mathbf{y} \in \mathbb{F}$ that does not match the assignments of $\x$ on $S$, we have that $f(\mathbf{y}) = 0$.


For Perceptrons, given some input $\x \in \mathbb{F}$ and a subset $S$, we construct a perceptron with $n$ input features. The single hidden layer $h^{1}$ is constructed as follows:

\begin{equation}
h^{1}_i := \begin{cases}
    1 \quad if \ \ \ \x_i=1 \wedge i\in S \\
    -1 \quad if \ \ \ \x_i=0 \wedge i\in S \\
    0 \quad if \ \ \ i\not\in S
\end{cases}
\end{equation}

We additionally define the single bias term as follows:

\begin{equation}
b^1:= 
- [\sum_{1\leq i\leq n}(h_{i}^{1}\cdot \x_{i})]+\frac{1}{2}
\end{equation}

It clearly satisfies that:

\begin{equation}
\sum_{1\leq i\leq n}(h_{i}^{1}\cdot \x_{i}) = |\{i\in S\wedge \x_i=1\}|
\end{equation}

Moreover, it holds that:

\begin{equation}
\sum_{i \in S \wedge \x_i=1}\z_i\cdot h^{1}_i + \sum_{i \in \Bar{S}\wedge \x_i=0}\z_i\cdot h^{1}_i = \sum_{i \in S \wedge \x_i=1}\z_i -\sum_{i \in \Bar{S}\wedge \x_i=0}\z_i
\end{equation}

Now assume some assignment $(\x_S;\z_{\Bar{S}})$ for any $\z\in\mathbb{F}$. It satisfies that:

\begin{equation}
\begin{aligned}
f(\x_S;\z_{\Bar{S}}) = step(\sum_{i \in S \wedge \x_i=1}(\x_S;\z_{\Bar{S}})_i - \sum_{i \in \Bar{S}\wedge \x_i=0}(\x_S;\z_{\Bar{S}})_i-[\sum_{1\leq i\leq n}(h_{i}^{1}\cdot \x_{i})]+\frac{1}{2})=\\
step(|\{i\in S\wedge \x_i=1\}|-\sum_{i \in S \wedge x_i=1}\z_i\cdot h^{1}_i+\frac{1}{2}) = step(\frac{1}{2})=1
\end{aligned}
\end{equation}

Now assume some assignment $\mathbf{y}\in\mathbb{F}$ for which $S$ does \emph{not} match the values over $\x$. It holds that:

\begin{equation}
\begin{aligned}
f(\mathbf{y}) = step(\sum_{i \in S \wedge \x_i=1}\mathbf{y}_i - \sum_{i \in \Bar{S}\wedge \x_i=0}\mathbf{y}_i-[\sum_{1\leq i\leq n}(h_{i}^{1}\cdot \x_{i})]+\frac{1}{2})=\\
step(|\{i\in S\wedge \x_i=1\wedge \mathbf{y}_i=1\}| - |\{i\in S\wedge \x_i=0\wedge \mathbf{y}_i=1\}|-\sum_{1\leq i\leq n}(h_{i}^{1}\cdot \x_{i})+ \frac{1}{2}) = 0
\end{aligned}
\end{equation}

Hence concluding the construction.

$\qedsymbol{}$

The construction for Perceptrons clearly holds for MLPs as well.


\begin{lemma}
\label{poly_subset_constructable_to_DNF}
Let $\phi$ be a DNF $\phi:=t_1\vee\ldots\vee t_n$ formula, and let $\mathcal{C}$ be a poly-subset constructable class of functions. Then, it is possible to construct, in polynomial time, a hard voting ensemble $f$ consisting of $2n-1$ base-models $f_i\in \mathcal{C}$.
\end{lemma}

\emph{Proof.} We follow a similar reduction to the one proposed in previous work, which demonstrated that obtaining a prime implicant on a random forest is $D^P$-Complete~\cite{izzaexplaining}. The key distinction here is that we need to show that our result holds for any poly-subset constructable class.

Let $\phi := t_1 \vee \ldots \vee t_n$ be a DNF formula. First, we construct $n$ models $\langle f_1, \ldots, f_n \rangle$, where each model $f_i$ corresponds to its respective clause $t_i$. This construction can be completed in polynomial time, given that we assume $\mathcal{C}$ is poly-subset constructable. Each clause $t_i$ corresponds to a partial assignment of values based on the literals that appear in $t_i$. For example, if $t_i := \x_i \wedge \overline{\x_j} \wedge \x_k$, then $t_i$ represents a partial assignment of the features $S := \{i,j,k\}$ with the corresponding assignment of $101$.



We can now leverage the fact that $\mathcal{C}$ is poly-subset constructable to build a model $f_i$ corresponding to each $t_i$ such that for all $i \in \{1,\ldots, n\}, f_i(\x) = t_i(\x)$. To complete the ensemble $f$, we add $n-1$ models that always return True, denoted as $f^t_1, \ldots, f^t_{n-1}$. Our final ensemble is $f := \langle f_i, \ldots, f_n, f^t_1, \ldots, f^t_{n-1} \rangle$. If $\phi$ is true, then at least one clause $t_i$ is satisfied, meaning that at least $n > \frac{2n-1}{2}$ models in $f$ return True, and therefore $f$ is True. If $\phi$ is false, all models $f_1, \ldots, f_n$ return False, which means that at least $n > \frac{2n-1}{2}$ models return False, and thus $f$ is false. This concludes the construction.

$\qedsymbol{}$




\begin{lemma}
\label{proof_of_msr_appendix}
Let $\mathcal{C}$ be some poly-subset constructable class. Then the MSR query for a $k$ ensemble of models from $\mathcal{C}$ is $\Sigma^P_2$-Complete.
\end{lemma}



\emph{Proof.} Membership is evident since we can guess an assignment of features $S$ of size $k$, and then use a coNP oracle to verify that $S$ is indeed sufficient. In other words: $\forall \z\in\mathbb{F}, \ f(\x_S;\z_{\Bar{S}})=f(\x)$. Next, we will prove $\Sigma^P_2$-Hardness. We begin by briefly discussing a proof proposed by~\cite{audemard2022trading}, highlighting a technical gap in this proof, which also appears in a different proof from~\cite{huang2021efficiently}. Finally, we will present an alternative proof that resolves this technical issue.


\textbf{A techincal gap in previous reductions.} 
We begin by referring to a previous reduction proposed by~\cite{audemard2022trading} (Proposition 5), which demonstrates that the \emph{MSR} query for random-forest classifiers is $\Sigma^P_2$-Hard. This is essentially the same objective as our proof, though in our case, we extend the result to a more general class of classifiers (those that are poly-subset constructible) rather than just decision trees. The proof relies on a reduction from the problem of finding \emph{minimal unsatisfiable sets} (\emph{MUS}s) of size $k$, a problem frequently discussed in this context~\cite{ignatiev2019relating}. We will first point out an \emph{existing gap} in this proof, which appears to be similar to gaps in other proofs suggested in previous works~(\cite{huang2021efficiently}, Proposition 7). Afterward, we will present how this gap can be addressed, and demonstrate that the aforementioned problem is indeed $\Sigma^P_2$-Hard. To the best of our knowledge, we are the first to propose a solution to resolve this gap.

The approach behind the reductions in~\cite{audemard2022trading, huang2021efficiently} begins with a given CNF $\phi:=c_1\wedge \ldots \wedge c_m$ and an integer $k$. For each clause, the reductions define $t_i := c_i \vee \overline{s_i}$, introducing a new variable $s_i$ (referred to in~\cite{huang2021efficiently} as the \emph{selector} variable, and denoted in~\cite{audemard2022trading} by $y_i$). Then, they define $\phi' := t_1 \wedge t_2 \wedge \ldots \wedge t_m$, which is still a valid CNF. Negating $\phi'$ produces a DNF equivalent to $\neg \phi'$. The work in~\cite{huang2021efficiently} halts the reduction at this point, as their focus is on proving hardness for DNF classifiers. However,\cite{audemard2022trading} proceeds further by reducing the DNF classifier to an ensemble of decision trees (via a procedure similar to the one provided in Lemma\ref{poly_subset_constructable_to_DNF}, which reduces a DNF to a more general ensemble of poly-subset constructable functions). Next, both reductions assume $\x$ to be a vector containing only ``1''s. The core argument in both reductions is that any selection of $k$ clauses $\{c'_1, \ldots, c'_k\}$ from $c_1, \ldots, c_m$ corresponds to selecting $k$ features $\{s'_1, \ldots, s'_k\}$ from $s_1,\ldots, s_m$. Consequently, they claim that a minimal unsatisfiable set in $\phi$ of size $k$ corresponds to a sufficient reason of size $k$ involving the variables $s_1, \ldots, s_m$ in $f$.

However, we identify a technical gap in these reductions. In the DNF proof from~\cite{huang2021efficiently} and the decision tree ensemble proof from~\cite{audemard2022trading}, the features that can be included in a sufficient reason may include not only the selector variables $s_1, s_2, \ldots, s_m$ (or equivalently $y_1, y_2, \ldots, y_m$ in~\cite{audemard2022trading}), but also the original variables $x_1, x_2, \ldots, x_n$ from the CNF $\phi$. Consequently, a sufficient subset may be chosen that includes both selector and original variables, which does not guarantee equivalence between the two problems.

\textbf{An alternative approach that avoids the technical gap.} Instead of reducing minimal unsatisfiable sets to DNFs and then transforming them further, we will take a different approach. We will reduce from the $\Sigma^P_2$-Complete \emph{Shortest Implicant Core} problem~\cite{umans2001minimum} for DNFs, a problem that is related but not entirely equivalent to finding cardinally minimal unsatisfiable sets.

It is already known that computing the \emph{MSR} query over MLPs is $\Sigma^P_2$-Hard~\cite{BaMoPeSu20}, and this complexity result can be derived from the \emph{Shortest-Implicant-Core} problem, not through DNFs, but rather through \emph{boolean circuits}. Specifically for MLPs, this reduction leverages the fact that any boolean circuit can be reduced to an MLP, making the reduction more adaptable. For ensembles, however, a more complicated approach is needed (which will enable its incorporation to DNFs). We begin by introducing the Shortest Implicant Core problem, first defining an implicant, followed by the formal definition of the Shortest-Implicant-Core problem:


\begin{definition} Consider $\phi$ as a boolean formula over the literals $\{x_1,\ldots,x_n\}$. An implicant $C:=x'_1x'_2,\ldots x'_l$ of $\phi$ is defined as a partial assignment over $\phi$'s literals (for all $1\leq i\leq l$, $x'_i$ is equal to either $x_j$ or $\overline{x_j}$ and each $x'_i$ contains an instance of a different literal), ensuring that any completion of this assignment results in $\phi$ evaluating to true. \end{definition}

For example for the DNF furmula: $\phi:=x_1\overline{x_2}x_3\vee x_1\overline{x_2}\overline{x_3}$, we have that $x_1\overline{x_2}$ is an implicant of $\phi$, since any completion of $x_1\overline{x_2}$ evaluates $\phi$ to True. We now define the \emph{Shortest-Implicant-Core} problem as follows:

\vspace{0.5em} 

\noindent\fbox{%
    \parbox{\columnwidth}{%
\mysubsection{Shortest Implicant Core}:

\textbf{Input}: A DNF Formula $\phi:=t_1\vee\ldots\vee t_n$, and an integer $k$.

\textbf{Output}: \yes{}, if there exists an implicant $C\subseteq t_n$ of $\phi$ of size $k$, and \no{} otherwise.
    }%
}

\vspace{0.5em} 

It is firstly important to mention that the \emph{Shortest-Implicant-Core} problem is generally more suitable than the \emph{Shortest-Implicant} problem for a reduction from the \emph{MSR} query since the \emph{core} $t_n$ is important for representing the specific \emph{local} assignment $\x\in\mathbb{F}$ for which the sufficient reason is provided. Additionally, since we are working with DNFs and not general boolean circuits, it is known that the Shortest-Implicant problem for DNFs is GC-Complete~\cite{umans2001minimum}, which is a class that is strictly less complex than $\Sigma^P_2$.

We will start by briefly presenting the proof presented by~\cite{BaMoPeSu20} for reducing the Shortest Implicant Core problem to the problem of solving the \emph{MSR} query for \emph{MLPs}. We will then discuss the desired modifications needed in order to produce this problem for ensembles containing poly-subset constructable functions instead.

For example, let us assume we have the following DNF formula:

\begin{equation}
    \phi:= x_1\overline{x_5}\vee \overline{x_2}\overline{x_6}\vee x_3 x_6 \vee \overline{x_1} \overline{x_2} x_4 \vee x_1 x_3 x_5
\end{equation}

A straightforward approach to reduce Shortest Implicant Core to MSR is to construct an MLP $f$ that is equivalent to $\phi$, which is feasible since any Boolean circuit can be reduced to an MLP~\cite{BaMoPeSu20}. For the input $\x$, we can create an input where the features corresponding to $t_n$ (in this case, features $1$, $3$, and $5$ are in $C$) are set to $1$, while the remaining features can be assigned other values.


There is an issue with this construction. The problem is that the sufficient reason in the constructed MLP may include features that are not part of $t_n$. Specifically, if we set $k:=2$, we observe that no implicant of size $2$ exists for $\phi$, yet setting features $3$ and $6$ to $1$ determines the prediction of $f(\x)$, indicating that a sufficient reason of size $2$ \emph{does} exist. To address this issue, ~\cite{BaMoPeSu20} suggests constructing a different formula as follows:


\begin{equation}
    \phi':= \bigwedge^3_{i=1}\big(x_1\overline{x_5}\vee \overline{x_2^i}\overline{x_6^i}\vee x_3 x_6^i \vee \overline{x_1} \overline{x_2^i} x_4^i \vee x_1 x_3 x_5\big)
\end{equation}



We can then once again transform this formula into an equivalent MLP $f$, which is possible since, as mentioned before, any Boolean circuit can be transformed into an MLP. More generally, this construction is formalized as follows. Let $X_c$ denote the set of variables that are not mentioned in $t_n$. Then, $\phi'$ will be defined as:

\begin{equation}
\begin{aligned}
    \phi^{(i)}:=\phi[x_j\to x_j^i, \text{ for all } x_j \in X_c] \\
    \phi':= \bigwedge^{k+1}_{i=1} \phi^{(i)}
    \end{aligned}
\end{equation}

However, in our case, the following construction does not apply directly. This is because, as proven in Lemma~\ref{poly_subset_constructable_to_DNF}, any DNF can be transformed into an equivalent ensemble of poly-subset constructable functions. However, this does not necessarily imply that we can reduce any Boolean circuit to such models, and since $\phi'$ is no longer a DNF, we cannot apply the same reduction as in~\cite{BaMoPeSu20}.

Instead, we will address this issue differently. Starting with the DNF $\phi$, we will create a new DNF, $\phi'$, which includes only the literals from $t_n$. We will subsequently demonstrate that any implicant $C \subseteq t_n$ for $\phi$ also serves as an implicant for $\phi'$. To facilitate this construction within polynomial time, we rely on the fact that any term $t_i$ in $\phi$, aside from $t_n$, has at most a constant size. We will thus formalize our problem as follows:

\vspace{0.5em} 

\noindent\fbox{%
    \parbox{\columnwidth}{%
\mysubsection{Shortest Implicant Core (Constant DNF)}:

\textbf{Input}: An integer $k$, a constant integer $d=O(1)$, and a DNF Formula $\phi:=t_1\vee\ldots\vee t_n$, where for any $1\leq i\leq n-1$, $t_i$ contains less than $d$ literals.

\textbf{Output}: \yes{}, if there exists an implicant $C\subseteq t_n$ of $\phi$ of size $k$, and \no{} otherwise.
    }%
}

\vspace{0.5em} 

\begin{claim}
    Shortest Implicant Core (Constant DNF) is $\Sigma^P_2$-Hard.
\end{claim}

\emph{Proof.} We note that this proof directly follows from the hardness proof of the $\Sigma^P_2$-hardness of the Shortest-Implicant problem~\cite{umans2001minimum}. The hardness is established using a reduction from $\textit{QSAT}_2$, where each term in $\phi$ is constructed from a 3-DNF formula conjoined to another constructed 3-DNF. This setup results in a DNF where each term in $\phi$ is of constant size. Consequently, it follows that the Shortest Implicant Core, where each term in $\{t_1,\ldots,t_{n-1}\}$ is of constant size, is also $\Sigma^P_2$-hard to decide.

$\qedsymbol{}$

We will first demonstrate how, from a DNF $\phi:=t_1\vee \ldots \vee t_n$ where the size of any $\{t_1,\ldots t_{n-1}\}$ is bounded by some $d$, we can construct another DNF $\phi'$ in polynomial time and size, where $\phi'$ exclusively includes literals from $t_n$. This approach allows us to later use $t_n$ directly as our desired $\x\in\mathbb{F}$ in the reduction. We will also demonstrate that any implicant $C \subseteq t_n$ for $\phi$ equally serves as an implicant $C \subseteq t_n$ for $\phi'$.

\textbf{The construction.} The idea is that if a term $t_i$ in $\phi$ includes literals not present in $t_n$, we can ``eliminate'' these non-$t_n$ elements by iterating over all possible assignments that cover the literals not included in $t_n$ (and this number is polynomial, as each term in $\phi$ is of constant size). Formally, let there be a DNF $\phi:=t_1 \vee t_2 \vee \ldots t_n$. For each $t_i$ where $1\leq i\leq n-1$, let us denote $r_i$ as the subset of literals from $t_i$ which do not participate in $t_n$. First, we define $\phi'$ to be the disjunction of any $t_i$ for which $1\leq i\leq n-1$ and $r_i$ is empty (in other words the ``part'' of the DNF that only contains assignments of literals that appear in $t_n$).

Now, for each $t_i$, if $r_i$ is not empty, we consider every subset $S \subseteq \{t_1, \ldots, t_{n-1}\}$ where the size of $S$ is $j \leq 2^{|r_i|}$. We use $\{x^i_1, x^i_2, \ldots, x^i_l\}$ to represent the literals participating in $r_i$. By iterating through $S$, we can verify whether any potential assignment for $\{x^i_1, x^i_2, \ldots, x^i_l\}$ is encompassed by $S \cup t_i$. Consider, for example, $r_i:= x_1\overline{x_7}$. The literals involved are $\{x_1, x_7\}$. Assume $S:={[x_1x_7x_8], [\overline{x_1} \overline{x_8} x_9], [\overline{x_1}x_7]}$. Here, every potential assignment to $x_1x_7$ (which includes $[x_1x_7], [\Bar{x_1}x_7], [x_1,\Bar{x_7}], [\Bar{x_1}\Bar{x_7}]$) is encompassed by $S\cup t_i$. Another example of a subset $S$ such that $S\cup t_i$ covers all assignments is: $S:=\{[\overline{x_1}x_5], [x_1x_7x_5]\}$. However, the subset $S:={[x_1x_7], [\overline{x_1}x_7], [x_1\overline{x_7}]}$ fails to enable $S\cup t_i$ to cover all assignments since it misses the assignment $\overline{x_1}\overline{x_7}$. It is worth noting that validating whether $S$ covers all literal assignments can be done in linear time with respect to $|S|$, as each $t_i \in S$ accounts for covered assignments, allowing us to save each covered assignment and, after iterating through the entire $S$, check whether all assignments were covered.


For each $t_i$, we initatie a set $\mathbb{S}_i$. For each $S$ that we iterate on concerning $t_i$, if we have that $S\cup t_i$ covers all assignments of literals of $r_i$ we will add $\bigwedge_{t_l\in S}\overline{r_l}$ to $\mathbb{S}_i$. Now, we are in a position to define our refined DNF formula:

\begin{equation}
    \phi'':= \phi' \vee [ \bigvee_{1\leq i \leq n-1}[\bigvee_{t_l\in \mathbb{S}_i}t_l  \ ] \ ]\vee t_n
\end{equation}

We first note that the construction of $\phi''$ is polynomial. For each $t_i$ we iterate over all subsets of size $2^{|r_i|}$ or less. Overall the number of subsets we iterate on is bounded by:

\begin{equation}
\begin{aligned}
    \sum_{1\leq i\leq n-1}\big(\sum_{1\leq j\leq 2^{|r_i|}}\binom{n}{j}\big)\leq     \sum_{1\leq i\leq n-1}\big(\sum_{1\leq j\leq 2^{|r_i|}}n^j\big)\leq \sum_{1\leq i\leq n-1}\big(2^{|r_i|}\cdot n^{2^{|r_i|}}\big) \leq \\
    \sum_{1\leq i\leq n-1}\big(2^{d}\cdot n^{2^{d}}\big)\leq n\cdot\big(2^{d}\cdot n^{2^{d}}\big)
\end{aligned}
\end{equation}

and since $d$ is constant, the derived term is polynomial in $n$. We have that both the runtime of constructing $\phi''$ as well as the size of $\phi''$ are polynomial in $n$. We now will prove the following lemma regarding the construction:

\begin{claim}
    Any $C\subseteq t_n$ is an implicant for $\phi$ if and only if it is an implicant for $\phi''$.
\end{claim}

\emph{Proof.} If $C \subseteq t_n$ is an implicant, this indicates the existence of a subset $S \subseteq \{t_1, \ldots, t_n\}$ such that the partial assignment $C$ ensures any completion of $C$ within $\bigvee_{t_l \in S} t_l$ evaluates to true (thereby guaranteeing that any completion of $C$ within $\phi$ also evaluates to true). Initially, we assume that each $t_l \in S$ incorporates literals from $t_n$. This leads to the identification of this subset as a subset of terms in $\phi'$, and consequently in $\phi''$. Thus, we identify a subset $S' \subseteq \{t'_1, \ldots, t'_m\}$, where $\{t'_1, \ldots, t'_m\}$ represents the terms of $\phi''$, satisfying $\bigvee_{t_l \in S} t_l = \bigvee_{t_l \in S'} t_l$. Therefore, any extension of $C$ in $\bigvee_{t_l \in S} t_l$ that holds true, ensures that any extension of $C$ in $\bigvee_{t_l \in S'} t_l$ is true (and by extension, any completion of $C$ in $\phi''$). This leads us to conclude that $C$ is also an implicant of $\phi''$.





Revisiting our earlier notation, let $r_i$ represent the set of literals from $t_i$ that are absent in $t_n$. For any $t_l\in S$ where $r_l$ is non-empty, it is required that all assignments to the literals of $r_l$ are covered by $S$ (failing which, there will be an assignment making $\bigvee_{t_l \in S} t_l$ untrue). Essentially, $S$ consists of each $t_i$ where $r_i$ is non-empty, including terms that cover all assignments for $r_i$, and potentially terms where $r_i$ is empty (which would then be included in $\phi'$). We observe that any assignment covering all assignments for $r_i$ can be at most of size $2^{|r_i|}$. This assignment will be generated by our approach and incorporated within $\bigvee_{t_i\in S}t_i$ and subsequently in $\phi''$. Consequently, we identify a subset $S'\subseteq \{t'_1,\ldots,t'_m\}$, where $\{t'_1,\ldots,t'_m\}$ are the terms of $\phi''$, such that $\bigvee_{t_l \in S} t_l=\bigvee_{t_l \in S'} t_l$. Thus, any completion of $C$ over $\bigvee_{t_l \in S} t_l$ that proves true, equally confirms that any completion of $C$ over $\bigvee_{t_l \in S'} t_l$ is true (and thus any completion of $C$ over $\phi''$ is true). Ultimately, this demonstrates that $C$ is also a prime implicant of $\phi''$.


For the second direction, let us assume that $C$ is an implicant for $\phi''$. Consequently, there is a subset $S \subseteq \{t'_1, \ldots, t'_m\}$ of terms from $\phi''$, denoted by $\{t'_1, \ldots, t'_m\}$, where any completion of $C$ over $\bigvee_{t_l \in S'} t_l$ invariably results in true (thus guaranteeing any completion of $C$ over $\phi''$ is true). If $S$ is entirely within $\phi'$, it follows that $S$ is also part of $\phi$, and therefore, there exists a corresponding set $S' \subseteq \{t_1, \ldots, t_n\}$ in which any completion of $C$ over $\bigvee_{t_l \in S} t_l$ is true, as is any completion of $C$ over $\phi$. Therefore, $C$ is also an implicant of $\phi$.


Now, if $S$ includes terms that are not from $\phi'$, these must be assignments in the form of $\bigwedge_{t_l\in S''}\overline{r_l}$, where $S''$ is a subset of $\{t_1,\ldots,t_{n}\}$ and covers all potential assignments over the literals of $r_i$ for some $t_i\in \{t_1,\ldots t_{n}\}$. Consequently, we can select the terms from each $S''$, as well as those from $\phi'$, to form a corresponding subset $S' \subseteq \{t_1,\ldots,t_{n}\}$. It is important to note that all terms in $S'$ that lack any literals from $t_n$ are part of $\phi'$ and thus appear equivalently in both $S$ and $S'$. Additionally, for any term $t_l \subseteq S'$ where $r_l \neq \emptyset$, all assignments over the literals in $r_l$ are covered by both $S$ and $S'$. Therefore, we conclude that $\bigvee_{t_l \in S} t_l = \bigvee_{t_l \in S'} t_l$; thus, if any completion of $C$ confirms that $\bigvee_{t_l \in S} t_l$ is true, it also confirms that $\bigvee_{t_l \in S'} t_l$ is true (and thus $\phi$). Ultimately, this shows that $C$ is also a prime implicant of $\phi$, completing the proof of the claim.

$\qedsymbol{}$

This claim proves that a prime implicant $C \subseteq t_n$ of size $k$ exists for $\phi$ if and only if a prime implicant $C \subseteq t_n$ of size $k$ exists for $\phi''$, thus concluding that the Shortest Implicant Core problem for constant DNF is $\Sigma^P_2$-Hard.

$\qedsymbol{}$

\textbf{Concluding the reduction.} We now finalize our proof for reducing the Shortest-Implicant-Core (Constant DNF) to the MSR for an ensemble of poly-subset constructable functions. Given a tuple $\langle \phi, C, k, d \rangle$, we can construct $\phi'$ from $\phi$ in polynomial time, a task feasible particularly because $d$ is constant. Utilizing Lemma~\ref{poly_subset_constructable_to_DNF}, we convert $\phi'$ into an equivalent ensemble $f$ of poly-subset constructable functions. Notably, the literals in $\phi'$ are exclusively those from $t_n$. Therefore, we use $t_n$ to construct our input vector, where any positive assignment in $C$ is represented as a ``1'' in $\x$ and any negative assignment as a ``0''. For example, if $t_n:=\x_1\overline{\x_2}\x_3$, we would set $\x:=(101)$.


Now, since the input $\x$ includes only features present in $f$, we can assert that an implicant in $t_n$ of size $k$ for $\phi'$ exists if and only if a sufficient reason of size $k$ exists for $\langle f,\x\rangle$. Given that any implicant $C$ in $t_n$ for $\phi'$ also serves as an implicant $C$ in $\phi$, it follows that an implicant of size $k$ for $\phi$ exists if and only if there is a sufficient reason of size $k$ for $\langle f,\x\rangle$, thereby completing our reduction.

$\qedsymbol{}$

In contrast to the \emph{MSR} query which was non-trivial, for the \emph{CSR}, \emph{MCR}, \emph{CC}, and \emph{SHAP} queries we can incorporate similar reductions to other works~\cite{izza2021efficient} which show hardness for \emph{DNFs}, since we know that an ensemble which is encompassed by models from a class that is poly-subset constructable, can be reduced to DNFs (Lemma~\ref{poly_subset_constructable_to_DNF}).

\begin{lemma}
Let $\mathcal{C}$ be some poly-subset constructable class of functions. Then the CSR query for a $k$ ensemble of models from $\mathcal{C}$ is coNP-Complete.
\end{lemma}

\emph{Proof.} While these results can be derived from the notable works of~\citep{audemard2022trading} and~\citep{ordyniak2024explaining} in the specific setting of majority-voting decision tree ensembles, we provide a direct proof here for the more general class of poly-subset constructable models, covering both majority and weighted voting schemes. Membership in coNP is straightforward since we can guess an assignment $\z\in\mathbb{F}$ and verify whether $f(\x_S; \z_{\Bar{S}}) \neq f(\x)$ to determine if $S$ is \emph{not} a sufficient reason. For the hardness result, we can apply the same proof provided by~\cite{BaMoPeSu20}, which involves a reduction from the tautology (\emph{TAUT}) problem for DNFs. In their work, DNFs are reduced to equivalent MLPs to establish coNP-hardness for MLPs. Similarly, in our case, we can utilize the exact same reduction since, in Lemma~\ref{poly_subset_constructable_to_DNF}, we demonstrated that any DNF can be reduced to an ensemble of models from $\mathcal{C}$, where $\mathcal{C}$ is poly-subset constructable. This leads to the conclusion that the problem is coNP-complete.

$\qedsymbol{}$

In the specific case of \emph{MCR}, we will establish the complexity of the problem for a set of models $\mathcal{C}$ that remains poly-subset constructable, but also adheres to an additional property, which we define as being ``closed under symmetric construction''. This property is formalized as follows:

\begin{definition}
Let $\mathcal{C}$ be a class of functions. Then $\mathcal{C}$ is closed under symmetric construction iff for any $f\in \mathcal{C}$, then $\neg f$ can be constructed in polynomial time.
\end{definition}

We will prove that ensembles of decision trees, Perceptrons, and MLPs all satisfy the aforementioned property:



\begin{lemma}
\label{ensembles_are_symmetric}
The class of ensembles of decision trees, the class of ensembles of MLPs, and the class of ensembles of Perceptrons are all closed under symmetric construction. In other words, suppose we are given $f$ --- an ensemble of $k$ models $f_1,\ldots,f_k$ that are all either Perceptrons, decision trees, or MLPs; then, $f':=\neg f$ can be constructed in polynomial time.
\end{lemma}


\emph{Proof.} We first note that negating individual decision trees, Perceptrons, and MLPs can be done in polynomial time, as demonstrated by~\cite{amir2024hard}. Applying this transformation to any decision tree $f_i$ within the ensemble results in negating the entire model $f$. The only asymmetric element in this negation arises in majority voting ensembles—where a ``1'' classification occurs iff $\ceil{\frac{k}{2}}$ of the models are classified as "1"—and the negation effectively changes the ensemble to classify ``1'' iff $\floor{\frac{k}{2}}$ models are classified as ``1''. This asymmetry can be addressed by constructing an additional model, when needed, that always classifies ``1''. This can be done using Perceptrons, decision trees, or MLPs. Similarly, in weighted voting ensembles, asymmetry may arise due to the distinction between strict and non-strict inequalities. In other words, while $f$ classifies as ``1'' iff $\sum_{1\leq i\leq k}\phi_i\geq0$, in the negated model $f'$, it will classify as ``1'' iff $\sum_{1\leq i\leq k}\phi_i>0$. This can again be resolved by adding an extra model $f_i$ that always classifies ``1'' with a very small weight, thereby addressing this issue. Given that any weight in $f_i$, $\phi_i$, is a rational number $\frac{p_i}{q_i}$, the weight of the newly constructed model can be set to $q_1\cdot q_2\ldots \cdot q_k$. This additional model will correct the asymmetry in cases where the weighted sum of all models is exactly zero, allowing the ensemble to be negated.

$\qedsymbol{}$

We are now prepared to prove the following claim, which will establish that solving \emph{MCR} for ensembles of decision trees, Perceptrons, and MLPs is NP-complete. This result follows directly from Lemma~\ref{ensembles_are_symmetric} and Lemma~\ref{perceptrons_fbdds_mlps_poly_subset}.

\begin{lemma}
Let $\mathcal{C}$ be some poly-subset constructable class of functions, which is also closed under symmetric construction. Then the MCR query for a $k$ ensemble of models from $\mathcal{C}$ is NP-Complete.
\end{lemma}

\emph{Proof.} Similarly to CSR, although these results can be derived from the notable works of~\citep{audemard2022trading} and~\citep{ordyniak2024explaining} in the specific case of majority-voting decision tree ensembles, we present a direct proof for the more general class of poly-subset constructable models, encompassing both majority and weighted voting schemes. Membership holds because we can guess a subset $S$ and check whether $|S|\leq d$ and $f(\x_S;\z_{\Bar{S}})\neq f(\x)$, which confirms the existence of a contrastive reason of size $d$. 

For hardness, we can use a similar proof to~\cite{BaMoPeSu20}, which demonstrates that the MCR problem is NP-Hard for MLPs. They reduce the problem from the \emph{Vertex-Cover} problem, which is NP-Complete. To achieve this, given a graph $G:=\langle V,E\rangle$, they construct an equivalent CNF formula:


\begin{equation}
    \bigwedge_{(u,v)\in E} (x_u \vee x_v)
\end{equation}

and encode the CNF formula as an equivalent MLP. In Lemma~\ref{poly_subset_constructable_to_DNF}, we proved that any DNF can be reduced to an ensemble with poly-subset constructable base-model functions. However, to extend the proof of~\cite{BaMoPeSu20} to apply to our ensembles, we must show that any \emph{CNF} can also be transformed into an ensemble constructed from these models. This is possible because the family of models $\mathcal{C}$ is not only poly-subset constructable but also closed under symmetric construction. We begin by proving the following relation:

\begin{claim}
\label{poly_subset_constructable_to_CNF}
Let $\phi$ be a CNF $\phi:=t_1\wedge\ldots\wedge t_n$ formula, and let $\mathcal{C}$ be a poly-subset constructable class of functions. Then, it is possible to construct, in polynomial time, a hard voting ensemble $f$ consisting of $2n-1$ base-models $f_i\in \mathcal{C}$ where $\mathcal{C}$ is both a poly-subset constructable class, and is closed under symmetric construction.
\end{claim}

\emph{Proof.} We begin by negating $\phi$ in polynomial time and derive a DNF formula equivalent to $\neg \phi$. From Lemma~\ref{poly_subset_constructable_to_DNF}, we know that any DNF can be transformed in polynomial time into an ensemble of models $f \in \mathcal{C}$ (since $\mathcal{C}$ is poly-subset constructable). Given that $\mathcal{C}$ is also closed under symmetric construction, we can negate $f$ and construct $\neg f$ in polynomial time. Thus, we obtain an ensemble $f' := \neg f$ equivalent to $\phi$, completing our proof.

$\qedsymbol{}$

We can now conclude our proof, as we know that ensembles of decision trees, Perceptrons, and MLPs are all poly-subset constructable and closed under symmetric construction. Therefore, any CNF can be reduced to an equivalent ensemble composed of models from this class, allowing us to leverage the vertex-cover problem to establish NP-Hardness.

$\qedsymbol{}$

\begin{lemma}
    Let $\mathcal{C}$ be some poly-subset constructable class of functions. Then the CC query for a $k$ ensemble of models from $\mathcal{C}$ is $\#P$-Complete.
\end{lemma}

\emph{Proof.} Membership is straightforward by definition. For the hardness result, we can reduce from the Model Counting problem for DNFs, which is known to be $\#P$-Hard~\cite{valiant1979complexity}. Given a DNF $\phi$, we can construct an equivalent ensemble of poly-subset constructable functions using Lemma~\ref{poly_subset_constructable_to_DNF} and set $S := \emptyset$ for the \emph{CC} query. Thus, solving \emph{CC} is equivalent to model counting, and the reduction holds.

$\qedsymbol{}$

\begin{lemma}
    Let $\mathcal{C}$ be some poly-subset constructable class of functions. Then the SHAP query for a $k$ ensemble of models from $\mathcal{C}$ is $\#P$-Hard.
\end{lemma}

\emph{Proof.} We follow the proof of~\cite{arenas2021tractability}, which demonstrated a connection between computing \emph{SHAP} under fully factorized distributions and the model counting problem. The following relation is established:

\begin{equation}
    \#f:=f(\x)-2^n\cdot \sum_{i\in \{1,\ldots,n\}} \phi_i(f,\x)
\end{equation}

where $\#f$ represents the number of positive assignments of $f$ and $\phi$ denotes the Shapley value. This establishes that computing \emph{SHAP} is at least as hard as the model counting problem. Since, as shown in Lemma~\ref{poly_subset_constructable_to_DNF}, any DNF can be constructed into an ensemble of poly-subset constructable functions, and model counting for DNFs is known to be $\#P$-Hard~\cite{valiant1979complexity}, this concludes the proof, demonstrating $\#P$-Hardness for computing \emph{SHAP}.

$\qedsymbol{}$

\section{Proof of Proposition~\ref{CC_and_SHAP_proposition}}
\label{pseudo_plynomial_theorem_appendix_section}

\begin{proposition}
\label{pseudo_plynomial_theorem_appendix}
    While the CC and SHAP queries can be solved in pseudo-polynomial time for Perceptrons, ensemble-Perceptrons remain $\#P$-Hard even if the weights and biases are given in unary.
\end{proposition}


\emph{Proof.} For the \emph{CC} query, we refer to the results from \cite{BaMoPeSu20}, which demonstrated that when the weights and biases are provided in unary, the \emph{CC} query can be solved in polynomial time for perceptrons. However, according to the findings in Proposition~\ref{not_paramaterized_results_appendix}, when the weights and biases are not given in unary, the \emph{CC} query is $\#P$-Complete for \emph{ensembles} of perceptrons.


For the \emph{SHAP} query, we use a proof that was proposed by the work of~\cite{arenas2023complexity} which showed a connection between the computation of shapley values for models (under some mild conditions) and the following portion:

\begin{equation}
H_{\mathcal{D}_p}(f,\x,S,k):=\sum_{S\subseteq[n], |S|=k}\mathbb{E}_{\z\sim \mathcal{D}_p}[f(\z)| \z_S=\x_S]
\end{equation}

Specifically, this relationship applies to models that are \emph{closed under conditioning}, defined as follows:

\begin{definition}
\label{closed_under_conditioning}
    A model $f:\mathbb{F}\to\{0,1\}$ is closed under conditioning if given some assignment $\x\in\mathbb{F}$ and some subset $S\in[n]$, we can construct in polynomial time a function $f':\mathbb{F}\to\{0,1\}$ for which it holds that for all $\y\in\mathbb{F}$: $f'(\y)=f(\x_S;\y_{\Bar{S}})$.
\end{definition}

The connection between models that are closed under conditioning and the computation of $H_{\mathcal{D}_p}$ is as follows~\cite{arenas2023complexity}:

\begin{lemma}
    Let there be a model $f$ which is closed under conditioning. Then solving the \emph{SHAP} query for $f$ and any $i\in[n]$ can be reduced, in polynomial time, to the problem of computing $H_{\mathcal{D}_p}$
\end{lemma}

To apply this lemma to our scenario, we begin by proving the following claim:

\begin{claim}
    Any perceptron $f:=\langle \wb,b\rangle$ is closed under conditioning.
\end{claim}

\emph{Proof.} Given a perceptron $f := \langle \wb, b \rangle$, an assignment $\x \in \mathbb{F}$, and a subset $S \in [n]$, we can construct a second perceptron $f' := \langle \wb', b' \rangle$, where $\mathbf{w}' := \mathbf{w} \cdot \mathbf{1}_{\Bar{S}}$. In other words, we ``zero-out'' all the feature values in $\mathbf{w}$ corresponding to the set $S$ and leave the features in $\overline{S}$ unchanged. Additionally, we define $b' := b + \sum_{i \in S, \x_i = 1} \mathbf{w}_i$. It now holds that for any $\y \in \mathbb{F}$:


\begin{equation}
\begin{aligned}
    f'(\y) = \sum_{i\in[n]}\mathbf{w}'_i\cdot \y_i+b'= \sum_{i\in S}\mathbf{w}'\cdot \y_i+\sum_{i\in \Bar{S}}\mathbf{w}'\cdot \y_i+(b+\sum_{i\in S, \x_i=1}\mathbf{w}_i)=\\
    \sum_{i\in \Bar{S}}\mathbf{w}\cdot \y_i+(b+\sum_{i\in S, \x_i=1}\mathbf{w}_i)=\\ \sum_{i\in \Bar{S}}\mathbf{w}\cdot \y_i+(b+\sum_{i\in S, \x_i=1}\mathbf{w}_i\cdot\x_i+\sum_{i\in S, \x_i=0}\mathbf{w}_i\cdot\x_i)=f(\x_S;\y_{\Bar{S}})
    \end{aligned}
\end{equation}

$\qedsymbol{}$

Now, we only need to prove that when the weights and biases of a perceptron $f$ are given in unary, $H_{\mathcal{D}_p}(S,k)$ can be computed in polynomial time. This will complete our proof that the \emph{SHAP} query can be solved in pseudo-polynomial time for perceptrons. We use the notation $sim(\z,\x)$ to denote the set of features $S$ such that for all $i \in S$, it holds that $\z_i = \x_i$. We begin by demonstrating the following relations:


\begin{equation}
\begin{aligned}
H_{\mathcal{D}_p}(S,k):=\sum_{S\subseteq[n], |S|=k}\mathbb{E}_{\z\sim \mathcal{D}_p}[f(\z)| \z_S=\x_S]=\\
\sum_{\z\in\mathbb{F}, |sim(\z,\x)|\geq k}\binom{|sim(\z,\x)|}{k}\cdot \big( \prod\limits_{i\in[n],\z_i=1} \big(p(i)\big) \prod\limits_{i\in[n],\z_i=0} \big(1-p(i))\big)\cdot f(\z)=\\
\sum_{\z\in\mathbb{F}, |sim(\z,\x)|\geq k, f(\z)=1}\binom{|sim(\z,\x)|}{k}\cdot \big( \prod\limits_{i\in[n],\z_i=1} \big(p(i)\big) \prod\limits_{i\in[n],\z_i=0} \big(1-p(i))\big)=\\
\sum_{j=k}^{j=n}\big(\sum_{\z\in\mathbb{F}, |sim(\z,\x)|= j, f(\z)=1}\binom{|sim(\z,\x)|}{k}\cdot \big( \prod\limits_{i\in[n],\z_i=1} \big(p(i)\big) \prod\limits_{i\in[n],\z_i=0} \big(1-p(i))\big)\big)=\\
\sum_{j=k}^{j=n}\binom{j}{k}\big(\sum_{\z\in\mathbb{F}, |sim(\z,\x)|= j, f(\z)=1}\cdot \big( \prod\limits_{i\in[n],\z_i=1} \big(p(i)\big) \prod\limits_{i\in[n],\z_i=0} \big(1-p(i))\big)\big)=\\
\sum_{j=k}^{j=n}\binom{j}{k}\big(\sum_{\z\in\mathbb{F}, |sim(\z,\x)|= j}\cdot \big( \prod\limits_{i\in[n],\z_i=1} \big(p(i)\big) \prod\limits_{i\in[n],\z_i=0} \big(1-p(i))\big)\big)\cdot\mathbf{1}_{\{f(\z)=1\}}=\\
\sum_{j=k}^{j=n}\binom{j}{k}\big(\sum_{\z\in\mathbb{F}, |sim(\z,\x)|= j}\cdot \big( \prod\limits_{i\in[n],\z_i=1} \big(p(i)\big) \prod\limits_{i\in[n],\z_i=0} \big(1-p(i))\big)\big)\cdot\mathbf{1}_{\{\sum_{i=1}^n \mathbf{w}_i\cdot \z_i+b\geq 0\}}
\end{aligned}
\end{equation}

We define $M := \max(\mathbf{w}) + 1$. Next, we define the vector $\mathbf{w}'$ as follows:


\begin{equation}
    \mathbf{w}'_i:= \begin{cases}
        -\mathbf{w}_i+M \quad  if \ \ \x_i = 1 \\
        \mathbf{w}_i+M \quad if  \ \ \x_i = 0
    \end{cases}
\end{equation}

We finally define $T := b + \big(\sum_{i \in [n], \x_i = 0} \mathbf{w}_i \big)$. We now prove the following relation:


\begin{claim}
    Given some integer $1\leq j\leq n$, the following relation between $\mathbf{w}$ and $\mathbf{w}'$ holds:
\begin{equation}
\sum_{\z\in\mathbb{F}, |sim(\z,\x)|= j}\mathbf{1}_{\{\sum_{i=1}^n \mathbf{w}_i\cdot \z_i+b\geq 0\}}= \sum_{S\subseteq [n], |S|= j}\mathbf{1}_{\{\sum_{i\in S} \mathbf{w}'_i\leq T+M\cdot j\}}
\end{equation}
\end{claim}

\emph{Proof.} We start by noting that the iterating over all vectors $\z\in\mathbb{F}$ for which $|sim(\z,\x)|=j$ is equivalent to iterating over all subsets $S\subseteq [n], |S|=j$ and taking the values of the features in $S$ to be those of $\x$ and those of $\overline{S}$ to be $\neg \x$. In other words, iterating over all vectors of the form $(\x_S;\neg\x_{\Bar{S}}$ for which $|S|=j$. From this, we can derive the following relation:

\begin{equation}
\begin{aligned}
\sum_{\z\in\mathbb{F}, |sim(\z,\x)|= j}\mathbf{1}_{\{\sum_{i=1}^n \mathbf{w}_i\cdot \z_i+b\geq 0\}}=
\sum_{S\subseteq [n], |S|=j}\mathbf{1}_{\{\sum_{i=1}^n \mathbf{w}_i\cdot (\x_i;\neg\x_i)+b\geq 0\}}=\\
\sum_{S\subseteq [n], |S|=j}\mathbf{1}_{\{(\sum_{i\in S} \mathbf{w}_i\cdot \x_i)+(\sum_{i\in \Bar{S}} \mathbf{w}_i\cdot (\neg\x_i))+b\geq 0\}}
\end{aligned}
\end{equation}

We can continue and show that the new relation holds the following:

\begin{equation}
\begin{aligned}
\sum_{S\subseteq [n], |S|=j}\mathbf{1}_{\{(\sum_{i\in S} \mathbf{w}_i\cdot \x_i)+(\sum_{i\in \Bar{S}} \mathbf{w}_i\cdot (\neg\x_i))+b\geq 0\}}=\\
\sum_{S\subseteq [n], |S|=j}\mathbf{1}_{\{(\sum_{i\in S,\x_i=1} \mathbf{w}_i)+(\sum_{i\in \Bar{S},\x_i=0} \mathbf{w}_i)+b\geq 0\}}
\end{aligned}
\end{equation}

We move to the second term and prove that the following holds:

\begin{equation}
\begin{aligned}
\sum_{S\subseteq [n], |S|= j}\mathbf{1}_{\{\sum_{i\in S} \mathbf{w}'_i\leq T+M\cdot j\}}=\\ 
\sum_{S\subseteq [n], |S|= j}\mathbf{1}_{\{\sum_{i\in S,\x_i=1} (-\mathbf{w}_i+M)+\sum_{i\in S,\x_i=0} (\mathbf{w}_i+M)\leq T+M\cdot j\}}=\\
\sum_{S\subseteq [n], |S|= j}\mathbf{1}_{\{\sum_{i\in S,\x_i=1} (-\mathbf{w}_i)+\sum_{i\in S,\x_i=0} \mathbf{w}_i\leq T\}}=\\
\sum_{S\subseteq [n], |S|= j}\mathbf{1}_{\{\sum_{i\in S,\x_i=1} (-\mathbf{w}_i)+\sum_{i\in S,\x_i=0} \mathbf{w}_i\leq b+(\sum_{i\in[n],\x_i=0} \mathbf{w}_i)\}}=\\
\sum_{S\subseteq [n], |S|= j}\mathbf{1}_{\{\sum_{i\in S,\x_i=1} (-\mathbf{w}_i)+\sum_{i\in S,\x_i=0} \mathbf{w}_i-\sum_{i\in S,\x_i=0} \mathbf{w}_i- \sum_{i\in \Bar{S},\x_i=0} \mathbf{w}_i\leq b\}}=\\
\sum_{S\subseteq [n], |S|= j}\mathbf{1}_{\{\sum_{i\in S,\x_i=1} \mathbf{w}_i+ \sum_{i\in \Bar{S},\x_i=0} \mathbf{w}_i+b\geq 0\}}
\end{aligned}
\end{equation}

Hence proving the equivalence.

$\qedsymbol{}$

We hence can equivalently derive in that $H_{\mathcal{D}_p}(S,k)$ is equivalent to:

\begin{equation}
\sum_{j=k}^{j=n}\binom{j}{k}\big(\sum_{|S|= j}\big( \prod\limits_{{i\in[n],(\x_S;\neg \x_{\Bar{S}})}_i=1} \big(p(i)\big) \prod\limits_{i\in[n],{(\x_S;\neg \x_{\Bar{S}})}_i=0} \big(1-p(i))\big)\big)\cdot\mathbf{1}_{\{\sum_{i\in S} \mathbf{w}'_i\leq T+M\cdot j\}}
\end{equation}




		
		


$\qedsymbol{}$

We now will conclude the proof by proving the following claim:

\begin{claim}
    $H_{\mathcal{D}_p}(S,k)$ can be computed in polynomial time.
\end{claim}

\emph{Proof.} We will demonstrate how $H_{\mathcal{D}_p}(S,k)$ can be computed in polynomial time. This will be done using a dynamic programming algorithm that computes a portion of this sum, utilizing the notation of $DP$. Specifically, for all $i\in\mathbb{N}$, $\ell\in\mathbb{N}_0$ such that $1\leq i-\ell\leq i\leq n$, and for all $C\in\mathbb{Z}$, we define:


\begin{equation}
    \begin{aligned}
        DP[i][C][i-\ell]:=\quad\quad\quad\quad\quad\quad\quad\quad\quad\quad\quad\quad\quad\\
        \begin{cases}
    \sum\limits_{S\subseteq[n], |S|\leq i-\ell}\big(\prod\limits_{{r\in[i],(\x_S;\neg \x_{\Bar{S}})}_r=1} \big(p(r)\big)\prod\limits_{{r\in[i],(\x_S;\neg \x_{\Bar{S}})}_r=0} \big(1-p(r)\big)\big) \quad if \ \ \sum_{i\in S} \mathbf{w}'_i\leq C \\
    0 \quad otherwise
\end{cases}
    \end{aligned}
\end{equation}

where, for any $i, \ell \in \mathbb{N}$ such that it does \emph{not} hold that $1 \leq \ell \leq i \leq n$, we define $DP[i][C][i-\ell] = 0$. Given some $1 \leq j \leq n$, if we take $i := n$, $C := T + Mj$, and $i - \ell := j$ (i.e., $\ell := i - j$), we find that $DP[n][T + Mj][j]$ is equal to:



\begin{equation}
    \begin{aligned}
\sum_{|S|\leq j}\big( \prod\limits_{{i\in[n],(\x_S;\neg \x_{\Bar{S}})}_i=1} \big(p(i)\big) \prod\limits_{i\in[n],{(\x_S;\neg \x_{\Bar{S}})}_i=0} \big(1-p(i))\big)\cdot\mathbf{1}_{\{\sum_{i\in S} \mathbf{w}'_i\leq T+M\cdot j\}}
    \end{aligned}
\end{equation}

Thus, we ultimately obtain that:

\begin{equation}
    \begin{aligned}
    DP[n][T+Mj][j] - DP[n][T+M(j-1)][j-1] =\quad\quad\quad\quad \\
\sum_{|S|= j}\big( \prod\limits_{{i\in[n],(\x_S;\neg \x_{\Bar{S}})}_i=1} \big(p(i)\big) \prod\limits_{i\in[n],{(\x_S;\neg \x_{\Bar{S}})}_i=0} \big(1-p(i))\big)\cdot\mathbf{1}_{\{\sum_{i\in S} \mathbf{w}'_i\leq T+M\cdot j\}}
    \end{aligned}
\end{equation}

Therefore, assuming that $DP[i][C][j]$ can be computed in polynomial time, we can iterate over all $j \leq k \leq n$ and compute $DP[i][C][j]$ for each $j$, thereby calculating $H_{\mathcal{D}_P}$ overall. To simplify the presentation, we define:

\begin{equation}
    \begin{aligned}
R(\x,S,i):=\big(\prod\limits_{{r\in[i],(\x_S;\neg \x_{\Bar{S}})}_r=1} \big(p(r)\big)\prod\limits_{{r\in[i],(\x_S;\neg \x_{\Bar{S}})}_r=0} \big(1-p(r)\big)\big)
    \end{aligned}
\end{equation}

Thus, we can simply express it as:

\begin{equation}
    \begin{aligned}
        DP[i][C][i-\ell]:=
        \begin{cases}
    \sum\limits_{S\subseteq[n], |S|\leq i-\ell}R(\x,S,i) \quad if \ \ \sum_{i\in S} \mathbf{w}'_i\leq C \\
    0 \quad otherwise
\end{cases}
    \end{aligned}
\end{equation}

For our final step, we will now prove the correctness of our dynamic programming algorithm:

\begin{lemma}
    There exists a polynomial dynamic programming algorithm that computes $DP[i][C][j]$ for any $C\in\mathbb{Z}$ and any $1\leq \ell\leq i\leq n$.
\end{lemma}

\emph{Proof.} We present the following inductive relation:

\begin{equation}
    \begin{aligned}
        DP[i+1][C][i+1-\ell]=\quad\quad\quad\quad\quad\quad\quad\quad\quad\\
        \begin{cases}
            DP[i][C][i-\ell]\cdot (1-p(i)) + DP[i][C-\mathbf{w}'_{i+1}][i-\ell]\cdot p(i) \ \ if \ \x_i=1 \\
            DP[i][C][i-\ell]\cdot p(i) + DP[i][C-\mathbf{w}'_{i+1}][i-\ell]\cdot (1-p(i)) \ \ if \ \x_i=0
        \end{cases}
    \end{aligned}
\end{equation}

We also define that $DP[i][C][j] = 0$ for any $i, C, j < 0$ and further define that when $i = 0$, then $DP[i][C][j] = 1$. We begin with the induction base, i.e., when $i = 1$. If $\ell \geq 1$, then we have $DP[1][C][1-\ell] = 0$, which satisfies our conditions. The only remaining case is when $\ell = 0$. In the case that $\x_i = 1$:



\begin{equation}
    \begin{aligned}
DP[1][C][1-\ell]=DP[1][C][1]= DP[0][C][0]\cdot (1-p(1)) + DP[0][C-s_1][0]\cdot p(1)=\\DP[0][C-s_1][0]\cdot p(1)
    \end{aligned}
\end{equation}

If it also holds that $s_1 \leq C$, we obtain:

\begin{equation}
    \begin{aligned}
DP[1][C][1-\ell]=DP[0][C-s_1][0]\cdot p(1)=p(1)=R(\x,\{s_1\},1)
    \end{aligned}
\end{equation}

However, if $s_1> C$ we get that:

\begin{equation}
    \begin{aligned}
DP[1][C][1-\ell]=DP[0][C-s_1][0]\cdot p(1)=0
    \end{aligned}
\end{equation}
As required. For the other scenario, we get that if $\x_i=0$ it holds that:

\begin{equation}
    \begin{aligned}
DP[1][C][1-\ell]=DP[1][C][1]= DP[0][C][0]\cdot (p(1)) + DP[0][C-s_1][0]\cdot p(1)=\\DP[0][C-s_1][0]\cdot (1-p(1))
    \end{aligned}
\end{equation}

If it additionally holds that $s_1\leq C$, then we have:

\begin{equation}
    \begin{aligned}
DP[1][C][1-\ell]=DP[0][C-s_1][0]\cdot (1-p(1))=(1-p(1))=R(\x,\{s_1\},1)
    \end{aligned}
\end{equation}

However, if $s_1> C$ we get that:

\begin{equation}
    \begin{aligned}
DP[1][C][1-\ell]=DP[0][C-s_1][0]\cdot (1-p(1))=0
    \end{aligned}
\end{equation}

as required. For the inductive step, we assume correctness holds for some $i \in \mathbb{N}$ and show that:

\begin{equation}
    \begin{aligned}
        DP[i+1][C][i+1-\ell]:=\\ 
        DP[i][C][j]\cdot p(i) + DP[i][C-\mathbf{w}'_{i+1}][j-1]\cdot (1-p(i)) = \\
    \sum\limits_{|S|\leq i-\ell, |\x_S|_1\leq C}R(k,\x,S,i)\cdot p(i+1) + \\
    \sum\limits_{|S| \leq i-\ell, |\x_S|_1\leq C-\mathbf{w}'_{i+1}}R(k,\x,S,i)\cdot (1-p(i+1))=\\
    \sum\limits_{|S| \leq i-\ell+1, |\x_S|_1\leq C, i+1\not\in S}R(k,\x,S,i)\cdot p(i+1) + \\
    \sum\limits_{|S| \leq i-\ell+1, |\x_S|_1\leq C, i+1\in S}R(k,\x,S,i)\cdot (1-p(i+1))=\\
    R(k,\x,S,i+1)
    \end{aligned}
\end{equation}

$\qedsymbol{}$

We conclude by presenting the complete algorithm described for computing $H_{\mathcal{D}_P}(S,k)$ with the following pseudocode:

\begin{algorithm}
    \newcommand{\algorithmicforeach}{\textbf{for each}}
\newcommand{\ForEach}[2]{\STATE \algorithmicforeach\ #1 \textbf{do} #2}
\newcommand{\Return}[1]{\STATE \textbf{return} #1}
\newcommand{\EndForEach}{\STATE \textbf{end for each}}

	\textbf{Input} $f$, $\x$, $S$, $k$
	\caption{Computing $H_{\mathcal{D}_P}(S,k)$}\label{alg:subset-minimal-local}
	\begin{algorithmic}[1]
 		\STATE $T\gets b+\big(\sum_{i\in[n],\x_i=0} \mathbf{w}_i\big)$
    	\STATE $H_{\mathcal{D}_P}\gets 0$
		\ForEach {$k\leq j \leq n$}\label{lst:line:orderingline}
    	\STATE $H_{\mathcal{D}_P}\gets H_{\mathcal{D}_P}+\binom{j}{k}\cdot(DP[n][T+Mj][j]-DP[n][T+M(j-1)][j-1])$
		\EndForEach
		\STATE \Return $H_{\mathcal{D}_P}$ 
	\end{algorithmic}
\end{algorithm}

The following proof demonstrates that solving the \emph{SHAP} query for Perceptrons can be done in polynomial time, assuming the weights and biases are provided in unary.

\section{Proof of Proposition~\ref{no_gap_original_paper_prop}}
\label{no_gap_in_complex_models_appendix_sec}

\begin{proposition}
\label{no_gap_in_complex_models_appendix}
There is no explainability query $Q$ for which the class of MLPs is strictly more c-interpretable than the class of ensemble-MLPs.
\end{proposition}

\emph{Proof.} We will specifically prove this claim for a much larger set of functions that we will refer to as models that are \emph{closed under ensemble construction}:

\begin{definition}
We say that a class of models $\mathcal{C}$ is \textbf{closed under ensemble construction} if given an ensemble $f$ containing models from $\mathcal{C}$, we can construct in polynomial time a model $g\in \mathcal{C}$ for which $\forall\x\in\mathbb{F}, f(\x)=g(\x)$.
\end{definition}

We first note that from the definition of a model that is closed under ensemble construction, the following claim holds: Let there be a class of models $\mathcal{C}$, which is closed under ensemble construction. Let us denote $\mathcal{C}_E$ as the class of ensemble models that consist of base-models from class $\mathcal{C}$. Then for any explainability query $Q$: then it holds that $Q(\mathcal{C}_E)\leq_P Q(\mathcal{C})$. Now, if $Q(\mathcal{C})$ belongs to a complexity class that is closed under polynomial reductions, it hence must hold that $Q(\mathcal{C}_E)$ does not belong to a strictly harder complexity class. We are now only left to prove the following claim:

\begin{claim}
    Let $\mathcal{C}_{MLP}$ denote the class of models that are represented by MLPs. Then $\mathcal{C}_{MLP}$ is closed under ensemble construction.
\end{claim}

\emph{Proof.} Consider a model $f$ comprising $k$ individual MLPs denoted as $f_1,\ldots,f_k$ We can devise a new MLP, $f'$, by integrating the hidden layers from each $f_i$. Specifically, $f'$'s first hidden layer is constructed by concatenating the first hidden layers of $f_1,\ldots,f_k$ , and similarly for subsequent layers up to the highest number of layers present in any $f_i$. For models with fewer layers, we introduce ``dummy'' layers equipped with weights of $1$ and biases of $0$, effectively passing their last actual output through unchanged. In this initial setup, the layers in $f'$ corresponding to each $f_i$ are only linked to their respective preceding layers within the same $f_i$, thus lacking full connectivity across the different models $f_j$ such that $j\neq i$. To amend this, connectivity can be enhanced by adding inter-layer connections with weights of $1$ and biases of $0$, ensuring each layer does not influence the next across the different sub-models.

Finally, we observe that the constructed $f'$ outputs $k$ distinct values corresponding to the outputs of each model $f_1,\ldots, f_k$ (the output value prior to the step function). We need to introduce a new output layer to $f'$ to implement a majority voting mechanism. This is conceptualized as a weighted voting process where each model $f_i$ is assigned a specific weight $\phi_i$. This can be realized by adding a fully connected layer that consolidates the $k$ outputs into a single output in the final layer of $f'$. In this layer, each output $i\in\{1,\ldots,k\}$ is assigned a weight $w_i:=\phi_i$, and we set the bias of this last layer to $0$. Consequently, applying a step function over $f'$ results in an output that represents a weighted majority vote of the ensemble $f$. Additionally, a non-weighted majority vote can be modeled in the setting where all weights $\phi_1=\phi_2=\ldots =\phi_k$ are equal. We thus establish that $f'\in\mathcal{C}_{MLP}$, and that for all $\z\in\mathbb{F}$, $f(\z)=f'(\z)$ thereby confirming that $\mathcal{C}_{MLP}$ is closed under ensemble construction.

$\qedsymbol{}$

\section{Proof of Proposition~\ref{paramaterized_model_size}}
\label{paramaterized_model_size_appendix_section}

\begin{proposition}
    \label{paramaterized_model_size_appendix}
    An ensemble consisting of either Perceptrons, decision trees, or MLPs, parameterized by the maximal base-model size is \begin{inparaenum}[(i)]
\item para-coNP Complete with respect to CSR, \item para-NP-Complete with respect to MCR, \item para-$\Sigma^P_2$-Complete with respect to MSR, and \item para-$\#P$-Complete with respect to CC, \item para-$\#P$-Hard with respect to SHAP. 
\end{inparaenum}
\end{proposition}


\emph{Proof.} Membership is straightforward from the definition of para-$\mathcal{K}$ and the completeness of all the non-paramaterized versions to each corresponding complexity class $\mathcal{K}$, as proven in Proposition~\ref{not_paramaterized_results_appendix}. For example, the non-parameterized version of \emph{CSR} for an ensemble of Perceptrons, decision trees, or MLPs is coNP-Complete. The same holds for the other complexity classes.

\textbf{Hardness.} The reduction for the \emph{CSR} query was provided as a direct reduction from the \emph{TAUT} problem. Since \emph{TAUT} is hard when restricted to \emph{3-DNF}s as well, then hardness for an ensemble consisting of \emph{constant} sized-base lines is straightforward (for any $k\geq 3$). In the case of the \emph{MCR} reduction, the result already holds for any $k=2$ since the reductino inherently produces ensembles consisting of input dimensions of at most $2$. For the \emph{CC} query.

Finally, in the case of \emph{MSR}, we note that the Shortest Implicant Core problem in its classic form presented by~\cite{umans2001minimum} describes general DNFs and not restricted DNFs. However, in a consequent work~\cite{dick2009improved}, it was proven that the Shortest implicant core problem for DNFs with \emph{constant} term size is also $\Sigma^P_2$-Hard. 

$\qedsymbol{}$

\section{Proof of Proposition~\ref{k_ensemble_perceptrons_hardness_paper}}
\label{linear_model_para_hardness_results_appendix_section}

\begin{proposition}
	\label{linear_model_para_hardness_results_appendix}
	$k$-ensemble-Perceptrons are
	\begin{inparaenum}[(i)]
		\item para-coNP Complete with respect to CSR, \item para-NP-Complete 
		with respect to MCR, \item para-$\Sigma^P_2$-Complete with respect to 
		MSR, and \item para-$\#P$-Hard with respect to CC and SHAP. 
	\end{inparaenum}
\end{proposition}

\emph{Proof.} All \emph{membership} results are a direct result from the 
non-parameterized complexity results~\ref{not_paramaterized_results_appendix}, 
and from the reasons described under the proof of 
Proposition~\ref{paramaterized_model_size_appendix}. We now will prove each 
hardness result seperattley.

\begin{lemma}
	The CSR query for a $k$-ensemble of Perceptrons is para-coNP Hard.
\end{lemma}

\emph{Proof.} We will equivalently prove that the \emph{CSR} query for an 
ensemble containing only $2$ perceptrons is coNP Complete. We will reduce from 
the \emph{Subset-sum problem} (\emph{SSP}), which is a classic NP-Complete 
problem. 

\vspace{0.5em} 

\noindent\fbox{%
	\parbox{\columnwidth}{%
		\mysubsection{SSP (Subset Sum)}:
		
		\textbf{Input}: $(z_1,z_2,\ldots,z_n)$ set of positive integers and a 
		positive (target) integer $T$
		
		\textbf{Output}: \yes{}, if there exists a subset $S\subseteq 
		(1,2,\ldots,n)$ such that $\sum_{i\in S}z_i=T$, and \no{} otherwise.
	}%
}

\vspace{0.5em} 
We reduce $CSR$ for an ensemble with $k=2$ of Perceptrons from 
$\overline{SSP}$. 
Given some $\langle (z_1,z_2,\ldots,z_n),T\rangle$, the reduction constructs 
the two following Perceptrons $f_1:=\langle \mathbf{w^1},b_1\rangle$ and 
$f_2:=\langle \mathbf{w^2},b_2\rangle$, where 
$\mathbf{w^1}:=(-z_1,-z_2,\ldots,-z_n)$, $b_1:=T-\frac{1}{2}$, 
$\mathbf{w^2}:=(z_1,z_2,\ldots,z_n)$, and $b_2:=-T-\frac{1}{2}$. The reduction 
constructs $\langle f:=(f_1,f_2),\x:=\mathbf{0}_n,S:=\emptyset\rangle$, and 
$\mathbf{0}_n$ denotes a vector of size $n$ where all values are set to $0$.

First, we notice that:

\begin{equation}
	\begin{aligned}
		f_1(\x)=f_1(\mathbf{0}_n) = T-\frac{1}{2}>0 \ \wedge \\
		f_2(\x)=f_2(\mathbf{0}_n) = -T-\frac{1}{2}<0
	\end{aligned}
\end{equation}

Since $f_1(\x)$ is positive and $f_2(\x)$ is negative, then the ensemble 
$f=\langle f_1,f_2\rangle$ returns 1 for the input $\x$ (is positive), by 
definition.

If $\langle (z_1,z_2,\ldots,z_n), T\rangle\in \overline{SSP}$ then there does 
not exist a subset of features $S'\subseteq (1,2,\ldots,n)$ such that 
$\sum_{i\in S'}z_i=T$. Since these are \emph{integers}, this implies that any 
subset $S'\subseteq (1,2,\ldots,n)$ is either equal or greater than $T+1$ or 
equal and smaller than $T-1$. Let us mark $S'$ as some subset for which it 
holds that $\sum_{i\in S'}z_i\geq T+1$. Let us denote $\mathbf{x'} = 
(\mathbf{1}_{S'};\mathbf{0}_{\Bar{S'}})$. Or in other words $\mathbf{x'}$ 
denotes a vector where all the values in $S'$ are set to $1$ and the rest to 
$0$. It clearly holds that:

\begin{equation}
	\begin{aligned}
		f_1(\mathbf{x'})=\sum_{i\in \Bar{S'}}\mathbf{x'}_i\cdot \mathbf{w^1_i} 
		+b_1 = \\
		-\sum_{i\in S'}z_i +b_1 \leq \\
		-(T+1) + T-\frac{1}{2} = -\frac{3}{2}<0
	\end{aligned}
\end{equation}

It also holds that:

\begin{equation}
	\begin{aligned}
		f_2(\mathbf{x'})=\sum_{i\in \Bar{S'}}\mathbf{x'}_i\cdot \mathbf{w^2_i} 
		+b_2 = \\
		\sum_{i\in S'}z_i +b_2 \geq \\
		(T+1) + (-T-\frac{1}{2}) = \frac{1}{2}>0
	\end{aligned}
\end{equation}

This implies $f_1(\mathbf{x'})$ is negative and $f_2(\mathbf{x'})$ is positive, 
and overall we get that for the ensemble $f$ it holds that $f(\mathbf{x'})$ is 
positive.

Now, let us assume that $\sum_{i\in S'}z_i\leq T-1$. Again, let us denote 
$\mathbf{x'} = (\mathbf{1}_{S'};\mathbf{0}_{\Bar{S'}})$. It clearly holds that:

\begin{equation}
	\begin{aligned}
		f_1(\mathbf{x'})=\sum_{i\in \Bar{S'}}\mathbf{x'}_i\cdot \mathbf{w^1_i} 
		+b_1 = \\
		-\sum_{i\in S'}z_i +b_1 \geq \\
		-(T-1) + T-\frac{1}{2} = \frac{1}{2} > 0
	\end{aligned}
\end{equation}

It also holds that:

\begin{equation}
	\begin{aligned}
		f_2(\mathbf{x'})=\sum_{i\in \Bar{S'}}\mathbf{x'}_i\cdot \mathbf{w^2_i} 
		+b_2 = \\
		\sum_{i\in S'}z_i +b_2 \leq \\
		(T-1) + (-T-\frac{1}{2}) = -\frac{3}{2}<0
	\end{aligned}
\end{equation}

Implying, that under this scenario the opposite case occurs: $f_1(\mathbf{x'})$ 
is positive and $f_2(\mathbf{x'})$ is negative, and overall we get again that 
for the ensemble $f$ it holds that $f(\mathbf{x'})$ is positive.

This shows, that for any feasible value of $\x'$, $f(\mathbf{x'})$ is positive, 
and this implies that $S=\emptyset$ is a sufficient reason of $\langle 
f,\x=\mathbf{0}_n\rangle$.

If $\langle (z_1,z_2,\ldots,z_n), T\rangle\not\in \overline{SSP}$, this means 
that there does exist a subset $S'\subseteq \{1,\ldots,n\}$ for which 
$\sum_{i\in S'}z_i=T$. We again denote $\mathbf{x'} = 
(\mathbf{1}_{S'};\mathbf{0}_{\Bar{S'}})$. It holds that:

\begin{equation}
	\begin{aligned}
		f_1(\mathbf{x'})=\sum_{i\in \Bar{S'}}\mathbf{x'}_i\cdot \mathbf{w^1_i} 
		+b_1 = \\
		-\sum_{i\in S'}z_i +b_1 = \\
		-(T) + T-\frac{1}{2} = -\frac{1}{2} < 0
	\end{aligned}
\end{equation}

It also holds that:

\begin{equation}
	\begin{aligned}
		f_2(\mathbf{x'})=\sum_{i\in \Bar{S'}}\mathbf{x'}_i\cdot \mathbf{w^2_i} 
		+b_2 = \\
		\sum_{i\in S'}z_i +b_2 = \\
		(T) + (-T-\frac{1}{2}) = -\frac{1}{2}<0
	\end{aligned}
\end{equation}

This means that both $f_1(\mathbf{x'})$ and $f_2(\mathbf{x'})$ are negative and 
hence the ensemble $f(\mathbf{x'})$ is also negative. Since 
$f(\x)=f(\mathbf{0}_n)$ is positive, it holds that $S=\emptyset$ is not a 
sufficient reason of $\langle f,\x\rangle$, which concludes the reduction. 

$\qedsymbol{}$

\begin{lemma}
	The MCR query for a $k$-ensemble of Perceptrons is para-NP Hard. 
\end{lemma}

We will equivalently show that the \emph{MCR} query for an ensemble of k=2 
Perceptrons is NP-Complete. We use a refined version of the SSP problem --- 
\emph{kSSP}, which is also known to be NP-Complete.

\vspace{0.5em} 

\noindent\fbox{%
	\parbox{\columnwidth}{%
		\mysubsection{kSSP (k Subset Sum)}:
		
		\textbf{Input}: $(z_1,z_2,\ldots,z_n)$ set of positive integers, a 
		positive integer $k$, and a positive (target) integer $T$.
		
		\textbf{Output}: \yes{}, if there exists a subset $S\subseteq 
		(1,2,\ldots,n)$ such that $|S|=k$ and $\sum_{i\in S}z_i=T$, and \no{} 
		otherwise.
	}%
}

\vspace{0.5em} 


We reduce $MCR$ for an ensemble with $2$ Perceptrons from $kSSP$. 
Given some $\langle (z_1,z_2,\ldots,z_n),k,T\rangle$, the reduction constructs 
the two following Perceptrons $f_1:=\langle \mathbf{w^1},b_1\rangle$ and 
$f_2:=\langle \mathbf{w^2},b_2\rangle$, where 
$\mathbf{w^1}:=(-z_1,-z_2,\ldots,-z_n)$, $b_1:=T-\frac{1}{2}$, 
$\mathbf{w^2}:=(z_1,z_2,\ldots,z_n)$, and $b_2:=-T-\frac{1}{2}$. The reduction 
constructs $\langle f:=(f_1,f_2),\x:=\mathbf{0}_n,k\rangle$, and $\mathbf{0}_n$ 
denotes a vector of size $n$ where all values are set to $0$.

First, we notice that:

\begin{equation}
	\begin{aligned}
		f_1(\x)=f_1(\mathbf{0}_n) = T-\frac{1}{2}>0 \ \wedge \\
		f_2(\x)=f_2(\mathbf{0}_n) = -T-\frac{1}{2}<0
	\end{aligned}
\end{equation}

Since $f_1(\x)$ is positive and $f_2(\x)$ is negative, then the ensemble 
$f=\langle f_1,f_2\rangle$ returns 1 for the input $\x$ (is positive), by 
definition.

If $\langle (z_1,z_2,\ldots,z_n),k, T\rangle\not\in SSP$ then there does not 
exist a subset of features $S'\subseteq (1,2,\ldots,n)$ such that $\sum_{i\in 
S'}z_i=T$. Since these are \emph{integers}, this implies that any subset 
$S'\subseteq (1,2,\ldots,n)$ \emph{of size $k$} is either equal or greater than 
$T+1$ or equal and smaller than $T-1$. Let us mark $S'$ as some subset of size 
$k$ for which it holds that $\sum_{i\in S'}z_i\geq T+1$. Let us denote 
$\mathbf{x'} = (\mathbf{1}_{S'};\mathbf{0}_{\Bar{S'}})$. Or in other words 
$\mathbf{x'}$ denotes a vector where all the values in $S'$ are set to $1$ and 
the rest to $0$. It clearly holds that:

\begin{equation}
	\begin{aligned}
		f_1(\mathbf{x'})=\sum_{i\in \Bar{S'}}\mathbf{x'}_i\cdot \mathbf{w^1_i} 
		+b_1 = \\
		-\sum_{i\in S'}z_i +b_1 \leq \\
		-(T+1) + T-\frac{1}{2} = -\frac{3}{2}<0
	\end{aligned}
\end{equation}

It also holds that:

\begin{equation}
	\begin{aligned}
		f_2(\mathbf{x'})=\sum_{i\in \Bar{S'}}\mathbf{x'}_i\cdot \mathbf{w^2_i} 
		+b_2 = \\
		\sum_{i\in S'}z_i +b_2 \geq \\
		(T+1) + (-T-\frac{1}{2}) = \frac{1}{2}>0
	\end{aligned}
\end{equation}

This implies $f_1(\mathbf{x'})$ is negative and $f_2(\mathbf{x'})$ is positive, 
and overall we get that for the ensemble $f$ it holds that $f(\mathbf{x'})$ is 
positive.

Now, let us assume that $\sum_{i\in S'}z_i\leq T-1$. Again, let us denote 
$\mathbf{x'} = (\mathbf{1}_{S'};\mathbf{0}_{\Bar{S'}})$. It clearly holds that:

\begin{equation}
	\begin{aligned}
		f_1(\mathbf{x'})=\sum_{i\in \Bar{S'}}\mathbf{x'}_i\cdot \mathbf{w^1_i} 
		+b_1 = \\
		-\sum_{i\in S'}z_i +b_1 \geq \\
		-(T-1) + T-\frac{1}{2} = \frac{1}{2} > 0
	\end{aligned}
\end{equation}

It also holds that:

\begin{equation}
	\begin{aligned}
		f_2(\mathbf{x'})=\sum_{i\in \Bar{S'}}\mathbf{x'}_i\cdot \mathbf{w^2_i} 
		+b_2 = \\
		\sum_{i\in S'}z_i +b_2 \leq \\
		(T-1) + (-T-\frac{1}{2}) = -\frac{3}{2}<0
	\end{aligned}
\end{equation}

Implying, that under this scenario the opposite case occurs: $f_1(\mathbf{x'})$ 
is positive and $f_2(\mathbf{x'})$ is negative, and overall we get again that 
for the ensemble $f$ it holds that $f(\mathbf{x'})$ is positive.

This shows that for any subset $S'$ of size $k$ the value of $f(\mathbf{x'})$ 
remains positive. Since the value of $f(\mathbf{x})$ is also positive, this 
implies that there is no contrastive reason $S'$ of size $k$ with respect to 
$\langle f,\x\rangle$. This also clearly implies that there is no contrastive 
reason $S'$ of size smaller than $k$ with respect to $\langle f,\x\rangle$.

If $\langle (z_1,z_2,\ldots,z_n), T\rangle\in SSP$, this means that there does 
exist a subset $S'\subseteq \{1,\ldots,n\}$ of size $k$ for which $\sum_{i\in 
S'}z_i=T$. We again denote $\mathbf{x'} = 
(\mathbf{1}_{S'};\mathbf{0}_{\Bar{S'}})$. It holds that:

\begin{equation}
	\begin{aligned}
		f_1(\mathbf{x'})=\sum_{i\in \Bar{S'}}\mathbf{x'}_i\cdot \mathbf{w^1_i} 
		+b_1 = \\
		-\sum_{i\in S'}z_i +b_1 = \\
		-(T) + T-\frac{1}{2} = -\frac{1}{2} < 0
	\end{aligned}
\end{equation}

It also holds that:

\begin{equation}
	\begin{aligned}
		f_2(\mathbf{x'})=\sum_{i\in \Bar{S'}}\mathbf{x'}_i\cdot \mathbf{w^2_i} 
		+b_2 = \\
		\sum_{i\in S'}z_i +b_2 = \\
		(T) + (-T-\frac{1}{2}) = -\frac{1}{2}<0
	\end{aligned}
\end{equation}

This means that both $f_1(\mathbf{x'})$ and $f_2(\mathbf{x'})$ are negative and 
hence the ensemble $f(\mathbf{x'})$ is also negative. Since 
$f(\x)=f(\mathbf{0}_n)$ is positive, it holds that there exists a contrastive 
reason ($S'$) of size $k$ with respect to $\langle f,\x\rangle$, concluding the 
reduction.

$\qedsymbol{}$

\begin{lemma}
	The CC and SHAP queries are para-$\#P$-Hard for a $k$ ensemble of 
	Perceptrons.
\end{lemma}

\emph{Proof.} The Hardness of the \emph{CC} query follows from tbe fact that 
this problem is already $\#P$-Hard for a single perceptron~\cite{BaMoPeSu20} 
($k=1$). For the \emph{SHAP} query, we follow the proof suggested 
by~\cite{arenas2023complexity} who showed a connection between the exact 
computation of shapley values computations and the model counting problem. 
Given some model $f$, we define $\#f$ as the number of assignments which output 
$1$. More formally: $\#f:|\{\z \ \ | \ \ f(\z)=1\}|$. 
\cite{arenas2023complexity} showed the following connection for the specific 
case where $\mathcal{D}_p$ is taken to be the uniform distribution. It holds 
that for \emph{all} $\x\in\mathbb{F}$ it holds that:

\begin{equation}
	\#f = 2^n\big(f(\x)-\sum_{i\in[n]}\phi_i(f,\x)\big)
\end{equation}

where the following relation is a consequence of the well-known 
\emph{efficiency} property of shapley values. This result establishes a direct 
reduction from the \emph{SHAP} query to the model counting problem. Moreover, 
the model counting problem is simply a private case of the \emph{CC} query 
where $S:=\emptyset$. This establishes a polynomial reduction from the 
\emph{CC} to the \emph{SHAP} query. We hence conclude that \emph{SHAP} is also 
$\#P$-Hard for an ensemble consisting of only a single Perceptron.


$\qedsymbol{}$

\begin{lemma}
	\label{MSR-k-perceptron-S2Pcomplete, berman2002complexity}
	The MSR query for a $k$-ensemble of Perceptrons is \StoPComplexity-Complete.
\end{lemma}

\emph{Proof.} \textbf{Membership.}
We establish membership in para-$\Sigma^P_2$ using the direct definition of the complexity class since we can non-deterministically guess a subset $S$ and then utilize a coNP oracle to verify whether $S$ is sufficient.

\textbf{Hardness.} We will equivalently show that the \emph{MSR} query for an ensemble of $k=5$ 
Perceptrons is \StoPComplexity-Hard. First, we will present the 
\emph{Generalized Subset Sum} problem, which is known to be 
\StoPCompleteComplexity-complete~\cite{schaefer2002completeness}.

\vspace{0.5em} 

\noindent\fbox{%
	\parbox{\columnwidth}{%
		\mysubsection{GSSP (Generalized Subset Sum)}:
		
		\textbf{Input}: Two vectors of positive integers 
		$\ub=(\ub_1,\ub_2,\ldots,\ub_l)$ and 
		$\vb=(\vb_1,\vb_2,\ldots,\vb_m)$, 
		and a positive (target) integer $T$.
		
		\textbf{Output}: \yes{}, if there \emph{exists} a binary vector 
		$\x\in\{0,1\}^l$ such that for \emph{any} binary vector 
		$\y\in\{0,1\}^m$, it holds that: 
		$\Sigma_{i=1}^{l}(\x_{i}\cdot \ub_{i}) + \Sigma_{j=1}^{m}(\y_{j}\cdot 
		\vb_{j}) \neq T$; and \no{} otherwise.
	}%
}

\vspace{0.5em} 

We will actually use a very close modified version of this problem which we 
term \emph{k-Generalized Subset Sum}, which requires an additional constraint 
that the size of the vector $\x$ is equal to some input integer $k$. More 
formally, the problem is defined as follows:

\vspace{0.5em} 

\noindent\fbox{%
	\parbox{\columnwidth}{%
		\mysubsection{$k$-GSSP ($k$-Generalized Subset Sum)}:
		
		\textbf{Input}: Two vectors of positive integers 
		$\ub=(\ub_1,\ub_2,\ldots,\ub_l)$ and 
		$\vb=(\vb_1,\vb_2,\ldots,\vb_m)$, 
		an integer $k$,
		and a positive (target) integer $T$.
		
		\textbf{Output}: \yes{}, if there \emph{exists} a binary vector 
		$\x\in\{0,1\}^l$ such that
		$\left\lVert \x \right\rVert_{1}= k$,
		and for \emph{any} binary vector $\y\in\{0,1\}^m$, it holds 
		that: 
		$\Sigma_{i=1}^{l}(\x_{i}\cdot \ub_{i}) + \Sigma_{j=1}^{m}(\y_{j}\cdot 
		\vb_{j}) \neq T$; and
		\no{} otherwise.
	}%
}

\vspace{0.5em} 


We next prove the following claim:

\begin{claim}
	\label{lemma:K-Gssp}
	The query $\kGSSP$ is \StoPCompleteComplexity-hard.
\end{claim}


\emph{Proof.} We present a polynomial-time reduction from $\GSSP$ to $\kGSSP$, enabling us to conclude that $\kGSSP$ is \StoPComplexity-hard. Given $\langle \ub=(\ub_1,\ub_2,\ldots,\ub_l), \vb=(\vb_1,\vb_2,\ldots,\vb_m), T\rangle$, we define some $G > 0$. The reduction then constructs $\langle \ub'=(\ub_1+G,\ub_2+G,\ldots,\ub_l+G,G_{1},\ldots,G_{l}), \textbf{v'}\coloneqq\textbf{v}=(\vb_1,\vb_2,\ldots,\vb_m), k'\coloneqq l, T'\coloneqq T + (l\cdot G) \rangle$ (where $G_{i}$ is the $i$-th occurrence of the value $G$).

First, assume that $\langle \ub, \vb, T\rangle \in \GSSP$. By construction, this implies that:


\begin{equation}
	\begin{aligned}
		\exists \x\in \{0,1\}^l \quad 
		\quad \forall \y\in \{0,1\}^m 
		\quad 
		\Sigma_{i=1}^{l}(\x_{i}\cdot \ub_{i}) + \Sigma_{j=1}^{m}(\y_{j}\cdot \vb_{j}) 
		\neq T
	\end{aligned}
\end{equation}

Let $\x$ be an element of $\{0,1\}^l$. Therefore, for this $\x$, the following holds:

\begin{equation}
	\label{Eq:firstDirectionX}
	\begin{aligned}
		\forall \y\in \{0,1\}^m 
		\quad 
		\Sigma_{i=1}^{l}(\x_{i}\cdot \ub_{i}) + \Sigma_{j=1}^{m}(\y_{j}\cdot 
		\vb_{j}) 
		\neq T 
	\end{aligned}
\end{equation}

We express: $\left\lVert \x \right\rVert_{1} = r \leq l$. Next, we divide the $l$ summations into two parts: the first sum will include the corresponding values in $\ub$ with non-zero coordinates in $\x$, and the other part will include the remaining values in $\ub$. For simplicity and w.l.o.g., we also reindex the coordinates accordingly so that, w.l.o.g., all $r$ nonzero coordinates are the first ones of $\x$ (this is w.l.o.g. since we can reorder the corresponding coordinates in $\ub$ accordingly). Next, given this $\x$, we can construct a new binary vector $\x' \in \{0,1\}^{2l}$ where the first $l$ coordinates are identical to the given vector $\x$, and the remaining $l$ coordinates are constructed as follows: the first $(l-r)$ coordinates of the second half, i.e., coordinates $(l+1)$ to $(2l-r)$ of $\x'$, will be ``1'', while the remaining coordinates of $\x'$, i.e., from $(2l-r+1)$ to $(2l)$, will be ``0''.

\begin{equation}
	\begin{aligned}	
		\Sigma_{i=1}^{2l}(\x'_{i}\cdot \ub'_{i}) = \\
		\Sigma_{i=1}^{r}(\x'_{i}\cdot \ub'_{i}) +
		\Sigma_{i=r+1}^{l}(\x'_{i}\cdot \ub'_{i}) +
		\Sigma_{i=l+1}^{2l-r}(\x'_{i}\cdot \ub'_{i}) +
		\Sigma_{i=2l-r+1}^{2l}(\x'_{i}\cdot \ub'_{i})=\\
  		\Sigma_{i=1}^{r}(\x_{i}\cdot (\ub_{i}+G)) +
		\Sigma_{i=r+1}^{l}(0\cdot (\ub_{i}+G)) +
		\Sigma_{i=l+1}^{2l-r}(1\cdot G) +
		\Sigma_{i=2l-r+1}^{2l}(0\cdot G) =\\
  		\Sigma_{i=1}^{r}(\x_{i}\cdot \ub_{i}) 
		+ (r\cdot G)   
		+ ((l-r)\cdot G)   
	\end{aligned}
\end{equation}





And as (w.l.o.g.) the \emph{last} $(l-r)$ coordinates for $\x$ are ``0'', it follows that for this constructed $\x'$, we can sum up to $l$ (and not ``only'' $r$):

\begin{equation}
	\label{eq:xTagResult}
	\begin{aligned}	
		\Sigma_{i=1}^{2l}(\x'_{i}\cdot \ub'_{i}) =		
		\Sigma_{i=1}^{l}(\x_{i}\cdot \ub_{i}) + (l\cdot G)
	\end{aligned}
\end{equation}

Now, suppose, by contradiction, that for this constructed $\x'$, it holds that:

\begin{equation}
	\begin{aligned}	
		\exists \y\in\{0,1\}^m \quad 
		\Sigma_{i=1}^{2l}(\x'_{i}\cdot \ub'_{i}) +
		\Sigma_{j=1}^{m}(\y_{j}\cdot \vb_{j})
		= T'
	\end{aligned}
\end{equation}

Then, according to Eq.~\ref{eq:xTagResult}, it follows that:

\begin{equation}
	\begin{aligned}	
		\exists \y\in\{0,1\}^m \quad 
		\Sigma_{i=1}^{l}(\x_{i}\cdot \ub_{i}) + (l\cdot G) +
		\Sigma_{j=1}^{m}(\y_{j}\cdot \vb_{j})
		= T' 
	\end{aligned}
\end{equation}

And from the definition $T'\coloneqq T + (l\cdot G)$, we infer from our assumption that:

\begin{equation}
	\begin{aligned}	
		\exists y\in\{0,1\}^m \quad 
		\Sigma_{i=1}^{l}(\x_{i}\cdot \ub_{i}) + (l\cdot G) +
		\Sigma_{j=1}^{m}(\y_{j}\cdot \vb_{j})
		= T + (l\cdot G) \iff \\
  		\exists \y\in\{0,1\}^m \quad 
		\Sigma_{i=1}^{l}(\x_{i}\cdot \ub_{i}) +
		\Sigma_{j=1}^{m}(\y_{j}\cdot \vb_{j})
		= T 
	\end{aligned}
\end{equation}



However, this contradicts the outcome from Eq.~\ref{Eq:firstDirectionX} concerning this specific $\x$. Therefore, our assumption is incorrect, and thus for the constructed $\x'\in\{0,1\}^{2l}$, with $\left\lVert \x' \right\rVert_{1}= l = k'$, it follows that:


\begin{equation}
	\begin{aligned}	
		\forall \y\in\{0,1\}^m \quad 
		\Sigma_{i=1}^{2l}(\x'_{i}\cdot \ub'_{i}) +
		\Sigma_{j=1}^{m}(\y_{j}\cdot \vb_{j})
		\neq T'
	\end{aligned}
\end{equation}

Therefore, we can conclude that: $\langle \ub', \vb', k', T'\rangle \in \kGSSP$.

Now, suppose $\langle \textbf{u}, 
\textbf{v}, T\rangle  \notin \GSSP$. This implies that $\forall \x\in\{0,1\}^{l}$:

\begin{equation}
	\label{eq:lemma32SecondDirection}
	\begin{aligned}
		\exists \y\in \{0,1\}^m \quad	
		\Sigma_{i=1}^{l}(\x_{i}\cdot \ub_{i}) + 
		\Sigma_{j=1}^{m}(\y_{j}\cdot \vb_{j}) 
		= T
	\end{aligned}
\end{equation}

We need to prove that this implies the following:

\begin{equation}
	\begin{aligned}	
		\forall \x'\in\{0,1\}^{2l} \quad \text{s.t} \quad \left\lVert \x' \right\rVert_{1} = k', \\ 
	\exists \y\in \{0,1\}^m, \quad
		\Sigma_{i=1}^{2l}(\x'_{i}\cdot \ub'_{i}) + 
		\Sigma_{j=1}^{m}(\y_{j}\cdot \vb'_{j}) 
		= T'
	\end{aligned}
\end{equation}

Let us assume, for the sake of contradiction, that this is not the case, i.e.:

\begin{equation}
	\begin{aligned}	
		\exists \x'\in 
		\{0,1\}^{2l} \quad \text{s.t} \quad 
		\left\lVert \x' \right\rVert_{1} = k', \\
		 \forall \y\in \{0,1\}^m  \quad
		\Sigma_{i=1}^{2l}(\x'_{i}\cdot \ub'_{i}) + 
		\Sigma_{j=1}^{m}(\y_{j}\cdot \vb'_{j}) 
		\neq T'	\end{aligned}
\end{equation}

Let $\x'\in \{0,1\}^{2l}$ represent a binary input $\x'\in \{0,1\}^{2l}$. Therefore, $\left\lVert \x' \right\rVert_{1} = k'\coloneqq l$. Assume $1\leq s \leq l$ corresponds to the number of ``1'' entries in $\x'$ located in the first $l$ coordinates of \textbf{u'}, with the remaining $(k'-s)=(l-s)$ ``1'' entries of $\x'$ located in the second half of \textbf{u'}. Further, assume without loss of generality that the \emph{first} $s$ coordinates of $\x'_{[1\ldots l]}$ are ``1'', and similarly, without loss of generality, that the \emph{first} $(l-s)$ coordinates of $\x'_{[(l+1)\ldots 2l]}$ (i.e., the second half of $\x'$'s $l$ coordinates) represent the locations of the remaining $(l-s)$ ``1'' values. Therefore, we establish that for this $\x'$:



\begin{equation}
	\begin{aligned}	
		\forall \y\in \{0,1\}^m,  \quad
		\Sigma_{i=1}^{s}(\x'_{i}\cdot \ub'_{i}) + 
		\Sigma_{i=s+1}^{l}(\x'_{i}\cdot \ub'_{i}) +
		\Sigma_{i=l+1}^{2l-s}(\x'_{i}\cdot \ub'_{i}) +\\
		\Sigma_{i=2l-s+1}^{2l}(\x'_{i}\cdot u'_{i}) +
		\Sigma_{j=1}^{m}(\y_{j}\cdot \vb'_{j}) 
		\neq T' \iff \\
  		\forall \y\in \{0,1\}^m,  \quad
		\Sigma_{i=1}^{s}(\x'_{i}\cdot \ub'_{i}) + 
		\Sigma_{i=s+1}^{l}(0\cdot \ub'_{i}) +
		\Sigma_{i=l+1}^{2l-s}(\x'_{i}\cdot \ub'_{i}) +\\
		\Sigma_{i=2l-s+1}^{2l}(0\cdot \ub'_{i}) +
		\Sigma_{j=1}^{m}(\y_{j}\cdot \vb'_{j}) 
		\neq T'	\iff \\
  		\forall \y\in \{0,1\}^m,  \quad
		\Sigma_{i=1}^{s}(\x'_{i}\cdot \ub'_{i}) + 
		\Sigma_{i=l+1}^{2l-s}(\x'_{i}\cdot \ub'_{i}) +
		\Sigma_{j=1}^{m}(\y_{j}\cdot \vb'_{j}) 
		\neq T'	\iff \\
  		\forall \y\in \{0,1\}^m,  \quad
		\Sigma_{i=1}^{s}(\x'_{i}\cdot (\ub_{i} + G)) + 
		\Sigma_{i=l+1}^{2l-s}(\x'_{i}\cdot G) +
		\Sigma_{j=1}^{m}(\y_{j}\cdot \vb'_{j}) 
		\neq T'	
	\end{aligned}
\end{equation}







Similarly, by partitioning the summation and considering the coordinates of $\x'$, we can deduce that the aforementioned equation can be rewritten as:

\begin{equation}
	\begin{aligned}	
		\forall \y\in \{0,1\}^m,  \quad
		\Sigma_{i=1}^{s}(1\cdot \ub_{i}) + 
		\Sigma_{i=1}^{s}(1\cdot G) + \\
		\Sigma_{i=l+1}^{2l-s}(1\cdot G) +
		\Sigma_{j=1}^{m}(\y_{j}\cdot \vb'_{j}) 
		\neq T'	\iff \\
  		\forall \y\in \{0,1\}^m  \quad
		\Sigma_{i=1}^{s}(1\cdot \ub_{i}) + 
		(s+(l-s))\cdot G + \\
		\Sigma_{j=1}^{m}(\y_{j}\cdot \vb'_{j}) 
		\neq T'	 = T + (l\cdot G) \iff \\
  		\forall \y\in \{0,1\}^m  \quad
		\Sigma_{i=1}^{s}(1\cdot \ub_{i}) + 
		\Sigma_{j=1}^{m}(\y_{j}\cdot \vb'_{j}) 
		\neq T 
	\end{aligned}
\end{equation}




Equivalently:

\begin{equation}
	\label{eq:contradictionResult}
	\begin{aligned}	
		\forall \y\in \{0,1\}^m,  \quad
		\Sigma_{i=1}^{s}(1\cdot \ub_{i}) +
		\Sigma_{i=s+1}^{2l}(0\cdot \ub_{i}) +  
		\Sigma_{j=1}^{m}(\y_{j}\cdot \vb'_{j}) 
		\neq T 
	\end{aligned}
\end{equation}

Let us define an input $\x''\in\{0,1\}^l$, such that the first $1\leq s \leq l$ coordinates are ``1'', and the remaining ones are``0''. In other words, $\x'' \coloneqq \x'_{[1\ldots l]}$. Therefore, given Eq.~\ref{eq:contradictionResult} and the specified $\x''\in\{0,1\}^{l}$, it follows that:



\begin{equation}
	\begin{aligned}
		\forall \y\in \{0,1\}^m \quad	
		\Sigma_{i=1}^{l}(\x''_{i}\cdot \ub_{i}) + 
		\Sigma_{j=1}^{m}(\y_{j}\cdot \vb_{j}) 
		\neq T
	\end{aligned}
\end{equation}

However, this contradicts Eq.~\ref{eq:lemma32SecondDirection}, and therefore we determine that our initial assumption is erroneous (i.e., no such $\x'\in\{0,1\}^{2l}$ exists), and:

\begin{equation}
	\begin{aligned}	
		\forall \x'\in\{0,1\}^{2l} \quad \text{s.t} \quad 
		\left\lVert \x' \right\rVert_{1} = k', \\
		 \exists \y\in \{0,1\}^m \quad
		\Sigma_{i=1}^{2l}(\x'_{i}\cdot \ub'_{i}) + 
		\Sigma_{j=1}^{m}(\y_{j}\cdot \vb'_{j}) 
		= T'
	\end{aligned}
\end{equation}

Thus, $\langle \ub', 
\vb', k', T'\rangle  \notin \kGSSP$. To conclude, we have proved that:
$\langle \ub, 
\vb, T\rangle  \in \GSSP \iff 
\langle \ub', 
\vb', k', T'\rangle  \in \kGSSP$, completing our reduction.

$\qedsymbol{}$

Next, we introduce a variant of the $\kGSSP$ problem, which we refer to as the $\ckGSSP$ problem.






\vspace{0.5em} 

\noindent\fbox{%
	\parbox{\columnwidth}{%
		\mysubsection{$\ckGSSP$ (Constrained k Generalized Subset Sum)}:
		
		\textbf{Input}:
		A set of positive integers
		$(\z_1,\z_2,\ldots,\z_n)$, a subset $S_0$,
		an integer $k$,
		and a positive (target) integer $T$.
		
		\textbf{Output}: \yes{}, if there \emph{exists} a subset of features 
		(indices) $S\subseteq 
		S_0$, such that, $\vert S \vert = k$, and for \emph{all} subsets $S' 
		\subseteq\Bar{S}$  it holds that
		$\Sigma_{i\in S}(\z_{i}) + \Sigma_{j\in S'}(\z_{j}) \neq T$; and \no{} 
		otherwise.
	}%
}

\vspace{0.5em} 

Now, we will establish the hardness of this refined query by demonstrating the following claim:

\begin{claim}
	\label{lemma:constrained-K-Gssp}
	The query $\ckGSSP$ is \StoPCompleteComplexity-hard.
\end{claim}

\emph{Proof.} We will demonstrate \StoPCompleteComplexity-hardness through a reduction from the \emph{k-GSSP} problem. Starting with $\langle \ub=(\ub_1,\ub_2,\ldots,\ub_l), \vb=(\vb_1,\vb_2,\ldots,\vb_m), k, T\rangle$, we produce $\langle \z=(\z_1,\z_2,\ldots,\z_n), S_{0}=(1,\ldots,l), k', T'\rangle$. The vector $\z$ is formed by concatenating $\ub$ and $\vb$ in the following manner:

\begin{itemize}
	\item 
	$\textbf{z} \coloneqq  (\ub'^\frown \vb') \in \mathbb{N}^{n \coloneqq  (l+m)}$, 
	i.e., for any $1 \leq i \leq l$: $\z_{i}=\ub_{i}$, and for any $(l+1) \leq j 
	\leq (l+m)$: $\z_{j}=\vb_{j-l}$.

	\item
	$\ub'\coloneqq
	[(2n+1)\cdot \ub] + \mathbbm{1}$, i.e., $\ub'_{j} = [(2n+1) \cdot 
	\ub_{j}] + 1$

	\item
	$\vb'\coloneqq (2n+1)\cdot \vb$, i.e., $\vb'_{i} = (2n+1) \cdot 
	 \vb_{i}$
\end{itemize}

We also set $T' \coloneqq T(2n+1)+k$ and $k' \coloneqq k$. We then will prove that: $\langle \ub, \vb, k, T\rangle \in \kGSSP \iff \langle \z, S_{0}, k', T'\rangle \in \ckGSSP$.


First, suppose $\langle \ub,\vb k, T\rangle \in \kGSSP$. Then, there \emph{exists} a binary vector $\x\in{0,1}^l$ such that $\left\lVert \x \right\rVert_{1}=k$, and for \emph{any} binary vector $\y\in{0,1}^m$, the following condition is met: $\Sigma_{i=1}^{l}(\x_{i}\cdot \ub_{i}) + \Sigma_{j=1}^{m}(\y_{j}\cdot \vb_{j}) \neq T$. Assuming, without loss of generality, that the \emph{first} $k$ entries of $\ub$ align with this particular binary vector $\x\in\{0,1\}^l$ (note that $\left\lVert \x \right\rVert_{1}= k$), we define the subset $S$ to include all corresponding indices, thus $S \coloneqq 
\{i \vert \x_{i}=1\} = \{1,\ldots, k\}\subseteq S_{0}$. We will now establish that for \emph{all} subsets $S' \subseteq\Bar{S} \coloneqq 
\{(k+1),\ldots,l,(l+1),\ldots,n\coloneqq (l+m)\}$, the following holds true for our chosen set $S$:




\begin{equation}
	\begin{aligned}
		\Sigma_{i\in S}(\z_{i}) + \Sigma_{j\in S'\subseteq\Bar{S}}(\z_{j}) \neq T'
	\end{aligned}
\end{equation}





We will demonstrate this in two parts: initially by considering subsets of $\Bar{S}$ that include only features from $\{(l+1),\ldots,(l+m)\}$, and subsequently by considering subsets of $\Bar{S}$ that intersect with $\{(k+1),\ldots,l\}$. In the first scenario, consider all $S'\subseteq\Bar{S}$ where $S'\subseteq \{(l+1),\ldots, (l+m)=n\}$, meaning the features correspond exclusively to those in the original $\vb$ vector. For this particular $\x\in\{0,1\}^l$, it is established that for \emph{any} $\y\in\{0,1\}^m$:




\begin{equation}
	\begin{aligned}
		\Sigma_{i=1}^{l}(\x_{i}\cdot \ub_{i}) + \Sigma_{j=1}^{m}(\y_{j}\cdot 
		\vb_{j}) 
		\neq T \iff \\
  		(2n+1)\Sigma_{i=1}^{l}(\x_{i}\cdot \ub_{i}) + 
		(2n+1)\Sigma_{j=1}^{m}(\y_{j}\cdot \vb_{j})
		\neq T(2n+1) \iff \\
  		(2n+1)\Sigma_{i=1}^{l}(\x_{i}\cdot \ub_{i}) + 
		(2n+1)\Sigma_{j=1}^{m}(\y_{j}\cdot \vb_{j}) + k
		\neq T(2n+1) + k = T'
	\end{aligned}
\end{equation}





Given that $\left\lVert \x \right\rVert_{1}=k$, we proceed under the assumption (w.l.o.g.) that the \emph{first} $k$ indices of $\x$ are set to ``1'', while the remaining ($l-k$) coordinates are 0''. Therefore, for \emph{any} $\y\in\{0,1\}^m$:


\begin{equation}
	\begin{aligned}
		(2n+1)\Sigma_{i=1}^{k}(\x_{i}\cdot \ub_{i}) + 
		(2n+1)\Sigma_{i=k+1}^{l}(\x_{i}\cdot \ub_{i}) + \\
		(2n+1)\Sigma_{j=1}^{m}(\y_{j}\cdot \vb_{j}) + k
		\neq  T' \iff \\
  		\Sigma_{i=1}^{k}((2n+1)\cdot \x_{i}\cdot \ub_{i}) + 
		\Sigma_{i=k+1}^{l}((2n+1)\cdot \x_{i}\cdot \ub_{i}) + \\
		\Sigma_{j=1}^{m}((2n+1)\cdot \y_{j}\cdot \vb_{j}) + k
		\neq  T'
	\end{aligned}
\end{equation}



And, assuming (w.l.o.g.) that the \emph{first} $k$ coordinates of $\x$ are ``1'', with the remaining ones set to ``0'', it follows that for \emph{any} $\y\in\{0,1\}^m$:


\begin{equation}
	\begin{aligned}
		\Sigma_{i=1}^{k}((2n+1)\cdot 1\cdot \ub_{i}) + 
		\Sigma_{i=k+1}^{l}((2n+1)\cdot 0 \cdot \ub_{i}) + \\
		\Sigma_{j=1}^{m}((2n+1)\cdot \y_{j}\cdot \vb_{j}) + k
		\neq  T' \iff \\
  		\Sigma_{i=1}^{k}((2n+1)\cdot 1\cdot \ub_{i}) + 
		\Sigma_{j=1}^{m}((2n+1)\cdot \y_{j}\cdot \vb_{j}) + k
		\neq  T' \iff \\
  		\Sigma_{i=1}^{k}([(2n+1) \cdot 
		\ub_{j}] + 1) + 
		\Sigma_{j=1}^{m}(\y_{j}\cdot (2n+1)\cdot  \vb_{j})
		\neq  T'
	\end{aligned}
\end{equation}





Given our definitions of $\ub'$ and $\vb'$, it follows that for \emph{any} $\y\in\{0,1\}^m$:


\begin{equation}
	\begin{aligned}
		\Sigma_{i=1}^{k}(\ub'_{i}) + 
		\Sigma_{j=1}^{m}(\y_{j} \cdot \vb'_{j})
		\neq  T'
	\end{aligned}
\end{equation}

%

Since $S$ encompasses the first $k$ coordinates of \textbf{u'} (which align with the $k$ ``1'' coordinates of the specified $\x$, by design), it results that for \emph{any} $\y\in\{0,1\}^m$:


\begin{equation}
	\begin{aligned}
		\Sigma_{i\in S}(\z_i) +
		\Sigma_{j=1}^{m}(\y_{j} \cdot \z_{l+j})
		\neq T'
	\end{aligned}
\end{equation}

%
%
%
%

Therefore, for all $S'\subseteq \Bar{S}$ where $S'\subseteq \{(l+1),\ldots,(l+m)\}$, it is established that:


\begin{equation}
	\begin{aligned}
		\Sigma_{i\in S}(\z_i) +
		\Sigma_{i \in S'\subseteq\Bar{S}}(\z_i) \neq
		T'
	\end{aligned}
\end{equation}

In the second scenario, we consider any subset $S'\subseteq\Bar{S}$ that intersects with $\{(k+1),\ldots,l\}$, meaning there is at least one feature $i\in S'$ such that $(k+1)\leq i \leq l$. We note that, according to our construction, $S$ contains \emph{exactly} $k$ indices, each corresponding to a value of $[(2n+1)\cdot \ub_i] + 1$. In this second case, $S'\subseteq\Bar{S}$ includes \emph{at least} one index that corresponds to a value of $[(2n+1)\cdot \ub_i] + 1$ (corresponding to $\ub'$) and potentially other values of $(2n+1)\cdot \vb_i$ (corresponding to $\vb'$). Consequently, when adding the values associated with the indices from $S$ and $S'$, we obtain \emph{at least} $(k+1)$ values (and at most, $n$), with \emph{at least} $k+1$ values $t$ being such that $t \mod (2n+1) = 1$. Therefore, the total sum will yield a value with a modulus of \emph{at least} $k+1$ over $(2n+1)$, implying that:




\begin{equation}
	\begin{aligned}
		[\Sigma_{i\in S}(\z_i)] mod 
		(2n+1) = k \implies \\
  		1 \leq [ \Sigma_{i\in S'\subseteq\Bar{S}}(\z_i)] mod 
		(2n+1) \leq l-k \implies \\
  		(\Sigma_{i\in S}(\z_i) + \Sigma_{i\in S'\subseteq\Bar{S}}(\z_i))mod 
		(2n+1) \neq k
	\end{aligned}
\end{equation}


Lastly, given that $T' \coloneqq (2n+1)T+k$, it follows that $(T') \mod (2n+1) = k$, therefore:

\begin{equation}
	\begin{aligned}
		\Sigma_{i\in S}(\z_i) + \Sigma_{i\in S'\subseteq\Bar{S}}(\z_i) \neq T' =
		(2n+1)T+k
	\end{aligned}
\end{equation}

Thus, we have demonstrated in both scenarios that for our defined $S \subseteq S_0$, it is true that $\lvert S \rvert = k$ and for \emph{all} subsets $S' \subseteq\Bar{S} \coloneqq \{(k+1),\ldots,l,(l+1),\ldots,n\coloneqq (l+m)\}$, it holds that:


\begin{equation}
	\begin{aligned}
		\Sigma_{i\in S}(\z_{i}) + \Sigma_{j\in S'\subseteq\Bar{S}}(\z_{j}) \neq T'
	\end{aligned}
\end{equation}

Hence, $\langle 
\textbf{z}, S_{0}, k', T'\rangle
\in\ckGSSP$

For the other direction, if $\langle \ub, \vb, k, T\rangle \notin \kGSSP$, then for \emph{every} binary vector $\x\in\{0,1\}^l$ with $\left\lVert \x \right\rVert_{1}=k$, there \emph{exists} a binary vector $\y'\in\{0,1\}^m$ such that: $\Sigma_{i=1}^{l}(\x_{i}\cdot \ub_{i}) + \Sigma_{j=1}^{m}(\y'{j}\cdot \vb{j}) = T$. For every $k$-sized subset $S \subseteq S_{0}$, we define $\x'\coloneqq \mathbbm{1}_{{i\in S}}$. Additionally, for each fixed $S$, we define the set $S' \subseteq \Bar{S}$ to be:




\begin{equation}
	\begin{aligned}
		S' \coloneqq 
		\left \{ 
		(\vert \ub \vert +i )
		\vert 
		\y'_{i}=1
		\right \}
	\end{aligned}
\end{equation}

for the corresponding $\y'$ that aligns with the indicator $\x'$ associated with $S'$. Next, we observe that:


\begin{equation}
	\begin{aligned}
		\Sigma_{i\in S}(\z_{i}) + \Sigma_{j\in S'\subseteq\Bar{S}}(\z_{j}) 
		\underset{(*)}{=} \\
		\Sigma_{i\in S}(\ub'_{i}) + \Sigma_{j\in S'\subseteq\Bar{S}}(\vb'_{j}) 
		= \\
		\Sigma_{i\in S}((2n+1)\ub_{i}+1) + \Sigma_{j\in 
		S'\subseteq\Bar{S}}((2n+1)\vb_{i})
		= \\
		(2n+1) 
		[\Sigma_{i\in S}(\ub_{i}) + \Sigma_{j\in S'\subseteq\Bar{S}}(\vb_{i})]
		+\vert S \vert \cdot 1
		\underset{(**)}{=} \\
		(2n+1) 
		[\Sigma_{i=1}^{l}(\x'_{i}\cdot \ub_{i}) + 
		\Sigma_{j=1}^{m}(\y'_{j}\cdot \vb_{j})]
		+k
		=\\
		(2n+1) 
		[T]
		+k
		=\\
		(2n+1) 
		[\frac{T'-k}{2n+1}]
		+k
		=\\
		[T'-k]
		+k
	\end{aligned}
\end{equation}

Where (*) arises because we have selected $S'\subseteq \Bar{S}$ to exclusively contain indices of values pertaining to \textbf{v'} (i.e., $S'\cap {1,\ldots,l}\neq\emptyset$), and (**) is due to considering any subset $S\subseteq S_0$ with a size of $\lvert S \rvert = k$. Therefore, we conclude that for \emph{all} $k$-sized subsets $S \subseteq S_{0}$, it is established that there \emph{exists} a subset $S' \subseteq\Bar{S}$ such that:



\begin{equation}
	\begin{aligned}
		\Sigma_{i\in S}(\z_{i}) + \Sigma_{j\in S'\subseteq\Bar{S}}(\z_{j}) = T'
	\end{aligned}
\end{equation}

Thus, $\langle 
\z, S_{0}, k', T'\rangle
\notin
\ckGSSP$, as required.

$\qedsymbol{}$


We will now present the final component of this proof by establishing the following claim:

\begin{claim}
	The MSR query for an ensemble of $k$-Perceptrons is \StoPCompleteComplexity-hard for $k=5$.
\end{claim}

\emph{Proof.} 
We will outline a polynomial-time reduction from $\ckGSSP$ to k-perceptron-MSR, specifically for $k=5$. Given $\langle \z, S_{0}, k, T \rangle$, our polynomial-time reduction produces $\langle (f_{1},\ldots, f_{5}), \x', k \rangle$, in the following manner: Initially, the reduction verifies (in polynomial time) whether: $\Sigma_{i=1}(\z_{i})=T$. If this condition is met, then since $\z$ consists solely of strictly positive integers, any \emph{strict} $k$-sized subset of these will not sum to $T$. Consequently, $\langle \z, S_{0}, k, T \rangle$ qualifies as belonging to k-GSSP*. As a result, the reduction returns, within polynomial time, $\langle (f_{1}\coloneqq \text{True},\ldots, f_{5}\coloneqq \text{True}), \x', k \rangle$ for k-Perceptron-MSR, as any $k$-sized set of features sufficiently justifies the condition. Otherwise, if $\Sigma_{i=1}(\z_{i}) \neq T$, the following steps are taken: we initialize $\x'\coloneqq 1^{n}$ and $k'\coloneqq k$. For each $1\leq i \leq 5$, the perceptron $f_{i}$ is defined as follows:




\begin{itemize}
	\item 
	$f_{1}\coloneqq (\wb_{1},b_{1})$, for $\wb_{1}\coloneqq 
	(-\z_{1},\ldots,-\z_{n})$, and $b_{1}\coloneqq T-\frac{1}{2}$
	
	\item
	$f_{2}\coloneqq (\wb_{2},b_{2})$, for $\wb_{2}\coloneqq (z_{1},\ldots,z_{n})$, 
	and $b_{2}\coloneqq -T-\frac{1}{2}$
	
	\item
	$f_{3}\coloneqq (\wb_{3},b_{3})$, for $\wb_{3}\coloneqq 
	(\mathbf{1}_{S_{0}};\mathbf{0}_{\Bar{S_{0}}}$) and 
	$b_{3}\coloneqq -k$ it holds that: 
	$f_{3}(\x)=1 
	\iff \Sigma_{i=1}^{n} (\x_{i})-k \geq 0 
	\iff\Sigma_{i=1}^{n} (\x_{i}) \geq k 
	\iff 
	\Sigma_{i=1}^{\vert S_{0} \vert} (\mathbf{1}_{S_{0}} \land \x_{i})
	+ \Sigma_{i=\vert S_{0} \vert + 1}^{n} (\mathbf{1}_{S_{0}} \land \x_{i})
	\geq k
	\iff
	\Sigma_{i=1}^{\vert S_{0} \vert} (\mathbf{1}_{S_{0}} \land \x_{i})
	\geq k
	\iff
	\vert \left \{ \x_{i} \vert \x_{i} = 1 \land \x_{i}\in S_{0} \right \} \vert 
	\geq k$, i.e., $f_{3}$ classifies as ``1'' if and only if the input has 
	\emph{at least} $k$ ``1'' values in $S_{0}$.
	
	\item
	$f_{4}\coloneqq (\wb_{4},b_{4})$, for $\wb_{4}\coloneqq (0,\ldots,0)$, and 
	$b_{4}\coloneqq 1$, i.e., $f_{4}$=True
	
	\item 
	$f_{5}\coloneqq \mathbf{1}_{\mathbf{1}^n}$, i.e., $f_{5}(\x)=1 \iff 
	\x\coloneqq \mathbf{1}^{n}$  (i.e., $f_5$ acts as a Perceptron serving as an indicator function for the constructed input $\x \coloneqq \mathbf{1}^{n}$. This setup can be implemented in polynomial time, as demonstrated in~\cite{AmBaKa24}.)
\end{itemize}

First, we observe that the ensemble $(f_{1},\ldots,f_{5})$ classifies $\textbf{x'}\coloneqq 1^{n}$ as ``1''. This classification results from a majority of at least three (out of five) ensemble members designating the input $\mathbf{1}^{n}$ as ``1'':

\begin{equation}
	\begin{aligned}
		f_{3}(\textbf{x'})=
		step(\Sigma_{i=1}^{\vert S_{0} \vert} (\x'_i - k) =
		step(\Sigma_{i=1}^{\vert S_{0} \vert} (1) - k) =
		step(\vert S_{0} \vert - k) =1
	\end{aligned}
\end{equation}

Thus, $f_{3}(\textbf{x'})=f_{3}(\mathbf{1}^{n})=1$. Additionally, $f_{4}(\mathbf{1}^{n})=1$, as it classifies every input as ``1'', and $f_{5}(\mathbf{1}^{n})=1$ as it functions as an indicator for this very input. We will prove that: $\langle \textbf{z}, S_{0}, k, T\rangle \in \ckGSSP \iff \langle (f_{1},\ldots,f_{5}), \textbf{x'},k' \rangle \in$ MSR for $k$-ensembles of Perceptrons when $k=5$.



Initially, assuming that $\langle \textbf{z}, S_{0}, k, T\rangle \in \ckGSSP$, there \emph{exists} a subset $S\subseteq S_{0}$ with $\vert S \vert =k$ and for every $S'\subset \Bar{S}$ it holds that $\Sigma_{i\in S}(\z_{i}) + \Sigma_{j\in S'}(\z_{j}) \neq T$. We then contend that $S$ serves as a $k$-sized explanation for input $\textbf{x'}\coloneqq \mathbf{1}^{n}$ within the 5-Perceptron ensemble $(f_{1},\ldots, f_{5})$. When considering $S$, we conclude that for every $S'\subseteq \Bar{S}$: $\Sigma_{i\in S}(\z_{i}) + \Sigma_{j\in S'}(\z_{j}) \neq T$, or equivalently, for each $S'$:




\begin{equation}
	\begin{aligned}
		[ \ \Sigma_{i\in S}(\z_{i}) + \Sigma_{j\in S'}(\z_{j}) \geq T + 1 \ ] \vee [ \ 		\Sigma_{i\in S}(\z_{i}) + \Sigma_{j\in S'}(\z_{j}) \leq T - 1 \ ]
	\end{aligned}
\end{equation}



We will demonstrate that in both cases, \emph{exactly} one of $f_1$ and $f_2$ classifies an input as ``1'', while the other classifies it as ``0''. In the scenario where $\Sigma_{i\in S}(\z_{i}) + \Sigma_{j\in S'}(\z_{j}) \geq T + 1$, for any input $\x\in\{0,1\}^n$ with $\x_{S}=\mathbf{1}{S}$, we construct $(\x_{S};\y_{\Bar{S}})$, applicable for $\x\in\{0,1\}^{n}$ and for \emph{every} $\y\in\{0,1\}^{n} $, ensuring that both $f_2((\x_{S};\y_{\Bar{S}}))=1$ and $f_1((\x_{S};\y_{\Bar{S}}))=0$. Moreover, $f_2((\x_{S};\y_{\Bar{S}}))=1$`, based on the following:




\begin{equation}
	\begin{aligned}
		f_{2}((\x_{S};\y_{\Bar{S}}))
		=\\
		step([(\x_{S};\y_{\Bar{S}})\cdot \wb_{2}] + b_{2})
		=\\
		step([(\x_{S};\y_{\Bar{S}})\cdot (z_{1},\ldots, z_{n})] + 
		(-T-\frac{1}{2}))
		=\\
		step(\Sigma_{i\in S}(\x_{i} \cdot \z_{i}) + \Sigma_{j\in \Bar{S}}(\y_{i} \cdot 
		\z_{j})
		-T-\frac{1}{2})
		\geq\\
		step((T+1)
		-T-\frac{1}{2})
		= step(\frac{1}{2}) =1
	\end{aligned}
\end{equation}

In this case, it also true that 
$f_1((\x_{S};\y_{\Bar{S}}))=0$, as the following holds:

\begin{equation}
	\begin{aligned}
		f_{1}((\x_{S};\y_{\Bar{S}}))
		=\\
		step([(\x_{S};\y_{\Bar{S}})\cdot \wb_{1}] + b_{1})
		=\\
		step([(\x_{S};\y_{\Bar{S}})\cdot (-\z_{1},\ldots, -\z_{n})] + 
		(T-\frac{1}{2}))
		= \\
		step(- \Sigma_{i\in S}(z_{i}) - \Sigma_{j\in \Bar{S}}(z_{j}) 
		 + 
		(T-\frac{1}{2}))
		\leq \\
		step(- (T + 1)  + 
		(T-\frac{1}{2})) = step(-\frac{3}{2})=0
	\end{aligned}
\end{equation}

With the last transition justified by the assumption that: $\Sigma_{i\in S}(\z_{i}) + \Sigma_{j\in S'}(\z_{j}) \geq T + 1$, and we also observe that we have selected $S'\coloneqq \Bar{S}$ (which is valid since trivially $\Bar{S}\subseteq \Bar{S}$). In the situation where $\Sigma_{i\in S}(\z_{i}) + \Sigma_{j\in S'}(\z_{j}) \leq T - 1$, for any input $\x\in\{0,1\}^n$ with $\x_{S}=\mathbf{1}$, we construct $(\x_{S};\y_{\Bar{S}})$, suitable for $\x\in\{0,1\}^{n}$ and for \emph{every} $\y\in\{0,1\}^{n} $, ensuring that both $f_1((\x_{S};\y_{\Bar{S}}))=1$ and $f_2((\x_{S};\y_{\Bar{S}}))=0$. Additionaly, $f_1((\x_{S};\y_{\Bar{S}}))=1$, based on the following:



 
\begin{equation}
	\begin{aligned}
		f_{1}((\x_{S};\y_{\Bar{S}}))
		=\\
		step([(\x_{S};\y_{\Bar{S}})\cdot \wb_{1}] + b_{1})
		=\\
		step([(\x_{S};\y_{\Bar{S}})\cdot (-\z_{1},\ldots, - \z_{n})] + 
		(T-\frac{1}{2}))
		=\\
		step(- [\Sigma_{i\in S}(\x_{i} \cdot \z_{i}) + \Sigma_{j\in \Bar{S}}(\y_{i} 
		\cdot 
		\z_{j})] +
		T-\frac{1}{2})
		\geq\\
		step((-1)(T-1) +
		T-\frac{1}{2})
		= step(1 - T + T - 
		\frac{1}{2}) = 
		step(\frac{1}{2}) =1
	\end{aligned}
\end{equation}

In this scenario, it also holds that 
$f_2((\x_{S};\y_{\Bar{S}}))=0$, as the following holds:

\begin{equation}
	\begin{aligned}
		f_{2}((\x_{S};\y_{\Bar{S}}))
		=\\
		step([(\x_{S};\y_{\Bar{S}})\cdot \wb_{2}] + b_{2})
		=\\
		step([(\x_{S};\y_{\Bar{S}})\cdot (\z_{1},\ldots, \z_{n})] + 
		(-T-\frac{1}{2}))
		= \\
		step(\Sigma_{i\in S}(\z_{i}) + \Sigma_{j\in \Bar{S}}(\z_{j}) 
		+ 
		(-T-\frac{1}{2}))
		\leq \\
		step(T - 1  + 
		(-T-\frac{1}{2})) = step(-\frac{3}{2})=0
	\end{aligned}
\end{equation}

The last transition remains valid under the assumption: $\Sigma_{i\in S}(\z_{i}) + \Sigma_{j\in S'}(\z_{j}) \leq T - 1$. We further observe that $S'\coloneqq \Bar{S}$, which is valid since $\Bar{S}\subseteq \Bar{S}$ by definition. Therefore, when $\Sigma_{i\in S}(\z_{i}) + \Sigma_{j\in S'}(\z_{j}) \neq T$, it follows that \emph{exactly} one of the Perceptrons in the pair $(f_1, f_2)$ classifies the input $(\x_{S};\y_{\Bar{S}})$ as ``1'', while the other classifies it as ``0''.




Furthermore, $f_{3}((\x_{S};\y_{\Bar{S}}))=1$ because $(\x_{S};\y_{\Bar{S}})$ contains $k$ ``1'' values, which is the threshold required by $f_3$ to activate. It is also observed that $f_{4}((\x_{S};\y_{\Bar{S}}))\geq 0$, implying that it evaluates to True (classifying all inputs as 1''). Therefore, when $k$ values of ``1'' are set in $S$, one of two outcomes occurs: either the majority $(f_{2}, f_{3}, f_{4})$ or the majority $(f_{1}, f_{3}, f_{4})$ classifies every input $(\x_{S},\y_{\Bar{S}})$ as ``1'' for \emph{every} possible $\y \in \{0,1\}^{n}$. Consequently, $S$ serves as a $k$-sized sufficient reason for input $\mathbf{1}^{n}$ with respect to $(f_{2}, \ldots, f_{5})$, thus $\langle (f_{1}, \ldots, f_{5}), \textbf{x'}, k' \rangle \in$ MSR for $k$-ensemble Perceptrons when $k=5$.




For the second direction, assume that $\langle \textbf{z}, S_{0}, k, T\rangle \notin \ckGSSP$. We aim to demonstrate that this leads to $\langle (f_{1},\ldots,f_{5}), \textbf{x'}, k' \rangle \notin$ MSR for $k$-ensemble Perceptrons when $k=5$. For the sake of contradiction, suppose the opposite is true, i.e., $\langle (f_{1},\ldots,f_{5}), \textbf{x'}, k' \rangle \in$ MSR for $k$-ensemble Perceptrons when $k=5$. In other words, we posit that there are $k$ features in $\textbf{x}=\mathbf{1}^{n}$ which, when fixed, result in the ensemble consistently classifying as ``1''. We will prove this to be impracticable by examining the only two scenarios: \begin{inparaenum}[(i)] \item that these features are all contained within $S_0$; \item that at least one of the fixed $k$ features resides in $\Bar{S_0}$. \end{inparaenum}



In the first scenario, assume that all $k$ features are within $S_0$. Yet, since $\langle \textbf{z}, S_{0}, k, T\rangle \notin \ckGSSP$, it follows that for \emph{any} input $(\x_{S};\y_{\Bar{S}})$ (for \emph{any} $k$-sized subset $S\subseteq S_0$ and \emph{any} $\x\in\{0,1\}^n$), there \emph{exists} a corresponding $\y\in\{0,1\}^n$ such that:


\begin{equation}
	\begin{aligned}
		\Sigma_{i\in S}(\z_{i}) + \Sigma_{j\in S'}(\z_{j}) = T
	\end{aligned}
\end{equation}

This suggests that \emph{both} $f_{1}$ and $f_{2}$ classify the input $(\x_{S};\y_{\Bar{S}})$ as ``0'', which we will demonstrate below. Additionally, $f_{1}((\x_{S};\y_{\Bar{S}})) = 0$, as evidenced by the following:

\begin{equation}
	\begin{aligned}
		f_{1}((\x_{S};\y_{\Bar{S}}))
		=\\
		step([(\x_{S};\y_{\Bar{S}})\cdot \wb_{1}] + b_{1})
		=\\
		step([(\x_{S};\y_{\Bar{S}})\cdot (-\z_{1},\ldots, - \z_{n})] + 
		(T-\frac{1}{2}))
		=\\
		step(- [\Sigma_{i\in S}(\x_{i} \cdot \z_{i}) + \Sigma_{j\in \Bar{S}}(\y_{i} 
		\cdot 
		\z_{j})] +
		T-\frac{1}{2})
		=\\
		step(- T + T - \frac{1}{2})
		= step(-\frac{1}{2}) = 0
	\end{aligned}
\end{equation}

Additionally, $f_{2}((\x_{S};\y_{\Bar{S}})) = 0$, as the following 
holds:
\begin{equation}
	\begin{aligned}
		f_{2}((\x_{S};\y_{\Bar{S}}))
		=\\
		step([(\x_{S};\y_{\Bar{S}})\cdot \wb_{2}] + b_{2})
		=\\
		step([(\x_{S};\y_{\Bar{S}})\cdot (\z_{1},\ldots, \z_{n})] + 
		(-T-\frac{1}{2}))
		=\\
		step(\Sigma_{i\in S}(\x_{i} \cdot \z_{i}) + \Sigma_{j\in \Bar{S}}(\y_{i} \cdot 
		\z_{j})
		-T-\frac{1}{2})
		=\\
		step(T - T - \frac{1}{2})
		= step(- \frac{1}{2}) = 0
	\end{aligned}
\end{equation}

Therefore, under the assumption that \emph{all} $k$ fixed features are within $S_0$, there \emph{exists} an input $\mathbf{y}\in\{0,1\}^n$ such that for the input $(\mathbf{x}_{S};\mathbf{y}_{\Bar{S}})$, both $f_{1}$ and $f_{2}$ classify $f_1((\mathbf{x}_{S};\mathbf{y}_{\Bar{S}}))=0$ and $f_2((\mathbf{x}_{S};\mathbf{y}_{\Bar{S}}))=0$. Next, we will demonstrate that the input $(\mathbf{1}_{S};\mathbf{0}_{\Bar{S}}) \neq \mathbf{1}^n$, i.e., it contains \emph{at least} one feature with value ``0''. This is because, for the defined $f_{1}$ and $f_{2}$, it is not possible that both $f_1(\mathbf{1}^n)=0$ and $f_2(\mathbf{1}^n)=0$, as this would require:



\begin{equation}
	\begin{aligned}
		f_{1}(\mathbf{1}^n)=0 \land  f_{2}(\mathbf{1}^n)=0 
		\iff \\
		[step(\Sigma_{i=1}^n(1 \cdot -\z_{i}) + (T-\frac{1}{2})) = 0] 
		\land 
		[step(\Sigma_{i=1}^n(1 \cdot \z_{i}) + (-T-\frac{1}{2})) = 0] \iff 
  \\
		[\Sigma_{i=1}^n(1 \cdot -\z_{i}) + (T-\frac{1}{2}) < 0] 
		\land 
		[\Sigma_{i=1}^n(1 \cdot \z_{i}) + (-T-\frac{1}{2}) < 0] 
		\iff \\
		[\Sigma_{i=1}^n(1 \cdot \z_{i}) > T-\frac{1}{2}] 
		\land 
		[\Sigma_{i=1}^n(1 \cdot \z_{i}) < T+\frac{1}{2}] 
		\iff \\
		T - \frac{1}{2} < \Sigma_{i=1}^n(1 \cdot \z_{i}) < T+\frac{1}{2} 	
	\end{aligned}
\end{equation}

Given that the $\z_i$ values are positive integers, this is only true if $\Sigma_{i=1}^n(1 \cdot \z_{i})=T$, which contradicts our assumption. Therefore, for $\mathbf{1}^n$, it is not the case that both $f_1(\mathbf{1}^n)=0$ and $f_2(\mathbf{1}^n)=0$. Consequently, the \emph{existing} input $\y\in\{0,1\}^n$ such that both $f_1((\x_{S};\y_{\Bar{S}}))=0$ and $f_2((\x_{S};\y_{\Bar{S}}))=0$ is not $\mathbf{1}^n$, i.e., $(\mathbf{1}_{S};\mathbf{0}_{\Bar{S}}) \neq (\mathbf{1}_{S};\mathbf{1}_{\Bar{S}})=\mathbf{1}^n$. In this scenario, it also holds that for this input $f_{5}((\mathbf{1}_{S};\mathbf{0}_{\Bar{S}})) = 0 $, since $f_3$ is an indicator for the input $\mathbf{1}^n$. Therefore, for \emph{any} $k$ fixed inputs originating solely from $S_0$, there always \emph{exists} an $(\mathbf{1}_{S};\mathbf{0}_{\Bar{S}})$, such that the perceptron majority $(f_{1},f_{2},f_{5})$, and thus the ensemble overall, classifies this input $(\mathbf{1}_{S};\mathbf{0}_{\Bar{S}})$ as ``0'', contrary to the initial classification. Consequently, given our premise that $\langle \textbf{z}, S_{0}, k, T\rangle \notin \ckGSSP$, no $k$ features from $S_0$ qualify as a sufficient reason with respect to $\mathbf{1}^n$.

In the second scenario, we examine the case where \emph{not all} $k$ fixed features are in $S_0$, which, under our assumption, means the MSR (Minimum Sufficient Reason) consists of \emph{at most} $(k-1)$ fixed values of ``1'' in $S_0$. We will show why this is also infeasible. We consider any completion with exactly $\vert S \vert -k$ ``0'' values in the features of $S$ to illustrate this point. Using the same logic as previously, based on the configuration of $f_{1}$ and $f_{2}$, it is established that for any input -—- \emph{at least} one of $f_{1}$ or $f_{2}$ classifies it as ``0''. As demonstrated, \emph{exactly} one Perceptron of the pair classifies it as ``0'' when the summation of $\z_i$-s does not equal $T$, and both of them classify it as ``0'' when the summation equals $T$.



Furthermore, any input with at most $k-1$ ``0'' values in features corresponding to $S$ is classified by $f_{5}$ as ``0'', since $f_3$ activates only when it detects \emph{at least} $k$ of $S$'s features as ``1''. Additionally, any input that includes more than a single zero value is clearly not equal to $\mathbf{1}^n$ and, therefore, will also result in $f_{5}$ classifying it as ``0'', given that $f_{5}$ serves as an indicator for $\mathbf{1}^n$ (classifying any other input as 0''). Thus, in this scenario as well, there is a majority of $(f_{1}, f_{3}, f_{5})$ or $(f_{2}, f_{3}, f_{5})$, and consequently the entire ensemble, classifying any such input as ``0''.




This indicates that any sufficient reason for our ensemble must have a size of at least $(k+1)$, and therefore: $\langle (f_{1}, \ldots, f_{5}), \x', k' \rangle \notin$ MSR for $k$-ensembles of Perceptrons when $k=5$. In summary, we have demonstrated that $\langle \z, S_{0}, k, T \rangle \in \ckGSSP \iff \langle (f_{1}, \ldots, f_{5}), \x', k' \rangle \in$ MSR for $k$-ensembles of Perceptrons when $k=5$. Thus, solving the MSR query for $k$-ensemble Perceptrons is para-\StoPComplexity-hard.

$\qedsymbol{}$

\section{Proof of Proposition~\ref{main_paper_fbdd_ensemble_paramterized_results}}
\label{xp_proofs_appendix_sec}


\begin{proposition}
\label{xp_proofs_appendix}
For ensembles of $k$-decision trees \begin{inparaenum}[(i)] \item the CSR query is coW[1]-Complete, \item the MCR query is W[1]-Hard and in W[P], \item the CC query is $\#$W[1]-Complete, and \item the SHAP query is W[1]-Hard and in XP.
\end{inparaenum} 
\end{proposition}


\emph{Proof.} We begin by demonstrating the initial complexity result stated in the proposition:

\begin{lemma}
\label{k-ensemble-fbdds-csr}
    For ensembles of $k$-decision trees, the CSR query is coW[1]-Complete.
\end{lemma}

\emph{Proof.} We execute a many-one FPT reduction from the complement of the Multi-Color Clique problem to the corresponding CSR query, and from the CSR query to the complement of the $k$-Clique problem. Both the Multi-Color Clique and the $k$-Clique problems are known to be W[1]-Complete. We will begin with the first direction to establish membership in coW[1]. We start by defining the $k$-Clique problem:


\vspace{0.5em} 

\noindent\fbox{%
    \parbox{\columnwidth}{%
\mysubsection{$k$-Clique}:

\textbf{Input}: A graph $G=\langle V,E\rangle$.

\textbf{Parameter:} Integer $k$.

\textbf{Output}: \yes{}, if there exists a clique of size larger than $k$ in $G$.
    }%
}

\vspace{0.5em} 




\textbf{Membership.} We initiate the reduction by demonstrating membership in coW[1]. As discussed in Section~\ref{appendix:model_types}, our aim is to establish membership for the weighted voting scenario. However, for clarity, we will initially focus on the simpler case of regular majority voting. Subsequently, we will elaborate on how this proof can be extended to the weighted version.

\textbf{Membership for standard majority vote.} We will specifically demonstrate a reduction from $\overline{\text{CSR}}$ to $k$-Clique, establishing that \emph{CSR} resides in coW[1]. Given an instance $\langle f,\x,S\rangle$, $\overline{\text{CSR}}$ inquires whether $S$ is \emph{not} a sufficient reason with respect to $\x$, or in other words, if $\overline{S}$ is contrastive. We will begin by establishing the following claim, which concerns the fact that ensembles of decision trees are closed under conditioning (refer to Definition~\ref{closed_under_conditioning}):

\begin{claim}
\label{ensemble_fbdds_conditioning}
    Given a model $f$ which is an ensemble of $k$ decision trees, some input $\x$, and subset $S\subseteq [n]$, then $f$ is closed under conditioning.
\end{claim}

\emph{Proof.} We first observe that any individual decision tree $f_i$ in the ensemble can be conditioned over $\x_S$. This is accomplished by creating a modified model $f'_i$ as follows: we start by duplicating $f_i$, that is, $f'_i:=f_i$. We then iterate through the tree from the top downwards, examining all splits. Each split corresponds to an assignment to a feature $j$, which can be set to either $0$ or $1$. If $j\in S$, we will retain all paths that extend from the subtree and follow the assignment $\x_j$, while deleting all paths that follow the assignment $\neg \x_j$. Ultimately, the resulting tree $f_i$ adheres to the following condition:

\begin{equation}
    \forall \z\in\mathbb{F} \quad [f_i(\x_S;\z_{\Bar{S}})=f'_i(\z]
\end{equation}

We observe that following the modifications, any split in the tree $f'_i$ concerning the features $j \in S$ results in only one viable path, as the subtree associated with the contrary assignment was removed earlier. Consequently, $f'_i$ effectively becomes a tree defined solely over the features in $\overline{S}$. Therefore, the following assertion is valid:

\begin{equation}
    \forall \z\in\mathbb{F} \quad [f_i(\x_S;\z_{\Bar{S}})=f'_i(\z_{\Bar{S}}]
\end{equation}

We have thus demonstrated that a single decision tree is closed under conditioning. By applying this procedure to each tree within the $k$-ensemble $f$, we arrive at the following conclusion:

\begin{equation}
    \forall \z\in\mathbb{F} \quad [f(\x_S;\z_{\Bar{S}})=f'(\z_{\Bar{S}}]
\end{equation}

$\qedsymbol{}$

We will establish a secondary minor claim that will be beneficial for our reduction:

\begin{claim}
\label{ensemble_fbdds_negating}
    Given a model $f$ which is an ensemble of $k$ decision trees, then $f':=\neg f$ can be constructed in polynomial time.
\end{claim}

\emph{Proof.} We first note that negating a single decision tree can be achieved by switching each leaf node's assignment from ``1'' to ``0'' and vice versa. Applying this modification to any decision tree $f_i$ within the ensemble results in negating the entire model $f$.

Now, employing Claim~\ref{ensemble_fbdds_conditioning}, an ensemble $f$ can be conditioned on a partial assignment of features, allowing us to develop a new model $f'$ that is conditioned on $\x_{S}$. In simpler terms, $f'$ retains only the features from $\overline{S}$ and satisfies the following condition:

\begin{equation}
    \forall \z\in\mathbb{F} \quad [f(\x_S;\z_{\Bar{S}})=f'(\z_{\Bar{S}})]
\end{equation}

Given the ensemble $f=(f_1, f_2, \ldots, f_k)$, the reduction constructs $f'=(f'_1, f'_2, \ldots, f'_k)$ and uses it to construct a graph $G$. The final setup of the reduction is $\langle G, k':=\ceil{\frac{k}{2}} \rangle$. We will now detail the construction of the graph $G$. First, we compute $f'(\x_{\Bar{S}})$, which is equivalent to $f(\x_S; \x_{\Bar{S}}) = f(\x)$. If $f'(\x_{\Bar{S}})=1$, we negate the ensemble $f'$. This negation is carried out using Lemma~\ref{ensemble_fbdds_negating}. Thus, we can generally assume that $f'(\x_{\Bar{S}})=0$.


The graph we construct will be a $k$-partite graph, with each part corresponding to each tree in the ensemble $f'$. The vertices and edges of the graph are constructed as follows: We iterate over the leaf nodes in each tree of $f'$, and every leaf node corresponding to the assignment of $\neg f'(\x)$ is designated as a vertex $v_i$ in the graph (associated with the specific tree this node is part of). We will now proceed to describe the construction of the edges of the graph.

First, it is important to note that any two vertices within the same part of the graph (i.e., associated with the same tree) will not be connected by an edge. We iterate over any two paths in two \emph{distinct} trees --— and hence associated with two different parts of the graph. We consider two paths, $\alpha$ and $\alpha'$, from different trees in $f'$ to ``match'' if they do not ``collide'' on any variable. Specifically, there should be no feature $i$ associated with both $\alpha$ and $\alpha'$ where the assignment of $i$ in $\alpha$ is $\x_i\in\{0,1\}$ and the assignment in $\alpha'$ is $\neg \x_i$. The two paths ``match'' if there is no such collision.


Now, each vertex $v_i$ in the graph is linked to a particular leaf node in one of the trees (and consequently to the path $\alpha$ that leads to this leaf node). The reduction will establish an edge between any two vertices $v_i$ and $v_j$, which correspond to paths $\alpha$ and $\alpha'$ from two separate trees, if and only if the paths $\alpha$ and $\alpha'$ ``match'' and both paths conclude at a terminating node with a True assignment, that is, classified as $1$.


We will now prove that $S$ is not a sufficient reason with respect to $\langle f,\x\rangle$ if and only if there exists a clique in the graph of size larger than $k':=\ceil{\frac{k}{2}}$. The proof will be divided into several distinct claims. First, we will establish the following claim:


\begin{claim}
    For the aforementioned reduction construction, $S$ is not a sufficient reason concerning $\langle f,\x\rangle$ if and only if there exists a partial assignment $\z_{\Bar{S}}$ for which $f'(\z_{\Bar{S}})=1$.
\end{claim}

\emph{Proof.} By definition, $S$ is not a sufficient reason for $\langle f,\x\rangle$ if and only if:

\begin{equation}
    \exists \z\in\mathbb{F} \quad [f(\x_S;\z_{\Bar{S}})\neq f(\x)]
\end{equation}

This equivalently means that $\overline{S}$ is contrastive with respect to $\langle f,\x\rangle$. Given that $f'$ is conditioned on $\x_S$ and includes only the features from $\Bar{S}$, it equivalently follows that $S$ is not a sufficient reason concerning $\langle f,\x\rangle$ iff:


\begin{equation}
\label{equation_proerty_to_ref}
    \exists \z\in\mathbb{F} \quad [f'(\z_{\Bar{S}})\neq f'(\x_{\Bar{S}})= 0] \iff 
    \exists \z\in\mathbb{F} \quad [f'(\z_{\Bar{S}})= 1] 
\end{equation}

This equation holds based on our assumption about the negation of $f'$ during its construction. Additionally, the condition where $f'(\z_{\Bar{S}})=1$ occurs if and only if at least $\ceil{\frac{k}{2}}$ models in the ensemble $f'$, specifically ${f'1,\ldots, f'{\ceil{\frac{k}{2}}}}$, have $f'i(\z_{\Bar{S}})=1$.

We will begin by establishing a smaller claim.

\begin{claim}

    For the aforementioned reduction construction, the following condition is satisfied: there exists some $\z\in\mathbb{F}$ for which $f'(\z_{\Bar{S}})=1$ if and only if there exists a subset of $j\geq \ceil{\frac{k}{2}}$ trees: ${f'_1,\ldots,f'_j}$, in which it is possible to select one path $\alpha_i$ in each tree $f'_i$ such that each path terminates at a ``True'' (classification $=1$) node, and each pair of distinct paths ``match''.
\end{claim}

\emph{Proof.} We first observe that upon proving this lemma, it will be established that the claim (the existence of $j\geq \ceil{\frac{k}{2}}$ trees satisfying the aforementioned property) holds if and only if $S$ is not a sufficient reason for $\langle f,\x\rangle$. This is a direct outcome of the equivalence property mentioned in equation~\ref{equation_proerty_to_ref}.

Let us assume we have a set of $j$ different paths ${\alpha_1,\ldots, \alpha_j}$ within $j$ distinct trees: ${f''_1,\ldots, f''_j}$, where each $f''_i$ is a model from the ensemble ${f_1,\ldots, f_k}$, with each path chosen within one of the trees. All paths are selected such that they terminate on a ``True'' node (assignment $1$) and every pair of paths from two distinct trees ``matches''. According to the definition of ``matching'' paths, this means that there is no feature $i$ for which paths in two different trees disagree on the assignment of that feature. Consequently, we can adopt the partial assignment $\z_i$ for each feature $i$ that is assigned a value in one of these paths. Let us denote this partial assignment as $\z_{S'}$ for some $S'\subseteq \overline{S}$. It therefore, follows that if we fix the features in $S'$ to their values in $\z$, the prediction of each one of the distinct trees: ${f''_1,\ldots,f''_j}$ will be classified as $1$. This is equivalent to stating that for any $S''\subseteq \Bar{S'}$, the following holds:


\begin{equation}
\begin{aligned}
    \forall \z'\in\mathbb{F} \quad [f''_1(\z_{S''};\z'_{\Bar{S}\setminus S''})=f''_2(\z_{S''};\z'_{\Bar{S}\setminus S''})=\ldots=f''_j(\z_{S''};\z'_{\Bar{S}\setminus S''})= 1] \implies \\
    \exists \z'\in\mathbb{F} \quad [f''_1(\z_{S''};\z'_{\Bar{S}\setminus S''})=f''_2(\z_{S''};\z'_{\Bar{S}\setminus S''})=\ldots=f''_j(\z'_{S''};\z'_{\Bar{S}\setminus S''})= 1] 
    \end{aligned}
\end{equation}


If we consider $j \geq \ceil{\frac{k}{2}}$, then it clearly follows that:

\begin{equation}
    \exists \z'\in\mathbb{F} \quad [f'(\z_{S''};\z'_{\Bar{S}\setminus S''})= 1] 
\end{equation}

Therefore, we can assign $\y_{\Bar{S}}:=(\z_{S''};\z'_{\Bar{S}\setminus S''})$ and it will be established that:

\begin{equation}
    \exists \y\in\mathbb{F} \quad [f'(\y_{\Bar{S}})= 1] 
\end{equation}

For the other direction, let us assume that there is \emph{no} set of $j\geq \ceil{\frac{k}{2}}$ trees that terminate at a $1$ ``True'' assignment, such that there is a viable choice of path $\alpha_i$ in each tree $f''_i$ where each pair of distinct paths in this set of trees ``matches''. From this assumption, it follows that there is \emph{no} group of $j\geq \ceil{\frac{k}{2}}$ trees for which a partial assignment $\z_{S'}$, with $S'\subseteq \overline{S}$, can be fixed to $\z$ ensuring that the prediction of all the trees ${f''_1,\ldots, f''_j}$ will remain $1$. More specifically, this indicates that there is no set of $j\geq \ceil{\frac{k}{2}}$ trees for which an assignment to $\overline{S}$ guarantees that the prediction of all ${f''_1,\ldots, f''_j}$ trees will remain $1$.



Since there are no $j\geq \ceil{\frac{k}{2}}$ trees where an assignment to $\overline{S}$ leads to all these trees predicting $1$, the value of $f'$ will consistently be $0$ for any possible assignment to the features in $\overline{S}$. Therefore, it is established that:

\begin{equation}
    \forall \y\in\mathbb{F} \quad [f'(\y_{\Bar{S}})= 0] 
\end{equation}

$\qedsymbol{}$

This concludes this segment of the proof. We have now demonstrated that $S$ is not a sufficient reason concerning $\langle f,\x\rangle$ if and only if there exists a subset of $j\geq \ceil{\frac{k}{2}}$ trees, each with a different path that finishes at a positive terminal node, and where each pair of paths ``match''. We will now proceed to prove the following claim, which will conclude our general proof:


\begin{claim}
    In the aforementioned reduction construction, there is a clique of size greater than $k' \geq \ceil{\frac{k}{2}}$ in $G$ if and only if there exist $k'$ distinct trees in $f'$ with $k'$ distinct paths (one per tree) that end at a True ``1'' node, and each pair of distinct paths ``match''.
\end{claim}

\emph{Proof.} Assuming that there are $k' \geq \ceil{\frac{k}{2}}$ paths in $k'$ distinct trees that ``match'' and end at a $1$ (True) node, the reduction construction implies that each of these paths corresponds to a vertex in $G$ (as they terminate on a $1$ node). Furthermore, given that these paths "match" according to our construction, there will be an edge connecting each pair of vertices. Consequently, the set of these $k'$ vertices forms a clique in $G$, establishing the existence of a clique of size $k'$ in $G$.


For the second direction, suppose there is no subset of $k'$ trees or more. This equivalently means that for any subset of $j \geq \ceil{\frac{k}{2}}$ paths chosen from $j$ distinct trees that end on a $1$ node, there exists at least one pair of distinct paths that do \emph{not} match. According to our construction, this implies that the subgraph $G' \subseteq G$, corresponding to these vertices (each representing one of the paths), is not a clique. Therefore, it follows that for any subgraph $G' \subseteq G$ with $\ceil{\frac{k}{2}}$ or more vertices, $G'$ is not a clique. This completes the reduction.


\textbf{Membership for weighted Vote.} In our previous proof, we conditioned $f$ on the partial assignment $\x_S$ to derive the model $f'$. We demonstrated that $S$ not being sufficient with respect to $\langle f,\x\rangle$ equates to a satisfying assignment in $f'$. We will now extend this proof to the weighted version, where a satisfying assignment does not necessarily correspond to a set of $\ceil{\frac{k}{2}}$ trees with a ``1'' classification. In this version, each tree $f'_i$ is associated with a weight $\phi_i$, and $f'$ is defined as follows:


\begin{equation}
f'(\x):=step(\sum_{1\leq i\leq n} \phi_i\cdot f'_i(\x))
\end{equation}

Therefore, we can implement the following procedure: Iterate over the power set of all possible trees (representing all potential sets of different trees), which includes iterating over subsets $S' \subseteq \{1, \ldots, k\}$. This enumeration is bounded by $O(2^k)$ and thus can be performed in FPT time. For each selected combination $S'$, representing a choice of $|S'|$ distinct trees, we check whether:

\begin{equation}
\label{weighted_voting_check}
step(\sum_{i \in S'} \phi_i\cdot f'_i(\x))>0
\end{equation}

This corresponds to checking whether:

\begin{equation}
\label{weighted_voting_check_2}
step(\sum_{i \in S'} \phi_i\cdot f'_i(\x))=  step(\sum_{i \in S'} \phi_i\cdot f'_i(\x)+ \sum_{i \in \Bar{S'}} \phi_i\cdot f'_i(\x))=step(\sum_{1\leq i\leq n} \phi_i\cdot f'_i(\x))>0
\end{equation}

Thus, we only need to ``check'' subsets $S'$ where equation~\ref{weighted_voting_check} is satisfied (as these alone correspond to a positive instance of $f'$). In our reduction, for each subset, $S'$ satisfying equation~\ref{weighted_voting_check}, we construct a subgraph $G'$ in the same manner as in the previous reduction: each path for each tree associated with $S'$ becomes a vertex in $G'$, and an edge between two vertices is formed between two edges iff two distinct paths "match". Additionally, we add $k-|S'|$ extra vertices to each such graph, which are connected to all other vertices in $G'$. We then construct $G$ as the union of all such sub-graphs $G'$ that were derived from each $S'$, and the reduction results in $\langle G, k \rangle$.


Previously, in the classic majority-vote scenario, we demonstrated that a positive assignment in $f'$ corresponds to a clique in $G$, with each vertex in the clique representing its associated path in $f'$. Thus, in our current construction, if there is a positive assignment to $f'$, then there exists some subset $S'$ such that:


\begin{equation}
\label{weighted_voting_check}
step(\sum_{i \in S'} \phi_i\cdot f'_i(\x))>0
\end{equation}

This implies that there is a subset $S'$ of distinct trees within a subgraph $G'$ that forms a clique of size $|G'|$ (where each vertex in $G'$ corresponds to a path in $f'$). Since these vertices are also connected to an additional $k-|S'|$ vertices included in our construction, there is also a corresponding clique of size $k$ within $G'$, and thus in $G$ as well.


However, if a positive assignment for $f'$ does not exist, it indicates that for any subset of trees $S'$ in $f'$, there is no corresponding clique of size $|S'|$ where each vertex corresponds to a path in $f'$. This implies that any clique is of size less than $k-|S'|+|S'|$. Since this holds for any subgraph $G'$ within $G$, it follows that there is no clique of size $k$ in $G$, thereby concluding the reduction.

$\qedsymbol{}$


\textbf{Hardness.} To demonstrate that CSR for a $k$-ensemble of decision trees is coW[1]-Hard, we will establish a many-to-one FPT reduction from the complementary version of the multi-color clique problem, which is known to be W[1]-Complete.


\vspace{0.5em} 

\noindent\fbox{%
    \parbox{\columnwidth}{%
\mysubsection{Multi Color Clique}:

\textbf{Input}: A graph $G=\langle V,E\rangle$, such that $V:=\langle V_1,\ldots, V_k\rangle$ where each $V_i$ denotes a set of distinct vertices of some color, for which any two vertices associated with a color are not neighbors (for all $i$ there is no edge $(u,v)\in E$ where $u,v\in V_i$).

\textbf{Parameter:} $k$ (the number of colors).

\textbf{Output}: \yes{}, if there exists a clique of size $k$ in $G$.
    }%
}

\vspace{0.5em} 

Let us consider an instance $\langle G,k\rangle$, where $G$ is a multi-colored graph with $k$ distinct colors. We can assume that $|V_1| = |V_2| = \ldots = |V_k| = m$. This assumption is valid because we can take the set with the maximum number of vertices, denoted $maxV$, and ``pad'' the parts of the graph with fewer than $maxV$ vertices by adding extra vertices that are unconnected (thereby not affecting the size of any potential clique). Consequently, the total number of vertices is $m \cdot k$.


The reduction constructs a model $f$, which is an ensemble of decision trees, in the following manner: Initially, each vertex is associated with a unique binary string of length $log(m)$, ensuring that vertices of the same color have different strings. We then iterate over pairs of distinct colors, ranging from $1 \leq i < j \leq k$. For each pair, we create a tree $f_{i,j}$, which is a complete binary tree with a depth of $2log(m)$. This tree comprises features ${x^i_1, x^i_2, \ldots, x^i_{log(m)}}$ (representing all possible assignments for the tree associated with color $i$) and ${x^j_1, x^j_2, \ldots, x^j_{log(m)}}$ (representing all possible assignments for the tree associated with color $j$). Each of the $2log(m)$ nodes in the tree corresponds to an assignment of all features of $x^i$ and $x^j$. Each terminal node in the tree, which corresponds to some path, represents a pair of vertices in colors $i$ and $j$. If there is an edge between these two vertices, this terminal node is marked with a $1$ (a ``True'' assignment); if not, it is marked with a $0$ (a ``False'' assignment).


Now, for each constructed tree (totaling $\binom{k}{2}$ trees), we also construct an additional ``dummy'' tree that consistently returns ``False'' (assignment $0$). Consequently, the total number of trees in the constructed ensemble is $\binom{k}{2} \cdot 2$.


We initially observe that we can assume the existence of at least one pair of vertices $(u,v)$ from different colors $i,j$ that are \emph{not} adjacent (if all pairs were adjacent, there would trivially exist a clique of size $k$). Therefore, we select an assignment where $f_{i,j}$ reaches a ``False'' ($0$) terminal node. Arbitrary assignments can be chosen for all other features. Let us denote this complete feature assignment by $\x$. Given that there is at least one tree among the first $\binom{k}{2}$ trees that results in a ``False'' ($0$) classification, and all the additional $\binom{k}{2}$ ``dummy'' trees are designed to reach a ``False'' ($0$) outcome, the majority of trees in $f$ will classify $\x$ as $0$. Thus, it is established that $f(\x) = 0$. The final structure of the reduction is $\langle f, \x, S:=\emptyset \rangle$. Notably, the number of models in $f$ is $2\cdot \binom{k}{2}$, setting the parameter for the instance $\langle f, \x, S:=\emptyset \rangle$ at $k':= 2\cdot \binom{k}{2}$, which is within the bounds of a computable function $g(k)$, thereby maintaining the FPT reduction.


We first note that this reduction operates in FPT time, as the size of $f$ is capped by $log(n)\cdot O(k^2)$, which is naturally bounded by $O(g(k)\cdot n^k)$ for some computable function $g$. We will now demonstrate that a multi-colored clique of size $k$ or greater exists in $G$ if and only if $S:=\emptyset$ is not a sufficient reason with respect to $\langle f,\x\rangle$.


Assume there exists a multi-colored clique of size $k$, denoted as $G'$, within $G$. This implies that for any two vertices $u,v$ in $G'$, $(u,v)\subseteq E$. Given that no edges exist between vertices of the same color, selecting two distinct colors $i,j$ ensures that there is at least one edge connecting a vertex from color $i$ to a vertex from color $j$. Therefore, for each tree $f_{i,j}$ associated with a pair of two different colors $i, j$, we can select the corresponding path that aligns with the edge connecting these two vertices, and which terminates at a ``True'' ($1$) leaf node, as per our construction.


We will now prove that when we select these specific paths, any two pairs of distinct paths ``match''. This occurs due to the following reason: Consider two paths chosen from two distinct trees. Initially, assume the first tree $f_{i,j}$ is associated with a pair of colors $i,j$ and the second tree $f_{k,l}$ is associated with a pair $k,l$, where $i\neq j\neq k \neq l$. Since the features of the tree $f_{i,j}$ are not shared with those of the tree $f_{k,l}$, any two paths from these trees ``match'' (they do not conflict over any feature assignment). The more complex scenario arises when two distinct trees, $f_{i,j}$ and $f_{j,k}$, involve the colors $i,j$ and $j,k$, where $i\neq j \neq k$. In this case, the features corresponding to $i, k$ are different, but those associated with color $j$ are shared. However, because a clique of size $k$ in a $k$-partite graph includes exactly one vertex from each color, the vertex corresponding to color $j$ associated with the paths chosen for both $f_{i,j}$ and $f_{j,k}$ is the same. Consequently, the binary string representing the path associated with these features is identical, ensuring these two paths do not conflict over any feature assignment. Thus, in any scenario, any pair of distinct paths selected in this manner for these trees ``match''.


Therefore, we can assign each of the features in these paths to their respective values within the path. This is feasible because all these paths ``match'' and have no conflicting assignments. Given that all of these trees are complete, all features can be assigned (i.e., this constitutes a full assignment, not a partial one). We will denote this assignment by $\z \in \mathbb{F}$. As previously noted, and according to our construction, all of these paths terminate at a ``True'' ($1$) leaf node. Consequently, the assignment $\z$ results in the first $\binom{k}{2}$ trees receiving a ($1$) assignment. Since exactly half of the trees in the ensemble are assigned a value of $1$, the overall classification by $f$ is $1$. In summary, there exists an assignment $\z$ for which:


\begin{equation}
    \exists \z\in\mathbb{F} \quad [f(\z)= 1 \neq f(\x) = 0] 
\end{equation}

If no clique of size $k$ exists in $G$, then any subgraph $G'$ of $G$ containing $k$ vertices is \emph{not} a clique. Consider such a subgraph $G'$ with $k$ vertices. Since it is not a clique, this means there must be at least one pair of vertices $(u, v)$ within it that are not connected by an edge.


Consider some assignment $\z\in\mathbb{F}$. When examining the specific path associated with $f_{i,j}(\x)$ (for $1\leq i,j\leq k$), it follows a designated path leading to a specific terminal node of the tree $f_{i,j}$. Additionally, any pair of distinct paths from two distinct trees associated with $\z$ necessarily ``match'' (since otherwise, they would not correspond to a non-contradicting assignment). Suppose, for the sake of contradiction, that each of these paths ends at a ``True'' (``1'' classification) node.


For each tree $f_{i,j}$, we can identify the pair of vertices $(u,v)$ that correspond to the path representing the binary string of that path. This involves selecting vertices $(u,v)$ associated with two distinct colors $i,j$. Consequently, this specific assignment $\z$ gives rise to a subgraph $G'$, which includes at least one vertex from each color (since every tree $f_{i,j}$ is associated with two vertices —-- one for each color). However, we will now prove that there must be \emph{exactly} one vertex from each color in $G'$. Assume, for the sake of contradiction, that there are two vertices $v_1,v_2$ of the same color $i$ in $G'$. Since $v_1,v_2$ are in $G'$, there exist (without loss of generality) two distinct trees $f_{i,j}$ and $f_{i,k}$ where $i\neq k$, and the path associated with $f_{i,j}$ includes vertex $v_1$, while the path associated with $f_{i,k}$ includes vertex $v_2$. This configuration implies that the paths for $f_{i,j}$ and $f_{i,k}$ do not ``match'' (as the vertex chosen for color $i$ in these two trees differs and is associated with a different binary string).


We have established that $G'$, the graph associated with a specific assignment $\z$, contains exactly one vertex from each color in $G$, and therefore has a size of $k$. From the assumption of this direction in the reduction, this indicates that $G'$ is \emph{not} a clique. Consequently, this also means there must be two vertices $(u,v)$ from two different colors $i\neq j$ where $(u,v)\not\subseteq E$. In terms of our reduction construction, this means that examining $f_{i,j}$, the binary string associated with vertices $u,v$ leads to a ``False'' ($0$) terminal node, which contradicts our initial assumption.


Thus, we have demonstrated that if there is no clique in $G$ of size $k$, then for any arbitrary assignment $\z$, not all of the first $\ceil{\binom{k}{2}}$ trees in $f$ receive a $1$ assignment. In simpler terms, at least one tree receives a $0$ assignment. Consequently, in the ensemble, any assignment results in at least $\ceil{\binom{k}{2}}+1$ trees being assigned $0$, ensuring that the ensemble classification is always $0$. In other words:


\begin{equation}
    \forall \z\in\mathbb{F} \quad [f(\z)= 0 = f(\x)] 
\end{equation}

This indicates that $S:=\emptyset$ is a sufficient reason with respect to $\langle f,\x\rangle$, thereby concluding our reduction.

$\qedsymbol{}$

\begin{lemma}
\label{cc_appendix_k_ensemble_fbdd_proof}
    The CC query for a $k$-ensemble of decision trees is \#W[1]-Complete.
\end{lemma}

\emph{Proof.} We note that the CC query is the counting version for the CSR query, for which we already proved coW[1]-Completness. The proofs of membership and hardness where from the $k$-Clique, and $k$-Multicolored-Clique problems. It is well known that the counting version of $k$-Clique is \#W[1]-Complete~\cite{flum2004parameterized}. Moreover, there exist FPT reductions to and from $k$ Multi-Color Clique to $k$-Clique, which shows us that Multicolored clique is also \#W[1]-Complete. Finally, we arrive at that the CC query for a $k$ ensemble of decision trees is \#W[1]-Complete.

$\qedsymbol{}$

\begin{lemma}
    The MCR query for a $k$-ensemble of decision trees is W[1]-Hard and in W[P].
\end{lemma}

\emph{Proof.} \textbf{Hardness}. Specifically, the hardness results are consistent with those presented by~\cite{ordyniak2024explaining}. However, we can also directly prove hardness by presenting a reduction from the previous (complement of) the \emph{CSR} problem for ensembles of decision trees, which we have shown to be coW[1]-Complete (via FPT reductions from $k$-Clique and $k$-Multicolored Clique), which will prove hardness beyond majority-voting ensembles. The complement of the \emph{CSR} problem is equivalent to validating whether, given some $f, \x$, and a subset $S$, it can be checked whether $S$ is \emph{not} a sufficient reason for $\langle f, \x \rangle$. This is equivalent to checking whether $\overline{S}$ is a contrastive reason for $\langle f, \x \rangle$.


Hence, given an instance $\langle f, \x, S \rangle$, we can apply Lemma~\ref{ensemble_fbdds_conditioning}, which states that ensembles of decision trees are closed under conditioning. We will condition $f$ on $\x_S$ to construct $f'$. The resulting model $f'$ will have $|\overline{S}|$ features, and it holds that:


\begin{equation}
    \forall \z\in\mathbb{F} \quad [f'(\z_{\Bar{S}})=f(\x_S;\z_{\Bar{S}})] 
\end{equation}

Now, given the instance $\langle f, \x, S \rangle$, the reduction will construct: $\langle f', \x, k := |\overline{S}| \rangle$. If $S$ is not a sufficient reason for $\langle f, \x \rangle$, this implies that $\overline{S}$ is a contrastive reason for $\langle f, \x \rangle$, and therefore:


\begin{equation}
    \exists \z\in\mathbb{F} \quad [f(\x_S;\z_{\Bar{S}})\neq f(\x)] 
\end{equation}

This further implies that:

\begin{equation}
    \exists \z\in\mathbb{F} \quad [f'(\z_{\Bar{S}})= f(\x_S;\z_{\Bar{S}})\neq f(\x)=f'(\x_{\Bar{S}})] 
\end{equation}

which indicates that $\overline{S}$ is a contrastive reason for $\langle f', \x \rangle$. Therefore, there exists a contrastive reason of size $|\overline{S}|$ for $\langle f', \x \rangle$. If we assume that $S$ is a sufficient reason for $\langle f, \x \rangle$, then it holds that:



\begin{equation}
\begin{aligned}
    \forall \z\in\mathbb{F} \quad [f(\x)=f(\x_S;\z_{\Bar{S}})] \iff \\
    \forall \z\in\mathbb{F} \quad [f'(\x_{\Bar{S}})=f(\x)=f(\x_S;\z_{\Bar{S}})=f'(\z_{\Bar{S}})] 
    \end{aligned}
\end{equation}

This indicates that $\overline{S}$ is not a contrastive reason for $\langle f', \x \rangle$, and that any subset $S' \subseteq S$ is also not a contrastive reason for $\langle f', \x \rangle$. Hence, it follows that there is no contrastive reason of size $|\overline{S}|$ or smaller for $\langle f', \x \rangle$. This completes the reduction, thereby proving that the \emph{MCR} query for an ensemble of $k$ decision trees is W[1]-Hard.


\textbf{Membership.} We will prove membership in W[P] by reducing the MCR query for $k$-ensemble decision trees to the WCS[$C_{d,t}$] problem, as described in Section~\ref{parametrieed_background}. Given an instance $\langle f,\x,D\rangle$, where $D$ represents the size of the contrastive reason we are looking for, we construct a Boolean circuit $C$. Although the weft of $C$ is 2 (and could be reduced to 1 with a more refined construction), its depth will depend on a parameter $D$ and will not be bounded by a constant. 


Our reduction begins by creating a modified model $f'$ based on the original model $f$ and the input $\x$. Essentially, $f$ and $f'$ will maintain the same structure; however, the vector $\mathbf{1}_n$ (consisting solely of $1$s) in $f'$ will correspond to the vector $\x$ in $f$, and conversely, the vector $\neg\x$ in $f$ will correspond to the vector $\mathbf{0}_n$ (consisting solely of $0$s) in $f'$.


To carry out this construction, we start by replicating $f$ to create $f'$. For each decision tree $f_i$ in the ensemble $f$, we examine every node $v$ within $f_i$. For each node assignment $v_i\in\{0,1\}$ where $v_i \neq \x_i$, we reverse the $0$ and $1$ assignments, with the typical convention that $1$ represents the right branch and $0$ the left branch. This flipping will be done such that $1$ will now correspond to the left branch. If $v_i=\x_i$, we retain the original order. This process is repeated across all paths in each decision tree of the ensemble. As a result, we generate a new model $f'$ where each assignment to a value of $\x_i$ in $f$ corresponds to an assignment to $1$ in $f'$. If $f(\x)=1$, we can apply the negation principle using Lemma~\ref{ensemble_fbdds_negating} to negate $f'$, and if $f(\x)=0$, we can leave it as is. Consequently, any vector $\z\in\mathbb{F}$ where $f(\z)\neq f(\x)$ translates to a vector $\z'\in\mathbb{F}$ for which $f'(\z)\neq 1$. Therefore, the problem of determining whether there exists a subset $S$ of size $D$ such that $f(\x_{\Bar{S}};\z_S)\neq f(\x)$ equates to finding a vector $\z'\in\mathbb{F}$ with $D$ assignments to $1$ (i.e., of Hamming weight $D$) where $f'(\z')\neq 1$.

Now, with $f'$ in place, we will develop a Boolean circuit $C$, as outlined earlier. Specifically, we will create a Boolean circuit $C$ with a weft of $2$ and arbitrary depth, designed to return True if $f'$ has a positive assignment of Hamming weight $D$, and False otherwise. The construction will proceed through the following steps:

\begin{enumerate}
    \item In our ensemble of $k$ trees, for each leaf $v$ in every tree $f'_i$, we will designate $y_{\{v,f'_i\}}$ as an input node in the circuit. We will consider two input nodes, $y_{\{v,f'_i\}}$ and $y_{\{v',f'_j\}}$, as inconsistent if $i=j$ and $v\neq v'$. In such cases, we will introduce a node: $[\neg y_{\{v,f'_i\}}] \vee [\neg y_{\{v',f'_i\}}]$, which is equivalent to $\neg [y_{\{v,f'_i\}} \wedge y_{\{v',f'_i\}}]$, where both $y_{\{v,f'_i\}}$ and $y_{\{v',f'_i\}}$ are input nodes. This setup essentially encodes the $k$-\emph{Clique} problem, except for the final AND encoding involving all input nodes, which we will address in the last step. Recalling our proof for the \emph{CSR} query, the reduction to the $k$-\emph{Clique} problem assists in determining whether a positive assignment exists for $f'$. We are now tasked with a more challenging problem: determining whether there is a positive assignment to $f'$ with a Hamming weight of $d$. This necessitates the inclusion of additional constraints.
    
    \item To this circuit $C$, currently encoding the Clique problem, we will introduce more nodes. For each feature $i \in [n]$, we will create a new node $u_i$ functioning as an OR gate. This node will take inputs from any $y_{\{v,f'_j\}}$ where the assignment represented by the leaf $v$ assigns ``True'' (i.e., a $1$ assignment) to feature $i$.
    \item For the final component, we will add another layer to our circuit. For every $1 \leq j \leq n$ and for every $0 \leq d' \leq D$, we define a variable $u_{\{j,d'\}}$. This variable is configured to be set to True if and only if exactly $d'$ of the features $1,\ldots,j$ are set to True. Specifically, $u_{\{1,0\}}$ will take $\neg u_1$ as its input, and $u_{\{1,1\}}$ will take $u_1$ as its input. All other $u_{\{1,d'\}}$ variables are set to False. For $j > 1$, we construct $u_{\{j,d'\}}$ to take the input:
    

\begin{equation}
    [ \ u_{\{j-1,d'-1\}} \wedge u_{j} \ ] \vee [ \ u_{\{j-1,d'-1\}} \wedge \neg u_j \ ]
\end{equation}

\item Finally, the output of the circuit is derived from a large AND node that collects inputs from all nodes established in step $2$ and those from step $4$ of the form $u_{{n,d'}}$, where $d'$ ranges from $0$ to $D$.
\end{enumerate}

Since that circuit $C$ effectively encodes a scenario where $f'$ receives a positive assignment of Hamming weight $D$, and considering that $C$ possesses a weft of $2$ and arbitrary depth, we conclude that the \emph{MCR} query for ensembles of decision trees is in W[P].

$\qedsymbol{}$

\textbf{The W[1]-W[P] gap for the MCR query.} We note that while we have proven the \emph{MCR} query for ensembles of decision trees is W[1]-Hard and belongs to W[P], the exact complexity class for which this problem is complete remains unknown. Unfortunately, the current definition of the W-hierarchy is not well-suited to capture the complexity of this problem. Specifically, containment in W[t] for a fixed constant $t$ means that the problem instance at hand should be encoded using a Boolean circuit of constant depth (as well as weft $t$), so that the instance at hand is a yes-instance if and only if the circuit has a satisfying assignment having exactly $k$ variables assigned true, where $k$ is the parameter (or $k$ is bounded by a function of the parameter) of the instance at hand. We refer to Chapter 13.3 in \cite{cygan2015parameterized} for more information. However, our problem involves a ``weight measure'' $d$ (the Hamming distance) that may be arbitrarily larger than $k$. In particular, encoding that a potential solution to the instance at hand assigns true to at most $d$ variables cannot be done with a constant-depth circuit, as this requires the implementation of a counter. 

The problem described above is a fundamental one in the definition of the W-hierarchy in the context of \emph{ weighted problems in general}, where the desired total weight of the solution is not bounded by the parameter. Indeed, it is not known how to classify basic W[1]-hard problems in the field such as Weighted $k$-Clique, where given a vertex-weighted graph, along with integers $k$ and $w$, we seek a clique of size exactly $k$ and weight at least/at most $W$, and the parameter is $k$. The same situation holds for problems where the ``weight measure'' is implicit similarly to our problem, such as the Partial Vertex Cover problem, where given a (non-weighted) graph $G$ and integers $k$ and $t$, we seek a set of vertices of size exactly $k$ that altogether cover at least $t$ edges. From personal communications with other researchers in the field, we have gathered that the definition and study of a ``Weighted W-Hierarchy'' can be a topic of independent interest in parameterized complexity, but this is outside the scope of this paper.

\begin{lemma}
    The SHAP query for a $k$-ensemble of decision trees is \#W[1]-Hard, and in XP.
\end{lemma}

\emph{proof.} \textbf{Hardness.} We begin by proving that this problem is \#W[1]-Hard. Our approach follows the proof outlined by~\cite{arenas2023complexity}, which demonstrates a link between the computation of Shapley values and the model counting problem. First, we provide a definition of the model counting problem. Given a model $f: \{0,1\}^n \to \{0,1\}$, we denote $\#f$ as the number of assignments where $f$ outputs $1$. In other words:



\begin{equation}
    \#f:=|\{\z\in\mathbb{F}, f(\z)=1\}|
\end{equation}

Now, the work of~\cite{arenas2023complexity} demonstrates the following relationship when the feature distribution $\mathcal{D}_p$ is assumed to be the \emph{uniform distribution}, which is a specific case of the fully factorized distribution considered in this paper. For any $f, \x$, it holds that:


\begin{equation}
    \#f:=f(\x)-2^n\cdot \sum_{i\in \{1,\ldots,n\}} \phi_i(f,\x)
\end{equation}

This provides direct proof that computing the Shapley value under a uniform distribution (and consequently for the fully factorized distribution) is as difficult as the model counting problem. It remains to be shown that the model counting problem for an ensemble of $k$-decision trees is \#W[1]-Complete, when parameterized by $k$. Since we have already established that the \emph{CC} query for a $k$-ensemble of decision trees is \#W[1]-Complete, we can demonstrate an FPT many-one reduction from the model counting problem to the computation of the \emph{CC} query.



Given a model $f$, we start by selecting an arbitrary vector $\x \in \mathbb{F}$. If $f(\x) = 1$, the reduction calculates the count for $\langle f, \x, S := \emptyset \rangle$, which we denote as $\#CC(f, \x, S)$. If $f(\x) = 0$, the reduction computes $2^n - \#CC(f, \x, S)$. We note that $2^n$ can be computed in polynomial time, as $n$ (representing the number of input assignments) is provided in unary.


The completion count seeks the number of assignments where the complement of $S$ (which, in this case, includes all possible assignments) results in a classification of $1$. Thus, if $f(\x) = 1$, we have $\#f = \#CC(f, \x, S)$, and if $f(\x) = 0$, we have $\#f = 2^n - \#CC(f, \x, S)$, concluding the reduction.


\textbf{Membership.} We will prove membership in \emph{XP} by presenting an algorithm that computes Shapley values for ensembles of $k$-decision trees in $O(|\mathcal{X}|^k)$ time. We utilize the following relation, established by~\cite{van2022tractability}, concerning Shapley values under conditional expectations and assuming feature independence. The following relation has been proven:

\begin{equation}
    \text{SHAP}(f,i,\mathcal{D}_p,\x)=_P \mathbb{E}_{\z\sim\mathcal{D}_p}[f(\z)]
\end{equation}

In other words --- the computational complexity of obtaining a Shapley value under this formalization is equivalent (under polynomial reductions) to the complexity of computing $\mathbb{E}_{\z \sim \mathcal{D}p}[f(\z)]$. This means we can focus on determining the complexity of obtaining $\mathbb{E}{\z \sim \mathcal{D}_p}[f(\z)]$, which can sometimes be easier to handle.


We now present the following algorithm for computing $\mathbb{E}_{\z \sim \mathcal{D}_p}[f(\z)]$ for ensembles of decision trees in $O(|\mathcal{X}|^k)$ time. The algorithm iterates over every combination of selecting one path from each tree in the ensemble $f$. We assume that $j$ of these selected paths correspond to trees that classify as ``1'', while the remaining $k-j$ trees classify as ``0''. We first check whether this selection of $j$ trees ${f'_1, \ldots, f'_j}$ that classify as ``1'' can result in a positive classification for $f$. For unweighted majority voting, this is equivalent to verifying whether $j \geq \ceil{\frac{k}{2}}$, and for weighted voting, it involves checking whether the weights $\phi_i$ associated with each tree $f'i$ satisfy $\sum{1 \leq i \leq j} \phi_i > 0$.


We can then verify whether each pair of paths chosen from the set of paths across all trees satisfies that any two paths ``match'' (i.e., they do not contain any features with conflicting assignments). For each combination of trees, we examine the partial assignment to all features involved in these paths (there must be only one such assignment since the paths ``match''). In practice, this is equivalent to iterating over all possible positive assignments of the ensemble $f$. However, we note that a single iteration over a selection of paths in each tree may yield only a partial assignment to some features, denoted $\z_{S'}$ for some $S' \subseteq [n]$. As a result, this assignment corresponds to any assignment of the form $(\z_{S'}; \z'_{\Bar{S'}})$ for any $\z' \in \mathbb{F}$ (all of which are also classified as ``1'').


We note that this entire process is bounded by $O(m^k)$, where $m$ represents the maximum number of leaves in a tree and $k$ is the number of trees. This is also bounded by $O(|\mathcal{X}|^{O(k)})$. We now proceed to prove the following claim:


\begin{claim}
For two distinct partial assignments $\z_{S}$ and $\z'_{S'}$, obtained by iterating through the aforementioned procedure of selecting $j$ paths in $j$ distinct trees, it holds that there exists a feature $i \in S, S'$ such that $\z_i \neq \z'_i$.
\end{claim}

\emph{Proof.} Since $\z_S$ and $\z'{S'}$ are selected from iterating over two distinct choices of paths from trees, there must be at least one tree where the paths chosen for $\z_S$ and $\z'{S'}$ differ. Any two distinct paths in a tree contain at least one feature with differing assignments for some feature $i$. Therefore, based on the previous construction, it follows that there exists at least one feature $i$ where the partial assignments of $\z_S$ and $\z'_{S'}$ differ.


By definition, the following sum holds:

\begin{equation}
    \mathbb{E}_{\z\sim \mathcal{D}_p}[f(\z)] = \sum_{\z\in\mathbb{F}, f(\z)=1}\mathcal{D}_p(\z)= \sum_{\z\in\mathbb{F}, f(\z)=1}\Big({\displaystyle \prod_{i\in [n], \z_i=1} p(i)}\Big)\cdot \Big({\displaystyle \prod_{i\in [n], \z_i=0} (1-p(i))}\Big)
\end{equation}

That is, computing the expectation involves summing each $\mathcal{D}p(\z)$ for every positive assignment. Now, assume we have some partial assignment $\z_S$ obtained in the previous phase. Since this is only a partial assignment, we need to account for all possible completions of $\z_S$, or, in other words, any vector of the form $(\z_S; \z'{\Bar{S}})$. Let the set of all partial assignments computed using the aforementioned procedure be denoted by $\mathbb{S}$. Since any two partial assignments have a conflicting feature that does not ``match'', there is no ``overlap'' in the assignment completions corresponding to two distinct partial assignments. In other words, for two distinct partial assignments $\z_S$ and $\z'{S'}$, it holds that for all $\z'' \in \mathbb{F}$, $(\z_S; \z''{\Bar{S}}) \neq (\z'{S'}; \z''{\Bar{S'}})$. This leads to the following equivalence:


\begin{equation}
    \mathbb{E}_{\z\sim \mathcal{D}_p}[f(\z)] = \sum_{\z_S\in\mathbb{S}}\Big(\sum_{\z'_{\Bar{S}}\in\{0,1\}^{|\overline{S}|}}\mathcal{D}_p(\z_S;\z'_{\Bar{S}})\Big) 
\end{equation}

We do note, however, that for a fixed partial assignment $\z_S$, the following holds:

\begin{equation}
\begin{aligned}
     \sum_{\z'_{\Bar{S}}\in\{0,1\}^{|\overline{S}|}}\mathcal{D}_p(\z_S;\z'_{\Bar{S}})= \sum_{\z'_{\Bar{S}}\in\{0,1\}^{|\overline{S}|}}\Big({\displaystyle \prod_{i\in [n], (\z_S;\z'_{\Bar{S}})_i=1} p(i)}\Big)\cdot \Big({\displaystyle \prod_{i\in [n], (\z_S;\z'_{\Bar{S}})_i=0} (1-p(i))}\Big)=\\
     \sum_{\z'_{\Bar{S}}\in\{0,1\}^{|\overline{S}|}}\Big({\displaystyle \prod_{i\in S, \z_i=1} p(i)}\Big) \Big({\displaystyle \prod_{i\in S, \z_i=0} (1-p(i))}\Big) \Big({\displaystyle \prod_{i\in \overline{S}, \z'_i=1} p(i)}\Big) \Big({\displaystyle \prod_{i\in \overline{S}, \z'_i=0} (1-p(i))}\Big)=\\
     \Big({\displaystyle \prod_{i\in S, \z_i=1} p(i)}\Big)\cdot \Big({\displaystyle \prod_{i\in S, \z_i=0} (1-p(i))}\Big)
     \end{aligned}
\end{equation}

Overall, we obtain that:

\begin{equation}
\begin{aligned}
    \mathbb{E}_{\z\sim \mathcal{D}_p}[f(\z)] =  
\sum_{\z_S\in\mathbb{S}}\Big({\displaystyle \prod_{i\in S, \z_i=1} p(i)}\Big)\cdot \Big({\displaystyle \prod_{i\in S, \z_i=0} (1-p(i))}\Big)
     \end{aligned}
\end{equation}

Hence, after computing $\mathbb{S}$ in $O(|\mathcal{X}|^{O(k)})$ time, we can iterate over each partial assignment, compute its corresponding weight, and sum all the weights to obtain $\mathbb{E}_{\z \sim \mathcal{D}p}[f(\z)]$. This proves that the complexity of computing $\mathbb{E}{\z \sim \mathcal{D}_p}[f(\z)]$ is in XP. 

$\qedsymbol{}$

As explained earlier, this also establishes that the complexity of computing SHAP for some $\langle f, \x, i, \mathcal{D}_p \rangle$ is likewise in XP.

$\qedsymbol{}$

We have demonstrated that solving \emph{SHAP} when $\mathcal{D}_p$ represents any fully factorized distribution is $\#$W[1]-Hard and in XP. However, if we specifically set $\mathcal{D}_p$ to the \emph{uniform} distribution (i.e., where for any $i \in [n]$, $p(i) = \frac{1}{2}$), a specific type of fully factorized distribution, then this query becomes $\#$W[1]-Complete.

\begin{lemma}
\label{lemma_of_uniform_shap}
    Assuming that $\mathcal{D}_p$ is the uniform distribution, then the SHAP query for ensembles of $k$-decision trees is $\#$W[1]-Complete.
\end{lemma}

\emph{Proof.} The hardness is derived directly from our proof concerning fully factorized distributions, assuming a uniform distribution. For membership, we leverage the analysis by~\cite{arenas2021tractability}, which demonstrates that when $\mathcal{D}_p$ is uniform, computing \emph{SHAP} can be polynomially reduced to the problem of model counting (i.e., computing $\#f$). As outlined in Proposition~\ref{cc_appendix_k_ensemble_fbdd_proof}, the task of model counting for a $k$-ensemble of decision trees is $\#W[1]$-Complete. Thus, we conclude that computing \emph{SHAP} under the uniform distribution for $k$-ensembles of decision trees also achieves $\#$W[1]-Complete status.

$\qedsymbol{}$

\section{Proof of Proposition~\ref{msr_prop_main_paper}}
\label{msr_k_ensemble_fbdds_appendix_sec}

\begin{proposition}
    \label{msr_k_ensemble_fbdds_appendix}
    The MSR query for a $k$-ensemble of decision trees is para-NP-Hard and is in XNP.
\end{proposition}


\emph{Proof.} \textbf{Hardness.} Para-NP hardness is straightforward since the \emph{MSR} query for a single decision tree is already NP-Hard~\cite{BaMoPeSu20} and hence is obtained for $k=1$.

\textbf{Membership.} For membership in XNP we need to show that there exists a non-deterministic algorithm that solves this problem in $O(|\mathcal{X}|^k)$ time. Specifically, we will make use of the minimum-hitting-set (MHS) duality between sufficient and contrastive reasons to prove this claim~\cite{ignatiev2020contrastive}. First, we will define the MHS: 

\begin{definition}
Given a collection $\mathbb{S}$ of sets from a universe U, a hitting set $h$ for
$\mathbb{S}$ is a set such that
$\forall S \in \mathbb{S}, h\cap S \neq \emptyset$. A hitting set $h$
is said to be \emph{minimum} when it has the smallest possible cardinality among
ll hitting sets.
\end{definition}

We note that a subset minimal contrastive (sufficient) reason $S$ of $\langle f,\x\rangle$ is a contrastive (sufficient) reason that ceases to be a contrastive (Sufficient reason) when any feature $i$ is removef from it. In other words, for all $i$, it holds that $S\setminus i$ is not sufficient (contrastive). We now are in a position to use the following MHS duality between sufficient and contrastive reasons~\cite{ignatiev2020contrastive}:

\begin{lemma}
    The MHS of all subset minimal contrastive reasons with respect to $\langle f,\x\rangle$ is a cardinally minimal sufficient reason of $\langle f,\x\rangle$. Moreover, the MHS of all subset minimal sufficient reasons with respect to $\langle f,\x\rangle$ is a cardinally minimal contrastive reason of $\langle f,\x\rangle$.
\end{lemma}

Now, we describe the following preprocessing stage which runs in time $O(|\mathcal{X}|^k)$ and computes all of the subset minimal contrastive reasons of $\langle f,\x\rangle$.

\begin{claim}
Given an ensemble of decision trees $f$ with $k$ decision trees, there exists an algorithm that computes all of the subset minimal contrastive reasons of $\langle f,\x\rangle$ in time $O(m^k)$, where $m$ denotes the maximal number of leaf nodes in a decision tree within $f$. 
\end{claim}

\emph{Proof.} We iterate over combinations of choosing one leaf (which corresponds to one path) from every distinct tree in $f$. This process can be done in $O(m^k)$ time. We check whether two conditions hold. First, we check whether every pair of two distinct paths that belong to two distinct trees ``matches'' and whether more than $\ceil{\frac{k}{2}}$ of these trees terminate over a leaf with a $\neg f(\x)$ assignment. For any combination of $k$ distinct paths in which any two paths ``match'', it means that there is some partial assignment $\z_{S'}$, which describes the corresponding assignments in each one of these paths (there is necessarily one such assignment since each two paths ``match'') and it holds that:

\begin{equation}
    \forall \y\in\mathbb{F} \quad [f'(\z_{S'};\y_{\Bar{S'}})\neq f(\x)] 
\end{equation}

We now will denote by $S''$, the subset of features in $\z_{S'}$ that do not match with $\x$. More specifically:

\begin{equation}
    S'':= |\{i\in\{1,\ldots S'\} \ \ \z_i\neq\x_i| 
\end{equation}

For each combination of $k$ paths, for which the corresponding two conditions hold, we compute $S''$. We denote the set of all such subsets as $\mathbb{S}$. We will now prove the following claim, which will finish the proof of our lemma:

\begin{claim}
Any subset minimal contrastive reason of $\langle f,\x\rangle$ is contained in $\mathbb{S}$
\end{claim}

\emph{Proof.} Let us assume towards contradiction that this claim does not hold. In other words, there exists a subset $S$ which is a subset minimal contrastive reason of $\langle f,\x\rangle$ and that is not chosen to be in $\mathbb{S}$ by our algorithm. Since $S$ is a contrastive reason it holds that:

\begin{equation}
    \exists \z\in\mathbb{F} \quad [f(\x_{\Bar{S}};\z_{S})\neq f(\x)] 
\end{equation}

However, we will prove a stronger property that holds if $S$ is a \emph{subset minimal} contrastive reason of $\langle f,\x\rangle$. Specifically, it holds that:

\begin{equation}
[f(\x_{\Bar{S}};\neg \x_{S})\neq f(\x)] 
\end{equation}

or in other words the specific vector $\z$ for which fixing $S$ to, changes the classification is $\neg \x$. The proof to this claim is straightforward --- let us assume towards contradiction that this is not the case, or in other words, it holds that:

\begin{equation}
[f(\x_{\Bar{S}};\neg \x_{S})= f(\x)] 
\end{equation}

Since $S$ is a contrastive reason, this means that there exists some other assignment to the features of $S$ (which is not $\neg \x$, which causes the classification to change. In other words, there exists some $\y\neq\neg\x$ for which:

\begin{equation}
[f(\x_{\Bar{S}};\y_{S})\neq f(\x)] 
\end{equation}

Since $\y\neq\neg\x$ over the features in $S$, this means that there is at least one feature $i\in S$ for which $\y_i=\x_i$. Let us now denote $S_0:=S\setminus \{i\}$. We hence get that:

\begin{equation}
[f(\x_{\Bar{S}};\y_{S})=f(\x_{\Bar{S_0}};\neg\x_{S_0})\neq f(\x)] 
\end{equation}

This implies that $S_0$ is a contrastive reason of $\langle f,\x\rangle$, hence contradicting the subset minimality of $S$. This concludes the proof of this claim, and we hence derive in the fact that since $S$ is a subset minimal contrastive reason of $\langle f,\x\rangle$, it must hold that:

\begin{equation}
[f(\x_{\Bar{S}};\neg \x_{S})\neq f(\x)] 
\end{equation}

We hence can take the assignment $(\x_{\Bar{S}};\neg \x_{S})$ and propagate it through $f$. The assignment $f_i(\x_{\Bar{S}};\neg \x_{S})$ will lead to some path, for which there are at least $\ceil{\frac{k}{2}}$ which terminate on a $\neg f(\x)$ node. These paths, of course ``match'' (meaning any pair of two distinct paths ``match''), because they correspond to \emph{one} assignment of features, and hence will be chosen as a combination of paths by our algorithm. 

We recall that our algorithm chooses the partial assignment which assigns a value to each one of the features in each path of each tree (there exists one such assignment). In our case this will be some partial assignment of $(\x_{\Bar{S}};\neg \x_{S})$. The algorithm then chooses the subset $S''$ which is the subset of features whose values (in our case of $(\x_{\Bar{S}};\neg \x_{S})$) that are different than those of $\x$. Hence, it necessarily holds that $S''\subseteq S$. 

We note that $S''$ is a contrastive reason concerning $\langle f,\x\rangle$, since by construction, it describes a subset of features such that if we set them to values that are not of $\x$ (specifically $\neg \x$), $\ceil{\frac{k}{2}}$ or more trees in the ensemble terminate on a $\neg f_i(\x)$ assignment. Since we have proven that $S''\subseteq S$, then it must hold that $S''=S$, since otherwise, this will contradict the subset minimality of $S$. We hence derive in the fact that $S''=S$ is chosen during our algorithm to be in $\mathbb{S}$, contradicting the assumption that $S$ is not in $\mathbb{S}$. This concludes the proof that any subset minimal contrastive reason of $\langle f,\x\rangle$ is in $\mathbb{S}$.

$\qedsymbol{}$

We will now use the MHS duality to conclude our proof regarding membership in XNP. We recall that if $\mathbb{S}$ is the set containing all subset minimal contrastive reasons of $\langle f,\x\rangle$, then the MHS of $\mathbb{S}$ is the cardinally minimal sufficient reason of $\langle f,\x\rangle$. We note that if we add to $\mathbb{S}$ (which contains all subset minimal contrastive reasons of $\langle f,\x\rangle$ other (non-subset-minimal) contrastive reasons of $\langle f,\x\rangle$ the MHS of $\mathbb{S}$ will remain the same. This is true since the hitting set of any two subsets $S$ and $S'$, when $S\subseteq S'$, is equal to $S$.

We have proven that the set $\mathbb{S}$ that is obtained by our algorithm contains all subset minimal contrastive reasons of $\langle f,\x\rangle$, as well as perhaps other contrastive reasons. Hence, the MHS of $\mathbb{S}$ is a cardinally minimal sufficient reason of $\langle f,\x\rangle$. 

Hence, we can simply non-deterministically guess some subset $S_1\subseteq \{1,\ldots,n\}$, and check whether $S_1$ intersects with all subsets in $\mathbb{S}$ (and hence is the MHS of $\mathbb{S}$, and a cardinally minimal sufficient reason of $\langle f,\x\rangle$. If $|S_1|\leq d$, our algorithm can return true, and otherwise, it will return false. Overall, the entire algorithm that we described performs a preprocessing step in $O(m^k)$ time (and hence is bounded by $O(|\mathcal{X}|^k)$ time), and then performs a non-deterministic step, also bounded by $O(|\mathcal{X}|^k)$ time. This concludes the proof that solving the MHS query for an ensemble of $k$ decision trees is contained in XNP.

$\qedsymbol{}$




\section{Proof of Proposition~\ref{subset_minimal_proof_main}}
\label{xp_subset_minimal_proof_sec}

\begin{proposition}
\label{xp_subset_minimal_proof}
    Obtaining a subset-minimal sufficient reason for a $k$-ensemble of decision trees is in XP.
\end{proposition}

\emph{Proof.} The specific result aligns with a result demonstrated by~\cite{ordyniak2024explaining}. However, we can further establish this result as a direct extension of the complexity result for the CSR query (Lemma~\ref{k-ensemble-fbdds-csr}), which will demonstrate this result not only for majority voting ensembles. To show this, we use a common greedy algorithm that computes a subset-minimal sufficient reason by invoking a linear number of queries, each checking whether a given subset is a sufficient reason~\cite{ignatiev2019abduction}. Intuitively, the algorithm attempts to ``free'' a feature from the subset $S$ at each iteration, until finally converging to a subset-minimal sufficient reason.

\begin{algorithm}
\label{algorithm_for_paper}
    \newcommand{\algorithmicforeach}{\textbf{for each}}
\newcommand{\ForEach}[2]{\STATE \algorithmicforeach\ #1 \textbf{do} #2}
\newcommand{\Return}[1]{\STATE \textbf{return} #1}
\newcommand{\EndForEach}{\STATE \textbf{end for each}}

	\textbf{Input} $f$, $\x$
	\caption{Greedy Subset Minimal Sufficient Reason Search}\label{alg:subset-minimal-local}
	\begin{algorithmic}[1]
		\STATE $S \gets \{1,\ldots,n\}$
		\ForEach {$i \in \{1,...,n\}$ by some arbitrary ordering}\label{lst:line:orderingline}
		\IF{$S\setminus \{i\}$ is a sufficient reason w.r.t $\langle f,\x\rangle$}\label{lst:line:sufficientline}
		\STATE $S \gets S\setminus \{i\}$
		\ENDIF
		\EndForEach\label{lst:line:endregularupper}
		\STATE \Return $S$ \COMMENT{$S$ is a \emph{subset minimal} sufficient reason}
	\end{algorithmic}
\end{algorithm}

Lemma~\ref{k-ensemble-fbdds-csr} proved that the \emph{CSR} query for an ensemble of decision trees is in coW[1] (and hence is also in XP). In essesnse, algorithm~\ref{algorithm_for_paper} implies that a linear number of queries to the \emph{CSR} query produces a subset-minimal sufficient reason. It hence, directly follows that obtaining some subset minimal sufficient reason for some $\langle f,\x\rangle$ is also in XP.

$\qedsymbol{}$

\end{document}